\documentclass[twoside,11pt]{article}

\usepackage[T1]{fontenc}
\usepackage[utf8]{inputenc}
\usepackage{times}
\usepackage{microtype}
\usepackage{booktabs}
\usepackage{caption}
\usepackage{url}
\usepackage[dvipsnames]{xcolor}
\usepackage{amsmath,amssymb,amsthm,amsopn,amsbsy,amsgen,amstext}
\providecommand{\quickfigs}{0}
\ifnum\quickfigs=1 \PassOptionsToPackage{draft}{graphicx}\fi
\usepackage[preprint,nohyperref]{jmlr2e}

\usepackage{bm}
\usepackage{mathtools}
\usepackage{mathrsfs}
\usepackage{dsfont}

\usepackage{graphicx}
\usepackage{subcaption}
\usepackage{array}
\usepackage{multirow}
\usepackage{float}
\usepackage{placeins}
\usepackage{nicematrix}
\usepackage{tikz}
\usepackage{pgfplots}
\pgfplotsset{compat=1.18}

\makeatletter
\long\def\@makecaption#1#2{%
  \vskip\abovecaptionskip
  \setbox\@tempboxa\hbox{#1: #2}%
  \ifdim\wd\@tempboxa>\hsize
    \begin{list}{#1:}{%
      \setlength{\topsep}{0pt}%
      \setlength{\partopsep}{0pt}%
      \setlength{\parsep}{0pt}%
      \setlength{\itemsep}{0pt}%
      \settowidth{\labelwidth}{#1:}%
      \setlength{\leftmargin}{\labelwidth}%
      \addtolength{\leftmargin}{\labelsep}%
    }\item #2\end{list}\par
  \else
    \hbox to\hsize{\hfil\box\@tempboxa\hfil}%
  \fi
  \vskip\belowcaptionskip
}
\makeatother
\usepackage[linesnumbered,ruled,vlined]{algorithm2e}
\RestyleAlgo{ruled}

\usepackage{enumitem}
\usepackage{tcolorbox}
\usepackage{titletoc}

\usepackage[dvipsnames]{xcolor}

\definecolor{niblue}{HTML}{4A6FA5}      %
\definecolor{nisoftblue}{HTML}{AFC6E9}  %
\definecolor{nired}{HTML}{C76C6C}       %
\definecolor{nisoftred}{HTML}{E7BCB5}   %
\definecolor{nigreen}{HTML}{6E9075}     %
\definecolor{nisoftgreen}{HTML}{CFE3D6} %
\definecolor{nigray}{HTML}{B9C0C8}      %
\definecolor{nisand}{HTML}{E7DCC3}      %

\definecolor{revcolor}{HTML}{7B3FA0}     %

\usepackage{subfiles}

\newcommand{\RR}{\mathbb{R}}
\newcommand{\EE}{\mathbb{E}}
\newcommand{\Cc}{\mathcal{C}}
\newcommand{\TT}{\mathbb{T}}
\newcommand{\QQ}{\mathbb{Q}}

\newcommand{\Fc}{\mathcal{F}}
\newcommand{\Dc}{\mathcal{D}}
\newcommand{\Pc}{\mathcal{P}}

\newcommand{\Xc}{\mathcal{X}}
\newcommand{\Yc}{\mathcal{Y}}

\newcommand{\Zc}{\mathcal{Z}}

\newcommand{\Tr}{{\rm Tr}}
\newcommand{\Diag}{{\rm Diag}}
\newcommand{\TTl}[1]{\mathbb{T}^{(#1)}}
\newcommand{\X}[1]{X^{(#1)}}
\DeclareMathOperator*{\argmin}{arg\,min}
\DeclareMathOperator*{\argmax}{arg\,max}

\newtheorem{Theorem}{Theorem}

\newtheorem{Proposition}{Proposition}
\newtheorem{Condition}{Condition}
\newtheorem{Lemma}{Lemma}

\newtheorem{Definition}{Definition}

\usepackage[colorlinks=true,linkcolor=blue,citecolor=blue,urlcolor=blue]{hyperref}
\usepackage{bookmark}

\ShortHeadings{StablePCA: Shared Representations from Multi-Source Data}{Wang, Liu, Lei, Bach and Guo}
\firstpageno{1}

\begin{document}

\title{StablePCA: Distributionally Robust Learning of Shared Representations from Multi-Source Data}

\author{\name Zhenyu Wang \email zw425@stat.rutgers.edu \\
\addr Department of Statistics, Rutgers University, USA
\AND
\name Molei Liu \email moleiliu@bjmu.edu.cn \\
\addr Department of Biostatistics, Peking University Health Science Center;\\
Beijing International Center for Mathematical Research, China
\AND
\name Jing Lei \email jinglei@andrew.cmu.edu \\
\addr Department of Statistics, Carnegie Mellon University, USA
\AND
\name Francis Bach$^*$ \email francis.bach@inria.fr \\
\addr Inria, \'Ecole Normale Sup\'erieure, PSL Research University, France
\AND
\name Zijian Guo$^*$ \email zijguo@zju.edu.cn \\
\addr Center for Data Science, Zhejiang University, China}

\maketitle
\thispagestyle{plain}
{\renewcommand{\thefootnote}{\fnsymbol{footnote}}\footnotetext[1]{Corresponding authors: Francis Bach (francis.bach@inria.fr) and Zijian Guo (zijguo@zju.edu.cn).}}

\begin{abstract}
When synthesizing multi-source high-dimensional data, a key objective is to extract low-dimensional representations that effectively approximate the original features across different sources. Such representations facilitate the discovery of transferable structures and help mitigate systematic biases such as batch effects.
We introduce \emph{Stable Principal Component Analysis} (StablePCA), a distributionally robust framework for constructing stable latent representations by maximizing the worst-case explained variance over multiple sources. A primary challenge in extending classical PCA to the multi-source setting lies in the nonconvex rank constraint, which renders the StablePCA formulation a nonconvex optimization problem. To overcome this challenge, we conduct a convex relaxation of StablePCA and develop an efficient Mirror-Prox algorithm to solve the relaxed problem, with global convergence guarantees. Since the relaxed problem generally differs from the original formulation, we further introduce a data-dependent certificate to assess how well the algorithm solves the original nonconvex problem and establish the condition under which the relaxation is tight.
Finally, we explore alternative distributionally robust formulations of multi-source PCA based on different loss functions.
\end{abstract}


\section{Introduction}

Extracting low-dimensional representations from high-dimensional data is a fundamental task in modern data science. Such representations are widely used for downstream tasks, including visualization, clustering, prediction, and knowledge discovery \citep{hastie2009elements, jolliffe2011principal}. 
To extract these representations, classical approaches such as principal component analysis (PCA) and matrix factorization (MF) learn a low-rank structure that captures the dominant variation in the observed data by minimizing the reconstruction error. However, a structure learned in this way is optimized for the training distribution and may fail to generalize to data drawn from a different distribution.
This raises a fundamental question: how can one learn a low-rank transformation that remains informative for future unseen data under distributional shift?

Multi-source data provides an opportunity to learn generalizable low-rank transformations when the unseen target data has a distributional shift from the training data.
Such multi-data arise commonly, including single-cell RNA sequencing (scRNA-seq) studies measured across multiple batches \citep{hicks2018missing, tran2020benchmark}, electronic health records (EHRs) collected from different hospitals \citep{hong2021clinical,li2024multisource}, and medical images acquired under varying protocols \citep{hari2021using}. Although each source carries its own source-specific variation, the sources tend to share an underlying structure, such as common biological programs, that generalizes to future data.
This observation motivates leveraging heterogeneous sources to identify the shared low-rank transformation while acknowledging the source-specific shifts and mitigating source-specific biases.

To achieve a shared low-rank transformation, a naive strategy is to pool data from multiple sources and then apply PCA to the aggregated dataset. However, this strategy relies on the questionable assumption that source-specific biases cancel out when averaged across sources, but such an assumption might not hold in practice, as these biases are frequently skewed or imbalanced \citep{hari2021using}.
For instance, in scRNA-seq studies, batch effects arise from technical variations in sequencing platforms, reagents, and experimental protocols. Directly pooling data for PCA fails to remove these artifacts, which can artificially separate cells of the same biological type, leading to biased cell-type annotation and distorted differential gene expression analyses \citep{tran2020benchmark}.
Moreover, when sources differ substantially in sample size or noise level, the learned low-rank structure may be dominated by a few large or high-variance sources, while underrepresented sources contribute little to the representation. As a result, naive pooling often leads to poor out-of-distribution generalization and raises fairness concerns when the learned representations are used downstream. 

\subsection{Our results and contributions}

In this work, we propose \emph{Stable Principal Component Analysis} (StablePCA), a distributionally robust framework to identify a stable low-rank structure based on multi-source data. 
As explained in Section \ref{subsec: interpretation revision}, under the shared latent factor model, StablePCA recovers the subspace shared across sources despite source-specific variation. This geometric interpretation motivates the name StablePCA for the work.

To ensure that the learned structure generalizes to future data, whose distribution may differ from the observed sources, we construct an uncertainty set for the unknown target distribution that consists of all mixtures of the source distributions.
StablePCA then seeks a low-rank subspace that maximizes the worst-case explained variance over this uncertainty set. This formulation is designed to learn a low-rank transformation that generalizes beyond the observed sources and performs well on new target data.

The primary challenge in solving StablePCA arises from the nonconvexity associated with the fixed-rank constraint set for projection matrices.
To address this issue, we employ a Fantope relaxation \citep{vu2013fantope}, enabling a convex reformulation of the original nonconvex StablePCA. 
We then develop a Mirror-Prox algorithm to efficiently solve the relaxed problem.
Unlike standard gradient descent–ascent methods, Mirror-Prox  \citep{nemirovski2004prox,bubeck2015convex} incorporates an extra-gradient step and is specifically designed for constrained minimax problems with non-Euclidean geometry, as in our case.
Building upon this, we derive explicit closed-form updates for each iteration of the Mirror-Prox algorithm, ensuring efficient implementation. From a theoretical perspective, we establish global convergence guarantees for the proposed algorithm, with rates characterized in terms of both the sample size $n$ and the number of iterations $T$.

Since the relaxed StablePCA generally differs from the original nonconvex formulation, we propose a data-dependent computable certificate to assess how well the proposed algorithm approximately solves the original StablePCA. 
Moreover, we show that, under an additional eigengap condition, the relaxation is tight, meaning that the solutions and optimal values of the relaxed and original StablePCA problems coincide. Consequently, the proposed algorithm also solves the original nonconvex StablePCA with global convergence guarantees.

Finally, we explore alternative distributionally robust formulations of multi-source PCA beyond the proposed
StablePCA, including variants based on squared loss and regret, where the regret loss recovers the FairPCA formulation in the literature \citep{samadi2018price, shen2025hidden}. 
An interesting observation is that different choices of loss functions lead to construction of different low-rank transformations in the multi-source setting. We design the proposed Mirror-Prox algorithm for these alternative formulations. In contrast to the semidefinite programming (SDP) method used in prior work \citep{samadi2018price}, which incurs a computational complexity of $\mathcal{O}(d^{6.5})$ for ambient dimension $d$, our method yields a scalable gradient-based algorithm with a runtime of $\mathcal{O}(d^3T)$, where $T$ denotes the number of iterations. 
As illustrated in Table \ref{tab:compare-sdp} of Section \ref{subsec: alternative}, our approach achieves substantial computational gains when $d$ is moderately large. For example, it is nearly 40 times faster than SDP when $d=300$.

We summarize the key contributions of this work as follows: 

\begin{enumerate}
    \item We propose StablePCA, a distributionally robust framework for learning stable low-rank transformations that generalize beyond the observed sources. We further explore and compare alternative robust formulations of multi-source PCAs using various loss functions.

    \item For the relaxed (convex) StablePCA, we develop a computationally efficient algorithm and establish global convergence guarantees that jointly characterize optimization convergence and statistical estimation errors (Theorem \ref{thm: new conv}).

    \item For the original (nonconvex) StablePCA, we introduce a data-dependent certificate to quantify how well the proposed algorithm solves it to  global optimality (Theorem \ref{thm: global conv rates alg}), and we establish a sufficient eigengap condition under which the convex relaxation is tight.

\end{enumerate}

\subsection{Related literature}
This section reviews relevant research and highlights how the proposed StablePCA relates to and differs from existing works. 

\vspace{0.5em}
\noindent{\bf Group DRO and Maximin effect.}
Group Distributionally Robust Optimization (DRO) \citep{sagawa2019distributionally} and the maximin
effect framework \citep{guo2022statistical,wang2023distributionally} have been extensively studied in supervised
learning for constructing prediction models that generalize across heterogeneous
source domains.
Recent extensions include the study of optimal sample allocation for Group DRO in online learning settings
\citep{zhang2024optimal} and adaptations to semi-supervised learning \citep{awasthi2024semi}.
In contrast to these works focusing on supervised prediction tasks, StablePCA addresses the unsupervised learning setting, aiming to extract shared latent structures from multi-source data. 

\vspace{0.5em}
\noindent{\bf Multi-source PCA.}
Several formulations of PCA in multi-source settings have been proposed in the literature.
For example, \citet{samadi2018price} and \citet{shen2025hidden} introduced FairPCA, which minimizes the worst-case regret across sources, while \citet{olfat2019convex} and \citet{kleindessner2023efficient} studied PCA variants that aim to remove source-specific information from learned representations.
In contrast, StablePCA is formulated from a distributionally robust optimization perspective, explicitly maximizing the worst-case explained variance over all mixtures of source distributions.
To address the nonconvexity of the fixed-rank constraint, we adopt the Fantope relaxation and develop the gradient-based Mirror-Prox algorithm to efficiently solve the resulting minimax problem. 
We note that our algorithm extends to solve FairPCA, avoiding the use of
semidefinite programming in the original work \citep{samadi2018price}, which is computationally prohibitive in large dimensions.
Moreover, in contrast to prior works, we establish the global convergence guarantees of our proposal in terms of both the sample size $n$ and the iteration number $T$.

 \vspace{0.5em}
\noindent{\bf Multi-source Learning.}
{ A broader body of work in multi-source learning aims to identify generalizable patterns from heterogeneous data sources. For transfer learning, \citet{li2022transfer} and  \citet{tian2023transfer} proposed to transfer knowledge from source domains to a target domain by leveraging structural similarity between source and target distributions. In meta-analysis, \citet{maity2022meta} and \citet{guo2025robust} identify and study models that the majority of source domains hold. For causal invariance learning, existing works \citep{peters2016causal, fan2024environment, wang2024causal} aim to learn causal models that remain invariant across sources. These works are typically designed for supervised settings with outcomes. In contrast, StablePCA addresses the problem in an unsupervised setting by identifying stable low-dimensional representations of data across heterogeneous sources.}

\subsection{Notations}

We define $[m] = \{1, 2, ..., m\}$ for the positive integer $m$.
For a vector $v\in \RR^d$, its $\ell_p$-norm ($p\geq 1$) is given by $\|v\|_p = (\sum_{j=1}^d |v_j|^p)^{1/p}$. 
We denote by ${\rm Diag}(v)$ the diagonal matrix whose diagonal entries are given by the components of $v$. We use ${\bf 1}_d$ to denote the $d$-dimensional vector of all ones.
For two matrices $A, B$ with compatible dimensions, we define the matrix inner product $\langle A, B\rangle={\rm Tr}(A^{\intercal}B)$ and the Frobenius norm $\|A\|_F=\sqrt{\langle A,A\rangle}$. 
If $A\in \RR^{d\times d}$ is symmetric, we denote its eigenvalues in nonincreasing order by  $\lambda_1(A)\geq \cdots \geq \lambda_d(A)$. 
If $A$ is positive definite with eigen-decomposition $A=U{\rm Diag}(\lambda) U^\intercal$, we define $\log(A)=U {\rm Diag}(\log \lambda) U^\intercal$. 
The $d$-dimensional identity matrix is denoted as ${\bf I}_d$. For sequences $a(n), b(n)$, we use $a(n) = \mathcal{O}(b(n))$ to represent that there exists some universal constant $C > 0$ such that $a(n) \leq C \cdot b(n)$ for all $n \geq 1$.

\section{StablePCA: Definitions and Interpretations}

This section introduces the definition of Stable PCA and provides interpretations of how it identifies a stable low-rank transformation across domains. We begin in Section~\ref{subsec: classical pca} with reviewing the classical PCA (in the single-source setting), which lays the foundation for defining StablePCA using multi-source data in Section~\ref{subsec: stablepca}. Finally, Section~\ref{subsec: interpretation revision} illustrates how StablePCA captures stable latent structures across heterogeneous sources.

\subsection{Review of Classical PCA}
\label{subsec: classical pca}
The central goal of PCA is to identify a low-rank linear subspace that captures the dominant variation in the data. We now present two equivalent formulations of PCA: one based on minimizing reconstruction error and the other based on maximizing explained variance. %

We start with the reconstruction-error formulation.
Let $d$ denote the ambient dimension. For a random feature vector $X\in \RR^{d}$, classical PCA seeks a $k$-dimensional linear subspace (with $k< d$) such that projecting $X$ onto this subspace yields the best approximation of $X$ itself.
Formally, let 
$\mathcal{O}^{d\times k}=\{V\in \RR^{d \times k}: V^\intercal V={\bf I}_{k}\}$ denote the set of $d \times k$ matrices with orthonormal columns. For each $V\in \mathcal{O}^{d\times k}$, its column space ${\rm col}(V)$ defines a $k$-dimensional linear subspace, and the corresponding matrix $P = VV^\intercal$ is the orthogonal projection onto this subspace (with dimension $k$). 
PCA is then formulated as identifying the rank-$k$ projection matrix $P$ that minimizes the expected reconstruction error $X - PX$:
\begin{equation}
P^* \in \argmin_{P\in \Pc^k}\  \EE\|X - P X\|_2^2,\quad \textrm{with}\quad \Pc^k = \left\{P\in \RR^{d\times d}: P=VV^\intercal, V\in \mathcal{O}^{d\times k}\right\},
\label{eq: standardPCA}
\end{equation}
where $PX$ represents the projection of $X$ onto the associated subspace, and $\Pc^k$ denotes the set of all rank-$k$ projection matrices. 

After learning $P^*$, we shall further leverage it to construct a low-dimensional representation of the original feature vector $X$. Specifically, decomposing $P^* = V^*V^{*\intercal}$ for some $V^*\in \mathcal{O}^{d\times k}$, we let $V^{*\intercal}X\in \RR^{k}$ be the $k$-dimensional representation of $X\in \RR^d$. 
 
Next, we provide an alternative interpretation of PCA in terms of explained variance. We rewrite the optimization problem in \eqref{eq: standardPCA} as follows:
\begin{equation}
P^* \in \argmax_{P\in \Pc^k}\  \EE\left(\|X\|_2^2 - \|X - P X\|_2^2\right)  = \argmax_{P\in \Pc^k} \ \langle \Sigma, P\rangle,
    \label{eq: standardPCA alter}
\end{equation}
where $\Sigma = \EE[XX^\intercal]$ denotes the second-moment matrix of $X$, and $\EE\left(\|X\|_2^2 - \|X - P X\|_2^2\right)=\langle\Sigma, P\rangle$ denotes the variance explained by the projection matrix $P$. Throughout this work, we slightly abuse terminology and refer to $\langle\Sigma, P\rangle$ as the variance explained by using $P$, even for the uncentered $X$. 
In practice, one can center the data by subtracting the sample mean prior to applying PCA. Therefore, $P^*$ is the rank-$k$ projection matrix that maximizes the explained variance of $X$.

Although the reconstruction-error and explained-variance formulations are equivalent for a single source, they behave differently when generalized to multi-source data. This work primarily focus on the explained-variance formulation \eqref{eq: standardPCA alter}, because its multi-source extension recovers the structure shared across heterogeneous sources, as shown in Section~\ref{subsec: interpretation revision}. We also consider the multi-source extension of the reconstruction-error formulation in Section~\ref{subsec: alternative}.

Despite the widespread use of classical PCA, the learned projection matrix $P^*$ is optimal only for the observed distribution and may fail to generalize to future target data whose distribution may shift. For example, applying $P^*$ to new data $\widetilde{X}$ drawn from a different distribution may yield a poor explained variance $\EE\left(\|\widetilde{X}\|_2^2 - \|\widetilde{X} - P^*\widetilde{X}\|_2^2\right)$. The challenge is thus to construct a low-rank structure that generalizes beyond the observed data. In what follows, we address it by leveraging data from multiple sources and constructing the low-rank transformation from the viewpoint of distributional robustness.

\subsection{Formulation of StablePCA}
\label{subsec: stablepca}

We now turn to the multi-source regime with data collected from multiple heterogeneous environments. Our goal is to identify a stable low-rank transformation that generalizes beyond the observed source distributions, in the presence of distributional shifts. We consider access to $L$ source datasets, where for  $1\leq l\leq L,$ we use $\TT^{(l)}$ to denote the $l$-th source distribution and $X^{(l)}\in \RR^d$ to denote the random feature vector drawn from $\TT^{(l)}$. 

We introduce an uncertainty class to account for the uncertainty about the unknown target distribution. Particularly, we introduce the following uncertainty class, comprising all mixtures of the source distributions:
\begin{equation}
\Cc=\left\{\QQ:\QQ=\sum_{l=1}^{L}\omega_{l}\cdot  \TT^{(l)}, \; \omega\in \Delta^{L}\right\},
\label{eq: uncertainty class}
\end{equation}
where $\Delta^L=\{\omega\in \RR_+^L: \sum_{l=1}^L \omega_l=1\}$ denotes the $(L-1)$-dimensional simplex. 

With such an uncertainty class, we generalize classical PCA in \eqref{eq: standardPCA alter} by considering the worst-case explained variance and define StablePCA as:
\begin{equation}
    P^* \in \argmax_{P\in \Pc^k} \min_{\QQ\in \Cc}\EE_{X\sim \QQ} \left(\|X\|_{2}^2 - \|X- P X\|_{2}^2\right),
    \label{eq: stablepca-original}
\end{equation}
where the expectation is taken with respect to the random feature $X$ drawn from the distribution $\QQ$. With a slight abuse of notation, we denote the optimal solution by $P^*$; unless otherwise specified, $P^*$ will henceforth refer to the solution to the StablePCA optimization problem defined in \eqref{eq: stablepca-original}.

When there is no distributional shift, that is, $\TT^{(1)}=\cdots=\TT^{(L)}$, the uncertainty class $\Cc$ collapses to a single distribution, and StablePCA reduces to classical PCA in \eqref{eq: standardPCA alter}. When the sources differ, the inner minimization of \eqref{eq: stablepca-original} selects, for a given projection matrix $P$, the mixture of sources in which $P$ explains the least variance. As a result, StablePCA tends to yield a more generalizable low-rank transformation by evaluating every subspace at its worst-case mixture.

We now simplify the optimization problem in \eqref{eq: stablepca-original} to facilitate algorithm design and further analysis. We apply the definition of the uncertainty class $\mathcal{C}$ and obtain 
\[
\begin{aligned}
    \min_{\QQ\in \Cc}\EE_{X\sim \QQ} \left(\|X\|_{2}^2 - \|X- P X\|_{2}^2\right) &=  \min_{\omega\in \Delta^L}\sum_{l=1}^{L}\omega_l\cdot \EE_{X\sim \TT^{(l)}} \left(\|X\|_{2}^2 - \|X- P X\|_{2}^2\right)\\
    &=\min_{\omega\in \Delta^L}\sum_{l=1}^L \omega_l\cdot  \langle \Sigma^{(l)}, P\rangle,
\end{aligned}
\]
where $\Sigma^{(l)} = \EE[X^{(l)} X^{(l)\intercal}]$. Therefore, our proposed StablePCA in \eqref{eq: stablepca-original} simplifies as follows:
\begin{equation}
    P^*\in \argmax_{P\in \Pc^k} \min_{\omega\in \Delta^L}\sum_{l=1}^L \omega_l\cdot  \langle \Sigma^{(l)}, P\rangle.
\label{eq: def diStablePCA}
\end{equation}
After learning $P^*$, we write $P^*=V^*V^{*\intercal}$ for some orthonormal matrix $V^*\in \mathcal{O}^{d\times k}$ and use $V^{*\intercal}{X}\in \RR^k$ as a low-dimensional representation of ${X}.$ {
In Appendix \ref{subsec: prior}, we further discuss incorporating prior information about the target distribution into defining StablePCA.}

\subsection{Geometric Interpretation: Shared Structure in Latent Factor Models}
\label{subsec: interpretation revision}
In this subsection, we provide the geometric interpretation for StablePCA by showing that it recovers the shared structure among multi-source data. Particularly, we study a shared latent factor model as follows:
\begin{equation}
 X^{(l)}=VZ^{(l)}+\Gamma^{(l)}U^{(l)},
 \label{eq: batch model revision}
\end{equation}
This shared latent factor model has also been studied in multi-study factor analysis \citep{devito2019multistudy} and personalized PCA \citep{shi2024personalized}.
In \eqref{eq: batch model revision}, $V\in\mathcal{O}^{d\times k}$ spans the shared subspace, and $\Gamma^{(l)}\in\mathcal{O}^{d\times q_l}$ spans the source-specific subspace. Both $\Gamma^{(l)}$ and its dimension $q_l$ may vary across sources. The factor vectors $Z^{(l)}\in\RR^k$ and $U^{(l)}\in\RR^{q_l}$ are centered and may have source-dependent covariance matrices.

For gene expression data collected across batches, this model can describe shared biological variation and batch-specific technical variation \citep{welch2019single,buettner2015computational,hicks2018missing}. Each column of $V$ represents a biological program shared across batches, such as a cell-type identity or a cell-cycle program, and $Z^{(l)}$ records the activities of these programs in a cell. Each column of $\Gamma^{(l)}$ represents a technical artifact specific to batch $l$, and $U^{(l)}$ records the magnitude of each artifact.

We assume that the source-specific directions are orthogonal to the shared directions,
\begin{equation}
 \Gamma^{(l)\intercal}V=0\qquad\text{for every source }l,
 \label{eq: orthogonality revision}
\end{equation}
and that $Z^{(l)}$ and $U^{(l)}$ are uncorrelated within each source. Then the second-moment matrix of $X^{(l)}$ is
\begin{equation}
\begin{aligned}
 \Sigma^{(l)}&=\EE[X^{(l)}X^{(l)\intercal}]=V\EE[Z^{(l)}Z^{(l)\intercal}]V^\intercal
 +\Gamma^{(l)}\EE[U^{(l)}U^{(l)\intercal}]\Gamma^{(l)\intercal}.
\end{aligned}
 \label{eq: shared model revision}
\end{equation}
Our goal is to recover the shared subspace ${\rm col}(V)$. In the gene expression example, projection onto this subspace removes batch artifacts and retains biological variation, with $V^\intercal X^{(l)}=Z^{(l)}$ providing the activities of the shared biological programs. %

The following proposition bounds the error in recovering the shared subspace ${\rm col}(V)$ for StablePCA. The proof is given in Appendix~\ref{appendix: general recovery}.
For each source $l\in[L]$, let $s_l=\EE\|Z^{(l)}\|_2^2$ be the total variance in the shared subspace, and define
\[
 \Delta_l=\lambda_{\min}\big(\EE[Z^{(l)}Z^{(l)\intercal}]\big)
 -\lambda_{\max}\big(\EE[U^{(l)}U^{(l)\intercal}]\big).
\]
The condition $\Delta_l>0$ indicates that, in source $l$, the shared subspace has greater variation strength than the source-specific subspace.

\begin{Proposition}[Recovery of a shared subspace]
\label{prop: shared recovery revision}
Assume \eqref{eq: batch model revision}--\eqref{eq: orthogonality revision} and suppose that $Z^{(l)}$ and $U^{(l)}$ are uncorrelated within each source. Then, for any source $l$ with $\Delta_l>0$, every StablePCA solution $P^*$ satisfies
\begin{equation}
 \|P^*-VV^\intercal\|_F^2
 \leq\frac{2(s_l-\min_{l'\in[L]}s_{l'})}{\Delta_l}.
 \label{eq: margin equal revision}
\end{equation}
Consequently, if $s_1=\cdots=s_L$ and $\Delta_l>0$ for at least one source $l\in[L]$, then $VV^\intercal$ is the unique solution of StablePCA in \eqref{eq: def diStablePCA}.
\end{Proposition}
When all sources have the same total variance in the shared subspace ($s_1=\cdots=s_L$), StablePCA recovers it exactly if the shared variation is stronger than the source-specific variation in just one source ($\Delta_l>0$), while the source-specific variances in the other sources can be arbitrarily large. In contrast, PooledPCA selects the directions of largest variance in the pooled data, so strong source-specific variation in some sources may dominate and fail to recover the shared the subspace. The following example illustrates this difference.

Consider three sources generated by
\begin{equation}
 X^{(l)}=e_1Z^{(l)}+e_2U^{(l)},\qquad
 U^{(l)}=\beta_lZ^{(l)}+\epsilon^{(l)},\qquad l\in[3],
 \label{eq: shared recovery 2d model revision}
\end{equation}
where $Z^{(l)}\sim N(0,1)$ and $\epsilon^{(l)}\sim N(0,\sigma_l^2)$ are independent. We fix $\beta_1=0$ and $(\sigma_1^2,\sigma_2^2,\sigma_3^2)=(0.4,1.4,1.2)$. We independently draw $\beta_2\sim\operatorname{Unif}(0.3,1.1)$, $\beta_3\sim\operatorname{Unif}(-1.1,-0.3)$, and the sample size $n_l$ uniformly from $\{20,21\ldots,200\}$ for each source $l$.

\begin{figure}[!ht]
 \centering
 \includegraphics[width=\linewidth]{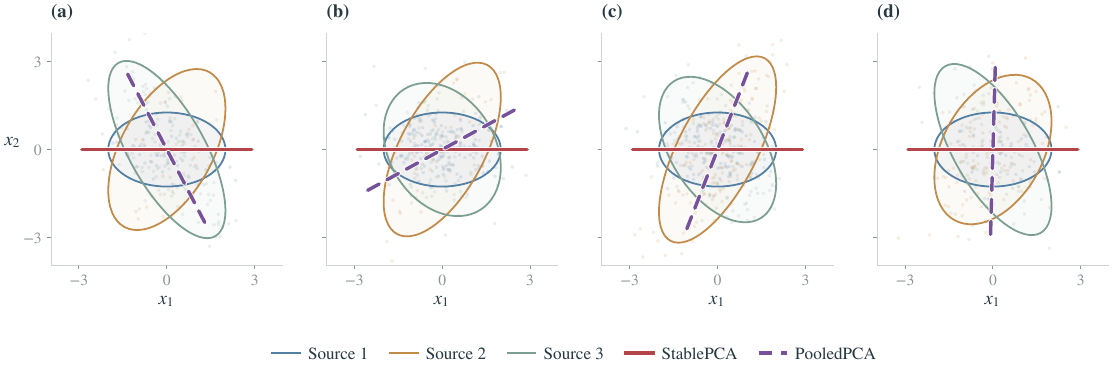}
 \caption{Comparison of subspaces yielded by StablePCA and PooledPCA in four realizations of latent factor models \eqref{eq: shared recovery 2d model revision}. Ellipses show two-standard-deviation covariance contours, and dots are sampled observations for each source.}
 \label{fig: shared recovery 2d revision}
\end{figure}

Figure~\ref{fig: shared recovery 2d revision} shows four representative realizations of \eqref{eq: shared recovery 2d model revision}. Among all four setups, StablePCA always recovers the shared subspace ${\rm span}(e_1)$, while PooledPCA yields a subspace that varies with the random slopes $\beta_l$ and sample sizes $n_l$. In Appendix~\ref{appendix: shared recovery revision illustration}, we provide another three-dimensional example to illustrate the different behaviors of StablePCA and PooledPCA.

\section{Fantope Relaxation and Algorithm for Solving StablePCA}
\label{sec: algo - efficient}
In this section, we focus on solving the StablePCA optimization problem in \eqref{eq: def diStablePCA}. {The main computational challenge stems from the interaction between the nonconvex constraint imposed by the set of rank-$k$ projection matrices $\Pc^k$, and the adversarial minimax structure of the problem.}
To address this issue, we introduce in Section~\ref{subsec: fantope} a convex Fantope relaxation of StablePCA, and in Section~\ref{sec: algo} we further develop an efficient gradient-based algorithm (Mirror-Prox) for computing the relaxed StablePCA.

\subsection{Fantope Relaxation}
\label{subsec: fantope}
 We introduce the Fantope  relaxation of the original StablePCA problem in \eqref{eq: def diStablePCA}.
The $k$-dimensional Fantope \citep{dattorro2010convex} is the convex hull of the nonconvex $\mathcal{P}^k$ defined as:
\begin{equation}
\mathcal{F}^k=\left\{M\in \RR^{d\times d}: M=M^{\intercal}, 0\preceq M\preceq {\bf I}_d, {\rm Tr}(M)=k\right\}.
\label{eq: fantope}
\end{equation}
Substituting the nonconvex constraint $P\in \Pc^k$ in the original nonconvex StablePCA formulation \eqref{eq: def diStablePCA} with the Fantope $M\in \Fc^k$ yields the following relaxed StablePCA problem:
\begin{equation}
M^*\in \argmax_{M\in \mathcal{F}^k}\min_{\omega\in \Delta^{L}} \langle {\Sigma}(\omega), M \rangle, \quad \textrm{where} \quad \Sigma(\omega) := \sum_{l=1}^L \omega_l \cdot \Sigma^{(l)}.
\label{eq: fantope MPCA}
\end{equation}
The Fantope relaxation has also been employed in other contexts; for example, in sparse PCA, where the classical PCA objective is combined with an $\ell_1$-penalty; see \cite{vu2013fantope} for more details. The relaxation in \eqref{eq: fantope MPCA} transforms the original nonconvex StablePCA into a convex optimization problem, for which we shall devise a computationally efficient algorithm that provably converges to the global optimum.

Even though we are able to solve the relaxed StablePCA problem efficiently, the solution to the relaxed problem may differ from the original StablePCA, since the optimal solution $M^*$ in \eqref{eq: fantope MPCA} is not guaranteed to be a rank-$k$ projection matrix. In Section~\ref{sec: dual-form}, we introduce a sufficient condition for when the Fantope relaxation is tight, meaning that the relaxed solution $M^*$ is also the optimal solution of the original StablePCA.
Throughout the remainder of the paper, unless otherwise specified, we refer to the original nonconvex problem in \eqref{eq: def diStablePCA} as {\bf original StablePCA}, and to its relaxed formulation \eqref{eq: fantope MPCA} simply as {\bf StablePCA}.

We now switch to the finite-sample setting, where we observe $n_l$ i.i.d. samples $\{\X{l}_i\}_{i=1}^{n_l}$ drawn from the $l$-th source distribution $\TT^{(l)}$ for $1\leq l\leq L$. By substituting the population-level $\Sigma^{(l)}$ with its empirical counterpart $\widehat{\Sigma}^{(l)} = \frac{1}{n_l}\sum_{i=1}^{n_l} \X{l}_i [\X{l}_i]^\intercal$ in \eqref{eq: def diStablePCA} and \eqref{eq: fantope MPCA}, respectively, we define the empirical version of the original StablePCA as:
\begin{equation}
    \widehat{P}\in \argmax_{P\in \Pc^k}\min_{\omega\in \Delta^L}\langle \widehat{\Sigma}(\omega), P\rangle,\quad \textrm{where}\quad \widehat{\Sigma}(\omega) := \sum_{l=1}^L \omega_l \cdot \widehat{\Sigma}^{(l)}
    \label{eq: empirical original stablepca}
\end{equation}
and the empirical StablePCA as:
\begin{equation}
    \widehat{M} \in \argmax_{M\in \Fc^k}\min_{\omega\in \Delta^L}\langle \widehat{\Sigma}(\omega), M\rangle.
    \label{eq: empirical stablepca}
\end{equation}

In the next subsection, we present a computationally efficient algorithm for solving the empirical StablePCA problem \eqref{eq: empirical stablepca} and further introduce a data-dependent certificate to assess how well the proposed algorithm solves the original StablePCA problem \eqref{eq: empirical original stablepca}. %

\subsection{Mirror-Prox Algorithm with Certificate}
\label{sec: algo}

We now introduce the algorithm for solving the empirical StablePCA problem \eqref{eq: empirical stablepca} , which we rewrite as a minimax problem:
\begin{equation}
    \widehat{M} \in \argmin_{M\in \Fc^k}\max_{\omega\in \Delta^L} {f}(M,\omega), \quad \textrm{with} \quad {f}(M,\omega) := -\langle \widehat{\Sigma}(\omega), M\rangle.
    \label{eq: finite sample}
\end{equation}
The objective function $f(M,\omega)$ is bilinear in $(M,\omega)$ and thus
defines a convex-concave minimax problem.
To solve the problem, we adopt the Mirror-Prox algorithm, which is designed for solving constrained minimax optimization problems \citep{nemirovski2004prox}; see Section 5.2 of \cite{bubeck2015convex} for a detailed discussion.

Mirror-Prox is a gradient-based algorithm that iteratively updates the variables $(M, \omega)\in \Fc^k\times \Delta^L$. To build intuition, we briefly recall standard gradient descent for minimizing an unconstrained differentiable convex function $\min_{x\in \RR^d}g(x)$, which takes the form:
\begin{equation*}
    x^{t+1} = x^t - \eta\nabla g(x^t) = \argmin_{x\in \RR^d} \left\{g(x^t)+\eta \langle \nabla g(x^t), x- x^t\rangle + \frac{1}{2}\|x - x^t\|_2^2\right\},
    \label{eq: standard gradient descent}
\end{equation*}
where $\eta>0$ is a pre-specified step size. 
In the rightmost expression, the term $g(x^t)+\langle\nabla g(x^t), x- x^t\rangle$ represents the first-order approximation of the objective at the current point $x^t$, while the distance $\frac{1}{2}\|x-x^t\|_2^2$ penalizes deviation from $x^t$. Therefore, each update seeks a point that achieves the largest reduction in the first-order approximated objective while remaining close to the current iterate.

Mirror-Prox employs mirror updates that follow the same underlying principle as standard gradient descent.
The key difference is that mirror updates replace the Euclidean distance $\frac{1}{2}\|x - x^t\|_2^2$ with Bregman divergences, which are better aligned with the geometry of constrained sets.
For completeness, we provide formal definitions and a brief overview of Bregman divergences and mirror updates in Appendix~\ref{appendix: defs}. In our setting, tailored to the constraints $M\in \Fc^k$ and $\omega\in \Delta^L$, we adopt the following Bregman divergences:
\begin{equation}
D_{\psi_1}(M,M')={\rm Tr}\left(M\left(\log M-\log M'\right)\right), \quad \textrm{and}\quad 
D_{\psi_2}(\omega, \omega')=\sum_{l=1}^{L}\omega_l \log \frac{\omega_l}{\omega'_l},
\label{eq: Breg separate}
\end{equation}
for any $M,M'\in \Fc^k$ and $\omega,\omega'\in \Delta^L$. Here, $D_{\psi_1}(M,M')$ represents the distance between two projection matrices while $D_{\psi_2}(\omega, \omega')$ represents the distance between two probability measures. These choices are standard in the literature \citep{bubeck2015convex}.

We now describe the Mirror-Prox algorithm for solving \eqref{eq: finite sample}.
We initialize the algorithm at a feasible point $(M^0,\omega^0)\in \Fc^k\times \Delta^L$, for example, $M^0=\frac{k}{d}{\bf I}_d$ and $\omega^0 = \frac{1}{L}{\bf 1}_L$. 
For each iteration $t=0,1,...,T-1$, the Mirror-Prox algorithm proceeds in two steps.
First, given the current iterate $(M^t, \omega^t)$ and the step sizes $\eta_M,\eta_\omega>0$, we compute the midpoint iterate $(M^{t+\frac{1}{2}}, \omega^{t+\frac{1}{2}})$ by performing mirror updates over the constraints $M\in \Fc^k$ and $\omega\in \Delta^L$:
\begin{equation}
\begin{aligned}
    M^{t+\frac{1}{2}} &= \argmin_{M\in \Fc^k}\left\{\eta_M\cdot  \left\langle \nabla_M f({M}^t, {\omega}^t), \;M\right\rangle + D_{\psi_1}(M, M^t)\right\},\\
    {\omega}^{t+\frac{1}{2}} &= \argmin_{\omega\in \Delta^L}\left\{-\eta_\omega\cdot \left\langle \nabla_\omega f({M}^t, {\omega}^t),\; \omega\right\rangle + D_{\psi_2}(\omega, \omega^t)\right\}.
\end{aligned}
    \label{eq: mirror middle update}
\end{equation}
For the update on $M$, the linear term 
$\eta_M \langle \nabla_M f(M^t, \omega^t), M \rangle$ 
corresponds to the first-order approximation of the objective $f(M,\omega^t)$ at the current iterate $M^t$, and the Bregman divergence $D_{\psi_1}(M, M^t)$ prevents the update from deviating excessively from $M^t$.
Similarly, the update on $\omega$ follows from a first-order approximation of $f(M^t,\omega)$ at $\omega^t$.

After obtaining the midpoint $(M^{t+\frac{1}{2}},\omega^{t+\frac{1}{2}})$, we compute the next iterate $(M^{t+1}, \omega^{t+1})$ using a similar idea as in \eqref{eq: mirror middle update} but evaluating gradients at the midpoint $(M^{t+\frac{1}{2}},\omega^{t+\frac{1}{2}})$
\begin{equation}
\begin{aligned}
    {M}^{t+1} &= \argmin_{M\in \Fc^k}\left\{\eta_M\cdot  \left\langle \nabla_M{f}({M}^{t+\frac{1}{2}}, {\omega}^{t+\frac{1}{2}}), \;M\right\rangle + D_{\psi_1}(M, M^t)\right\},\\
    {\omega}^{t+1} &= \argmin_{\omega\in \Delta^L}\left\{-\eta_\omega\cdot \left\langle \nabla_\omega {f}(M^{t+\frac{1}{2}}, {\omega}^{t+\frac{1}{2}}), \;\omega\right\rangle + D_{\psi_2}(\omega, \omega^t)\right\}.
\end{aligned}
    \label{eq: mirror final update}
\end{equation}
By evaluating the gradient at the midpoint iterate, the second step utilizes a more accurate approximation of the gradient at the subsequent iterate. Such an extra-gradient correction mitigates oscillations that commonly arise near saddle points in minimax optimization and stabilizes the trajectory of the iterates. As a result, the Mirror-Prox algorithm achieves the improved convergence rate $\mathcal{O}(T^{-1})$ for smooth convex-concave problems, in contrast to the slower rate $\mathcal{O}(T^{-1/2})$ for methods without the extra-gradient step. We refer to works \citep{mokhtari2020unified, bubeck2015convex} for a more detailed explanation.

For the mirror updates in \eqref{eq: mirror middle update} and \eqref{eq: mirror final update}, we derive the explicit closed-form expressions in the following proposition, which enable efficient updates. The proof is given in Appendix~\ref{proof of prop: closed form}.
\begin{Proposition}[Closed-form Updates]
The first step in \eqref{eq: mirror middle update} admits the following closed-form expression:
\begin{equation}
\begin{aligned}
    M^{t+\frac{1}{2}} &= \sum_{j=1}^d \min\{\exp(\lambda^t_j + \nu^t), 1\} U^{t}_j [U^{t}_j]^\intercal,\\
    \omega^{t+\frac{1}{2}}_l &= \frac{{\omega}^{t}_l \exp\left(-\eta_\omega\langle \widehat{\Sigma}^{(l)}, M^{t} \rangle\right)}{\sum_{r=1}^{L}{\omega}^{t}_r \exp\left(-\eta_\omega\langle \widehat{\Sigma}^{(r)}, M^{t} \rangle\right)},
\end{aligned}
\label{eq: omega update 1}
\end{equation}
for $1\leq l\leq L$, where $U^t{\rm Diag}(\lambda^t)(U^t)^\intercal$ is the eigen decomposition of $\log M^t+\eta_M\sum_{l=1}^L{\omega}_l^{t}\widehat{\Sigma}^{(l)}$,
and $\nu^t\in \RR$ is the unique solution to
$\sum_{j=1}^d \min\left\{\exp(\lambda_j^t + \nu), 1\right\} = k$.
Similarly, the second step in \eqref{eq: mirror final update} admits:
\begin{equation}
\begin{aligned}
    {M}^{t+1}&= \sum_{j=1}^d \min\left\{ \exp({\lambda}^{t+\frac{1}{2}}_j+\nu^{t+\frac{1}{2}}),1 \right\}U^{t+\frac12}_j[{U}^{t+\frac{1}{2}}_j]^\intercal,\\
    {\omega}^{t+1}_l &= \frac{{\omega}^{t}_l \exp\left(-\eta_\omega\langle \widehat{\Sigma}^{(l)}, M^{t+\frac{1}{2}} \rangle\right)}{\sum_{r=1}^{L}{\omega}^{t}_r \exp\left(-\eta_\omega\langle \widehat{\Sigma}^{(r)}, M^{t+\frac{1}{2}} \rangle\right)},
\end{aligned}
\label{eq: omega update 2}
\end{equation}
for $1\leq l\leq L$, where ${U}^{t+\frac{1}{2}}{\rm Diag}({\lambda}^{t+\frac{1}{2}})({U}^{t+\frac{1}{2}})^\intercal$ is the eigen decomposition of $\log M^t+\eta_M \sum_{l=1}^L\omega_l^{t+\frac{1}{2}}\widehat{\Sigma}^{(l)}$,
and $\nu^{t+\frac{1}{2}}$ is the unique solution to $\sum_{j=1}^d \min\left\{\exp(\lambda_j^{t+\frac{1}{2}} + \nu), 1\right\} = k$.
\label{prop: closed form revised}
\end{Proposition}

After completing $T$ iterations, we collect the midpoint iterates $\{(M^{t+\frac{1}{2}},\omega^{t+\frac{1}{2}})\}_{t=0}^{T-1}$ and return their average:
\begin{equation}
    \widehat{M}_T=\frac{1}{T}\sum_{t=0}^{T-1} M^{t+\frac{1}{2}},\quad \textrm{and}\quad \widehat{\omega}_T=\frac{1}{T}\sum_{t=0}^{T-1} \omega^{t+\frac{1}{2}}.
    \label{eq: averaged iterates}
\end{equation}
We use $\widehat{M}_T$ as an approximate solution to the empirical StablePCA optimization problem \eqref{eq: empirical stablepca}, and establish that the objective value attained by $\widehat{M}_T$ converges to the global optimum with increasing $T$ in later Section \ref{sec: theory}.

Since $\widehat{M}_T\in \Fc^k$ may not have rank exactly $k$, we enforce the rank constraint via a final projection step, by computing the nearest rank-$k$ projection matrix to $\widehat{M}_T$:
\begin{equation}
    \widehat{P}_T = \argmin_{P\in \Pc^k} \|P - \widehat{M}_T\|_F.
    \label{eq: proj}
\end{equation}
Thus $\widehat{P}_T=V_T V_T^\intercal$, where $V_T\in \mathcal{O}^{d\times k}$ comprises top-$k$ eigenvectors of $\widehat{M}_T$. Although $\widehat{P}_T\in \Pc^k$ satisfies the rank constraint, it does not necessarily solve the original nonconvex StablePCA problem \eqref{eq: empirical original stablepca} to global optimality. To quantify how close $\widehat{P}_T$ is to solving  the original nonconvex StablePCA problem \eqref{eq: empirical original stablepca}, we introduce a data-dependent {certificate}:
\begin{equation}
\begin{aligned}
    \tau&:=\min_{\omega\in \Delta^L}\langle\widehat\Sigma(\omega), \widehat{M}_T\rangle - \min_{\omega\in \Delta^L}\langle \widehat\Sigma(\omega), \widehat{P}_T\rangle=\min_{1\leq l\leq L} \langle \widehat{\Sigma}^{(l)},{\widehat{M}_T}\rangle - \min_{1\leq l\leq L}\langle \widehat{\Sigma}^{(l)}, {\widehat{P}_T}\rangle,
\end{aligned}
    \label{eq: certificate}
\end{equation}
where $\widehat\Sigma(\omega)$ is defined in \eqref{eq: empirical original stablepca}.
This certificate is directly computable and measures the difference  between the worst-case explained variances incurred by replacing the relaxed solution $\widehat{M}_T\in \Fc^k$ with the exact rank-$k$ $\widehat{P}_T\in \Pc^k$.

We now provide the rationale for why the certificate $\tau$ indicates whether $\widehat{P}_T$ solves the original nonconvex StablePCA problem \eqref{eq: empirical original stablepca} to global optimality. Since $\Fc^k\supseteq \Pc^k$, the objective gap between the objective value $\min_{\omega\in \Delta^L}\langle\widehat\Sigma(\omega), \widehat{P}_T\rangle$ and the global optimal value is upper bounded as follows:
\begin{equation}
\begin{aligned}
    0 \ &\ \leq \max_{P\in \Pc^k}\left[\min_{\omega\in \Delta^L}\langle \widehat\Sigma(\omega), P\rangle\right] - \min_{\omega\in \Delta^L}\langle\widehat\Sigma(\omega), \widehat{P}_T\rangle \\
    &\ \leq\ \max_{M\in \Fc^k}\left[\min_{\omega\in \Delta^L}\langle \widehat\Sigma(\omega), M\rangle\right] - \min_{\omega\in \Delta^L}\langle\widehat\Sigma(\omega), \widehat{P}_T\rangle \\
    &\ = \left\{\max_{M\in \Fc^k}\left[\min_{\omega\in \Delta^L}\langle \widehat\Sigma(\omega), M\rangle\right] - \min_{\omega\in \Delta^L}\langle\widehat\Sigma(\omega), \widehat{M}_T\rangle\right\} + \tau.
\end{aligned}
    \label{eq: upper bound for P_T}
\end{equation}

In Section~\ref{sec: theory}, we show that the first term, the objective gap of $\widehat{M}_T$ for the relaxed StablePCA problem, is of order $\mathcal{O}(T^{-1})$. Consequently, the data-dependent certificate $\tau$ controls how far $\widehat{P}_T$ is from global optimality for the original StablePCA problem.
We summarize the entire procedure in Algorithm \ref{algo: mp}.

\begin{algorithm}[!ht]
{
\DontPrintSemicolon
\SetAlgoLined
\SetNoFillComment
\LinesNotNumbered 
\caption{Mirror-Prox Algorithm for StablePCA}
\label{algo: mp}
\KwData{Empirical matrices $\{\widehat{\Sigma}^{(l)}\}_{l=1}^L$; subspace dimension $k$; step sizes $\eta_M,\eta_\omega$; iterations $T$}

Initialize $M^0 = \frac{k}{d}{\bf I}_d\in \Fc^k$ and $\omega^0=\frac{1}{L}{\bf 1}_L\in \Delta^L$;

\For{$t=0,1,\dots,T-1$}{
    First Step: compute $(M^{t+\frac{1}{2}},\omega^{t+\frac{1}{2}})$ as in \eqref{eq: omega update 1};

    Second Step: compute $(M^{t+1},\omega^{t+1})$ as in \eqref{eq: omega update 2};
}

Compute $(\widehat{M}_T$, $\widehat{\omega}_T)$ by averaging midpoints as in \eqref{eq: averaged iterates};

Compute $\widehat{P}_T$ as in \eqref{eq: proj} and the certificate $\tau$ as in \eqref{eq: certificate}.

\Return $\widehat{M}_T$, $\widehat{P}_T$, and certificate $\tau$.}
\end{algorithm}

\section{Convergence Analysis of Algorithm \ref{algo: mp}}
\label{sec: theory}
In this section, we establish the convergence of Algorithm~\ref{algo: mp}, considering both the optimization error from running the algorithm for a finite number of iterations and the statistical error arising from finite samples.
We first quantify the optimization error of Algorithm~\ref{algo: mp}, then add the statistical error to obtain a guarantee for the relaxed StablePCA \eqref{eq: fantope MPCA}, and finally use the data-dependent certificate $\tau$ to guarantee the original nonconvex StablePCA \eqref{eq: def diStablePCA}. 

We introduce additional notation used throughout the analysis. We define the maximum operator norms over sources as:
\[
\rho_{\rm max}
:= \max_{1\le l\le L}\|\Sigma^{(l)}\|_{\rm op},
\quad \textrm{and}\quad 
\hat{\rho}_{\rm max}
:= \max_{1\le l\le L}\|\widehat{\Sigma}^{(l)}\|_{\rm op}.
\]
Additionally, we define the effective rank of the source matrices \citep{koltchinskii2017concentration, vershynin2018high} as follows:
\begin{equation}
    \mathbf{r}(\Sigma^{(l)}) :=\frac{\sum_{j=1}^d \lambda_j(\Sigma^{(l)})}{\lambda_{\rm max}(\Sigma^{(l)})},\qquad \mathbf{r}_{\max} := \max_{1\leq l\leq L}\mathbf{r}(\Sigma^{(l)}).
    \label{eq: effective rank}
\end{equation}
The effective rank $\mathbf{r}(\Sigma^{(l)})$ is the ratio of the sum of the eigenvalues to the largest eigenvalue and satisfies $1\leq \mathbf{r}(\Sigma^{(l)})\leq {\rm rank}(\Sigma^{(l)})\leq d$. It is therefore small whenever the ordinary rank ${\rm rank}(\Sigma^{(l)})$ is small. Under the shared latent factor model \eqref{eq: shared model revision} of Section~\ref{subsec: interpretation revision}, $\Sigma^{(l)}$ has the ordinary rank at most $k+q_l$, where $k$ and $q_l$ are the dimensions of the shared and source-specific subspaces, respectively. Hence, $\mathbf{r}(\Sigma^{(l)})\leq k+q_l$, which can be much smaller than the ambient dimension $d$. Moreover, even when $\Sigma^{(l)}$ has full rank, its effective rank can remain small if its eigenvalues decay rapidly.

We run Algorithm~\ref{algo: mp} with the following step sizes
\begin{equation}
    \eta_M = \frac{\eta}{\log L}, \quad \eta_\omega = \frac{\eta}{k\log (d/k)}, \quad \textrm{where }\;  \eta := \frac{1}{4\hat{\rho}_{\max}}\sqrt{\frac{\log L}{k\log (d/k)}}.
    \label{eq: step sizes}
\end{equation}
The following theorem quantifies the optimization error of Algorithm~\ref{algo: mp}, following the standard analysis of Mirror-Prox for bilinear convex-concave problems \citep{nemirovski2004prox, bubeck2015convex}. Its proof is given in Appendix~\ref{proof of thm: conv rates alg}.
\begin{Theorem}[Optimization error of Mirror-Prox]
\label{thm: conv rates alg}
    With the step sizes in \eqref{eq: step sizes}, the output ${\widehat{M}_T}$ of Algorithm \ref{algo: mp} satisfies
    \begin{equation*}
0\leq \max_{M \in \Fc^k} \left[\min_{\omega \in \Delta^L}\langle \widehat{\Sigma}(\omega), M \rangle\right] - \min_{\omega \in \Delta^L} \langle \widehat{\Sigma}(\omega), {\widehat{M}_T} \rangle
\leq \frac{8\hat{\rho}_{\max}\cdot k\sqrt{k\log(d/k)\log L} }{T},
\end{equation*}
where $d$ is the ambient dimension, $k$ is the subspace dimension, $L$ is the number of data sources, and $T$ denotes the number of iterations of Algorithm~\ref{algo: mp}. 
\end{Theorem}
This theorem guarantees that $\widehat{M}_T$ attains an objective value $\min_{\omega \in \Delta^L} \langle \widehat{\Sigma}(\omega), {\widehat{M}_T} \rangle$ that is within an $\mathcal{O}(T^{-1})$ gap from the optimal value of the empirical relaxed StablePCA \eqref{eq: empirical stablepca}. Equivalently, $\widehat{M}_T$ approximately solves the empirical relaxed StablePCA, with an error of the order $\mathcal{O}(T^{-1})$. We note that the $\mathcal{O}(T^{-1})$ rate matches the lower bound for first-order methods on bilinear minimax problems with fixed dimensions \citep{ouyang2021lower}.

We next consider the finite-sample error and study how well $\widehat{M}_T$ solves the population relaxed StablePCA \eqref{eq: fantope MPCA}.
We impose the following standard sub-Gaussian assumption on the feature samples.
\begin{Condition}[Sub-Gaussian features]
\label{cond: subgauss}
For $1\leq l\leq L$, \(\{X^{(l)}_i\}_{i=1}^{n_l}\) are i.i.d. sub-Gaussian vectors, whose one-dimensional projections are sub-Gaussian relative to their $L_2$ norms, i.e., there is a constant $\sigma < \infty$ such that, for every $l\in[L]$ and every $u\in\RR^d$,
\[
\big\|\langle X_i^{(l)}, u \rangle\big\|_{\psi_2}
\;\leq\;\sigma\,\big\|\langle X_i^{(l)}, u \rangle\big\|_{L^2}
\;=\;\sigma\,\big(u^\intercal \Sigma^{(l)} u\big)^{1/2},
\]
where \(\|\cdot\|_{\psi_2}\) denotes the sub-Gaussian norm for a random variable.
\end{Condition}
The normalization by the $L_2$ norm $\|\langle X^{(l)}, u\rangle\|_{L^2}$ is the standard assumption under which the sample covariance matrix concentrates at a rate governed by the effective rank \eqref{eq: effective rank} instead of the ambient dimension $d$ \citep{koltchinskii2017concentration, vershynin2018high}.

The following theorem shows that $\widehat{M}_T$ approximately solves the relaxed StablePCA \eqref{eq: fantope MPCA}, with proof in Appendix \ref{proof of thm: new conv}. Throughout, we write $n=\min_{l} n_l$ for the smallest source sample size.
\begin{Theorem}[Convergence to the relaxed StablePCA]
\label{thm: new conv}
    Suppose Condition \ref{cond: subgauss} holds, the sample size satisfies $n\geq \mathbf{r}_{\max}$, and the step sizes are specified as in \eqref{eq: step sizes}.
    Then for any value $t \in [0, n-\mathbf{r}_{\max}]$, with probability at least $1-2Le^{-t}$, the output $\widehat{M}_T$ of Algorithm \ref{algo: mp} satisfies
    \[
    \max_{M\in\Fc^k}
    \left[\min_{\omega\in\Delta^L}
    \langle \Sigma(\omega), M\rangle\right]
    -
    \min_{\omega\in\Delta^L}
    \langle \Sigma(\omega), \widehat{M}_T\rangle
    \le
    C\rho_{\max}\cdot k\!\left(
    \sqrt{\frac{\mathbf{r}_{\max}+t}{n}}+
    \frac{\sqrt{k\log(d/k)\log L}}{T}
    \right)
    ,
    \]
    where $C>0$ denotes a constant depending only on the sub-Gaussian parameter $\sigma$ in Condition~\ref{cond: subgauss}, 
$d$ is the ambient dimension, $k$ is the subspace dimension, $\mathbf{r}_{\max}$ is the effective rank defined in \eqref{eq: effective rank} ,$L$ is the number of data sources, and $T$ denotes the number of iterations of Algorithm~\ref{algo: mp}.
\end{Theorem}
In the upper bound, the term $k\sqrt{(\mathbf{r}_{\max}+t)/n}$ is the finite-sample error, while the term $k{\sqrt{k\log (d/k)\log L}}/{T}$ is the optimization error from running a finite number of iterations. As $n$ and $T$ grow, the bound vanishes, and $\widehat{M}_T$ solves the population relaxed StablePCA to global optimality.

The finite-sample error depends on the effective rank $\mathbf{r}_{\max}$ of \eqref{eq: effective rank} rather than on the ambient dimension $d$. Under the shared latent factor model \eqref{eq: shared model revision}, $\mathbf{r}_{\max}$ is at most $k + \max_l q_l$, the number of shared programs plus the largest number of artifacts in a batch, which can be much smaller than $d$. The optimization error involves $d$ only through the logarithmic factor $\sqrt{k\log(d/k)}$, which comes from the diameter $k\log(d/k)$ of the Fantope under the Bregman divergence in \eqref{eq: Breg separate}.

We now turn to the original nonconvex StablePCA and analyze the projected output $\widehat{P}_T$ of Algorithm~\ref{algo: mp}.
Following \eqref{eq: upper bound for P_T}, the next theorem uses the data-dependent certificate $\tau$ in \eqref{eq: certificate} to quantify how well $\widehat{P}_T$ solves the original StablePCA \eqref{eq: def diStablePCA}. The proof is provided in \ref{proof of thm: global conv rates alg}.
\begin{Theorem}[Convergence to the original StablePCA]
\label{thm: global conv rates alg}
Under the conditions of Theorem~\ref{thm: new conv}, for any value $t\in [0, n-\mathbf{r}_{\max}]$, with probability at least $1-2Le^{-t}$, the output $\widehat{P}_T$ of Algorithm \ref{algo: mp} satisfies
    \begin{equation}
        \max_{P\in\Pc^k}
    \left[\min_{\omega\in\Delta^L}
    \langle \Sigma(\omega), P\rangle\right]
    -
    \min_{\omega\in\Delta^L}
    \langle \Sigma(\omega), \widehat{P}_T\rangle
    \le
    C\rho_{\max}\cdot k\!\left(
    \sqrt{\frac{\mathbf{r}_{\max}+t}{n}}
    +
    \frac{\sqrt{k\log(d/k)\log L}}{T}
    \right) + \tau,
    \label{eq: rate of original pop stablepca}
    \end{equation}
    where $C>0$ denotes a constant depending only on the sub-Gaussian parameter $\sigma$ in Condition~\ref{cond: subgauss},
$d$ is the ambient dimension, $k$ is the subspace dimension, $\mathbf{r}_{\max}$ is the effective rank defined in \eqref{eq: effective rank} ,$L$ is the number of data sources, and $T$ denotes the number of iterations of Algorithm~\ref{algo: mp}.
\end{Theorem}
As in Theorem~\ref{thm: new conv}, the bound contains both the finite-sample and the optimization errors. In addition, it contains the data-dependent certificate $\tau$, which measures the loss incurred by the projection step of Algorithm~\ref{algo: mp} and is computed from the data.

In practice, one aims to ensure that $\widehat{P}_T$ is an $\epsilon$-optimal solution that satisfies
\[
\max_{P \in \Pc^k}\left[ \min_{\omega \in \Delta^L}\left\langle {\Sigma}(\omega), P\right\rangle\right] - \min_{\omega \in \Delta^L} \left\langle {\Sigma}(\omega), \widehat{P}_T \right\rangle \leq \epsilon,
\]
for a small tolerance $\epsilon>0$. When $n$ and $T$ are large enough that the first term on the right-hand side of \eqref{eq: rate of original pop stablepca} is at most $\epsilon/2$, it suffices to compute the certificate $\tau$ defined in~\eqref{eq: certificate} and verify whether $\tau \le \epsilon/2$. 
When this holds, $\widehat{P}_T$ is guaranteed to be $\epsilon$-optimal for the original StablePCA problem. 

Empirically, the certificate $\tau$ is negligible across all simulated settings that we have explored. Table~\ref{tab:finite-certificate-projloss} reports $\tau$ of the simulated experiments in Section~\ref{sec: simus} with $T = 500$ iterations, averaged over 100 replications, across sample sizes $n\in[100, 40000]$ and ambient dimensions $d\in\{10, 20, 30\}$. Its magnitude satisfies $|\tau| < 0.003$ in every configuration. Moreover, under the eigengap condition introduced in Section~\ref{sec: dual-form}, we establish that $\tau$ vanishes to zeroas $n$ and $T$ grow; see Appendix~\ref{appendix: eigengap conv}.
\begin{table}[!htbp]
\centering
\renewcommand{\arraystretch}{1.15}  %
\setlength{\tabcolsep}{8pt}         %
\resizebox{0.9\linewidth}{!}{%
\begin{tabular}{c|ccccccccc}
\hline
\bm{$d\backslash n$} & 100 & 300 & 600 & 1200 & 2500 & 5000 & 10000 & 20000 & 40000 \\
\hline
$10$ & -0.0001 & 0.0018 & 0.0019 & 0.0024 & 0.0010 & -0.0002 & 0.0027 & 0.0004 & 0.0002 \\
$20$ & -0.0015 & -0.0008 & -0.0006 & -0.0005 & -0.0004 & -0.0003 & -0.0002 & -0.0003 & -0.0003 \\
$30$ & 0.0017 & 0.0002 & 0.0009 & -0.0014 & -0.0024 & -0.0014 & -0.0017 & -0.0028 & -0.0021 \\
\hline
\end{tabular}
}
\caption{Summary of the values of the certificate $\tau$ across samples sizes $n\in \{100,300,...,40000\}$ and ambient dimensions $d\in \{10,20,30\}$. We set the number of iterations $T=500$ in Algorithm \ref{algo: mp}. The table reports mean values averaged over 100 replications.}
\label{tab:finite-certificate-projloss}
\end{table}

\section{Tightness of Fantope Relaxation}
\label{sec: dual-form}

Since $\Pc^k\subseteq\Fc^k$, the relaxed StablePCA \eqref{eq: fantope MPCA} can have a larger optimal value than the original StablePCA \eqref{eq: def diStablePCA}. In this section, we establish conditions for the Fantope relaxation to be \emph{tight}, meaning that the two problems have the same optimal value.

To establish tightness, it suffices to show that an optimal relaxed solution $M^*$ is a rank-$k$ projection matrix, i.e., $M^*\in\Pc^k$. Such a solution is also feasible for the original problem, so the original problem can attain the relaxed optimal value as well. We next characterize the relaxed solutions and identify a condition under which $M^*\in\Pc^k$.

Note that the objective function $\langle\Sigma(\omega), M\rangle$ of \eqref{eq: fantope MPCA} is bilinear in $(M,\omega)$, and both $\Fc^k$ and $\Delta^L$ are convex and compact. By Sion's minimax theorem \citep{sion1958general}, the order of the two operators can be interchanged:
\begin{equation}
    \max_{M\in \Fc^k}\min_{\omega\in \Delta^L} \langle\Sigma(\omega), M\rangle = \min_{\omega\in \Delta^L} \max_{M\in \Fc^k}\langle \Sigma(\omega), M\rangle.
    \label{eq:minimax equality}
\end{equation}
The inner maximum on the right-hand side is explicit, and we denote its minimizers by
\begin{equation}
    \omega^* \in \argmin_{\omega\in \Delta^L} \phi(\omega),\quad \textrm{with}\quad \phi(\omega)=\max_{M\in \Fc^k}\langle \Sigma(\omega), M\rangle=\sum_{i=1}^k \lambda_i \left(\Sigma(\omega)\right),
    \label{eq: omega_star}
\end{equation}
where the last equality holds by Ky Fan's maximum principle \citep{fan1949theorem}, and $\phi(\omega)$ is a convex function as shown in Appendix~\ref{appendix: alter discussion}.

We call any such $\omega^*$ a \textbf{least favorable mixture} of the sources, because, among all mixtures, $\Sigma(\omega^*)$ is one whose top-$k$ explained variance is smallest.
Then every solution $M^*$ of the relaxed StablePCA \eqref{eq: fantope MPCA} maximizes $\langle\Sigma(\omega^*), M\rangle$ over $\Fc^k$, for \emph{every} least favorable mixture $\omega^*$. If $\Sigma(\omega^*)$ has a gap between its $k$-th and $(k+1)$-th eigenvalues, this maximization problem has a unique solution, which is the projection matrix onto the top-$k$ eigenspace of $\Sigma(\omega^*)$. Consequently, every relaxed solution $M^*$ must be a rank-$k$ projection matrix, suggesting that the relaxed StablePCA attains the same optimal value as the original StablePCA.

Theorem~\ref{thm: fantope tight} below formalizes this argument. We first state the eigengap condition.

\begin{Condition}[Eigengap at a least favorable mixture]
\label{cond: eigengap}
There exists a least favorable mixture $\omega^*\in\argmin_{\omega\in\Delta^L}\phi(\omega)$ in \eqref{eq: omega_star} such that
\[
\delta^*:=\lambda_k\big(\Sigma(\omega^*)\big) - \lambda_{k+1}\big(\Sigma(\omega^*)\big) > 0.
\]
\end{Condition}

We give two clarifications in order. 
First, the least favorable mixture $\omega^*$ in \eqref{eq: omega_star} need not be unique, since the convex function $\phi$ is not necessarily strictly convex. Second, when there are multiple least favorable mixtures, Condition~\ref{cond: eigengap} only requires the eigengap to hold at one of them, rather than all of them.
The following proposition lists concrete situations in which the condition holds, formulated directly in terms of the source matrices $\{\Sigma^{(l)}\}_{l=1}^L$. The proof is given in Appendix~\ref{proof of prop: instances}

\begin{Proposition}[Instances of Condition~\ref{cond: eigengap}]
\label{cor: tightness sufficient}
Condition~\ref{cond: eigengap} holds in each of the following cases.
\begin{enumerate}[label=(\alph*)]
    \item \textbf{Shared latent structure.} Under the shared latent factor model of Proposition~\ref{prop: shared recovery revision}, suppose all sources have the same total variance in the shared subspace, $s_1=\cdots=s_L$, and $\Delta_l>0$ for at least one source $l\in[L]$. For any such source, $\omega^*=e_l$ is a least favorable mixture with eigengap $\delta^*=\Delta_l>0$.
    \item \textbf{A dominated source.} Some source $l_0$ satisfies $\lambda_k(\Sigma^{(l_0)}) > \lambda_{k+1}(\Sigma^{(l_0)})$, and the projection $P_{l_0}$ onto its top-$k$ eigenspace satisfies $\langle \Sigma^{(l)}, P_{l_0}\rangle\geq\langle \Sigma^{(l_0)}, P_{l_0}\rangle$ for all $l$. This holds, for instance, if $\Sigma^{(l)}\succeq\Sigma^{(l_0)}$ for all $l$. Then $\omega^* = e_{l_0}$ with eigengap $\delta^* = \lambda_k(\Sigma^{(l_0)}) - \lambda_{k+1}(\Sigma^{(l_0)})>0$.
    \item \textbf{Common leading eigenspace.} All sources share the same top-$k$ eigenspace and the same explained variance. That is, there is a rank-$k$ projection $P$ such that, for every $l$, $P$ is the projection onto the top-$k$ eigenspace of $\Sigma^{(l)}$ with $\lambda_k(\Sigma^{(l)}) > \lambda_{k+1}(\Sigma^{(l)})$, and $\langle\Sigma^{(l)}, P\rangle \equiv c$ for all $l$. Then every $\omega\in\Delta^L$ is a least favorable mixture, with eigengap $\delta^*\geq\min_l\{\lambda_k(\Sigma^{(l)}) - \lambda_{k+1}(\Sigma^{(l)})\}>0$.
\end{enumerate}
\end{Proposition}

We now state the tightness result under Condition~\ref{cond: eigengap}. The proof is given in Appendix~\ref{proof of thm: fantope tight}.

\begin{Theorem}[Tightness of Relaxation]
\label{thm: fantope tight}
Suppose Condition~\ref{cond: eigengap} holds. Then there exists a unique $P^*\in\Pc^k$ such that
\[
\argmax_{M\in\Fc^k}\min_{\omega\in\Delta^L}\langle\Sigma(\omega),M\rangle
=\argmax_{P\in\Pc^k}\min_{\omega\in\Delta^L}\langle\Sigma(\omega),P\rangle=\{P^*\},
\]
and the relaxed and original StablePCA problems have the same optimal value:
\begin{equation}
\max_{M\in\Fc^k}\min_{\omega\in\Delta^L}\langle\Sigma(\omega),M\rangle
=\max_{P\in\Pc^k}\min_{\omega\in\Delta^L}\langle\Sigma(\omega),P\rangle.
\label{eq: tight values}
\end{equation}
\end{Theorem}

Since the relaxation is tight under the eigengap condition, Algorithm~\ref{algo: mp}, which solves the relaxed problem to global optimality, also solves the original nonconvex problem to global optimality.

Moreover, in Appendix~\ref{appendix: eigengap conv}, we show that under the eigengap Condition \ref{cond: eigengap}, both the relaxed output $\widehat{M}_T$ and the projected output $\widehat{P}_T$ of Algorithm~\ref{algo: mp} converge to the solution $P^*$ itself, not merely in objective value as in Section \ref{sec: theory}, and that the data-dependent certificate $\tau$ vanishes as $n$ and $T$ grow. Treating the subspace dimension $k$, the number of sources $L$, the operator norm $\rho_{\max}$, and the eigengap $\delta^*$ as constants, with probability at least $1 - 4Le^{-t}$, 
\begin{equation}
    \max\left\{\|\widehat{M}_T - P^*\|_F^2, \ \|\widehat{P}_T - P^*\|_F^2 \right\}\ \lesssim \ \sqrt{\frac{t}{n}} + \frac{\mathbf{r}_{\max}+t}{n} + \frac{\sqrt{\log d}}{T},
    \label{eq: conv rate of obj original stablepca}
\end{equation}
where the effective rank $\mathbf{r}_{\max}$ enters at the order of $n^{-1}$ and the ambient dimension $d$ enters only through a logarithmic factor. Compared with Theorem~\ref{thm: global conv rates alg}, the data-dependent certificate $\tau$ is replaced there by an explicit term that vanishes with $n$ and $T$.

\section{Exploration of Alternative Robust Multi-source PCA}
\label{subsec: alternative}
While the proposed StablePCA is formulated based on the explained variance, as defined in \eqref{eq: def diStablePCA}, this section further explores alternative distributionally robust formulations of multi-source PCA based on other loss functions. Different choices of the loss function result in different learned low-rank representations. The impact of loss function selection  has also been observed in multi-source regression problems \citep{wang2023distributionally}. 

\medskip
\noindent \textbf{Formulations of Alternative Robust Multi-source PCA.} The first alternative approach is designed to minimize the worst-case squared reconstruction error over the uncertainty set $\Cc$ defined in \eqref{eq: uncertainty class}:
\begin{equation}
\label{eq: squaredPCA}
    P^{\rm sq} \in \argmin_{P\in\Pc^k}\max_{\QQ\in \Cc}\EE_{\QQ}\|X - PX\|_2^2 = \argmin_{P\in\Pc^k}\max_{l\in [L]}\EE\|X^{(l)} - PX^{(l)}\|_2^2,
\end{equation}
where the equality follows because the inner maximization over $\Cc$ is attained at one of the extreme points, corresponding to the worst-case source.
Since it extends the squared reconstruction-error formulation of classical PCA in \eqref{eq: standardPCA}, we refer to it as SquaredPCA. 

Note that $\EE\|X^{(l)} - PX^{(l)}\|_2^2 = \langle \Sigma^{(l)}, {\bf I}-P\rangle$, which corresponds to the \emph{unexplained variance} of $X^{(l)}$ by $P$, where $\Sigma^{(l)}=\EE[X^{(l)}X^{(l)\intercal}]$. Therefore, SquaredPCA in \eqref{eq: squaredPCA} can be equivalently interpreted as minimizing the worst-case unexplained variance across sources, in contrast to the proposed StablePCA, which maximizes the worst-case explained variance. These two formulations generally lead to different solutions under the multi-source regime, although they are equivalent in the single source setting, as shown in Section \ref{subsec: classical pca}.

Moreover,
one may instead consider a regret-based objective that measures the excess reconstruction error relative to the best subspace for each mixture of distributions. Given a distribution $\QQ$ and a projection matrix $P\in \Pc^k$, the corresponding regret is defined as:
\[
{\rm Regret}_\QQ(P):= \EE_{\QQ}\|X - PX\|_2^2 - \min_{P'\in \Pc^k} \EE_{\QQ}\|X - P'X\|_2^2,
\]
where $\min_{P'\in \Pc^k} \EE_{\QQ}\|X - P'X\|_2^2$ denotes the smallest reconstruction error for the distribution $\QQ$.
We then minimize the worst-case regret across the uncertainty set $\Cc$:
\begin{equation}
    P^{\rm fair} \in \argmin_{P\in \Pc^k}\max_{\QQ\in \Cc} {\rm Regret}_\QQ(P).
    \label{eq: fairPCA}
\end{equation}
We refer to this approach as FairPCA, as it coincides with the worst-case regret formulation studied in \citet{samadi2018price}. In their work, FairPCA was defined as $P^{\rm fair} \in \argmin_{P\in \mathcal{P}^k}\max_{l\in [L]}{\rm Regret}_{\mathbb{T}^{(l)}}(P)$, which minimizes the maximum regret across individual source distributions. 
Our formulation instead takes the worst-case regret over the entire uncertainty class $\mathcal{C}$, consisting of all mixtures of the source distributions $\{\mathbb{T}^{(l)}\}_{l=1}^L$, thereby providing a distributionally robust interpretation of FairPCA. We show in Appendix~\ref{subsec: dual appendix - connection} that these two formulations are in fact equivalent.

\begin{figure}[!ht]
    \centering
    \includegraphics[width=\linewidth]{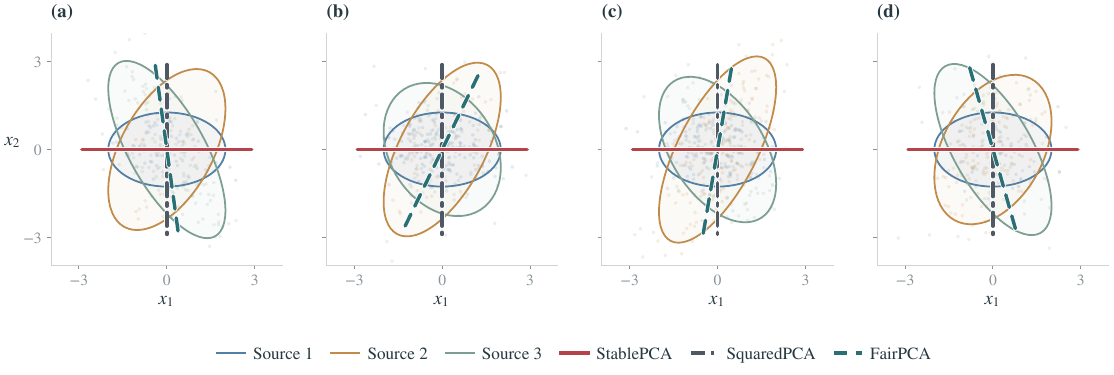}
    \caption{Comparison of subspaces yielded by  StablePCA, SquaredPCA, and FairPCA subspaces for the same setups as Figure~\ref{fig: shared recovery 2d revision}. Ellipses show two-standard-deviation covariance contours, and dots are sampled observations, for each source. }
    \label{fig: alt 2d}
\end{figure}

Figure~\ref{fig: alt 2d} compares the three formulations in the same setups as Figure~\ref{fig: shared recovery 2d revision}. StablePCA consistently recovers the shared subspace ${\rm span}(e_1)$. SquaredPCA instead selects ${\rm span}(e_2)$, while FairPCA returns a tilted subspace. Appendix~\ref{appendix: shared recovery revision illustration} provides another three-dimensional example to illustrate their differences.

We emphasize that the three methods differ only in the criterion they optimize, and each criterion responds to a different feature of the data. SquaredPCA minimizes the worst-case unexplained variance. FairPCA minimizes the worst-case regret relative to the best subspace for each source. StablePCA maximizes the worst-case explained variance, so it seeks directions along which every source carries variance, that is, the variance shared by all sources. The explained-variance criterion is therefore suited to recovering shared structure, as illustrated in Section~\ref{subsec: interpretation revision}.
Recently, \citet{fries2026worst} also studied the worst-case explained-variance, reconstruction-error, and regret formulations of multi-source PCA, and characterized when these three objectives lead to the same subspace.

\medskip
\noindent \textbf{Mirror-Prox Algorithm  for SquaredPCA and FairPCA.}
We now generalize Algorithm \ref{algo: mp} to solve SquaredPCA and FairPCA. We then compare this Mirror-Prox algorithm with the semidefinite programming approach adopted in prior work \citep{samadi2018price} for solving FairPCA. We begin by presenting reformulations of SquaredPCA and FairPCA that parallel the StablePCA formulation in \eqref{eq: def diStablePCA}. The proof is given in Appendix~\ref{proof of prop: alternative reformulation}. 
\begin{Proposition}[Reformulations of SquaredPCA and FairPCA]
\label{prop: alternative reformulation}
    Let $\Sigma^{(l)} = \EE[X^{(l)}X^{(l)\intercal}]$ have eigenvalues $\lambda_1^{(l)}\geq \cdots\geq \lambda_d^{(l)}$ for each source $1\leq l\leq L$. 
    Then SquaredPCA defined in \eqref{eq: squaredPCA} and FairPCA defined in \eqref{eq: fairPCA} admit the following equivalent reformulations:
    \[
    P^{\rm sq}\in \argmax_{P\in \Pc^k} \min_{\omega\in \Delta^L}\sum_{l=1}^L\omega_l\cdot \left\langle \Sigma^{(l)}-\frac{1}{k}\sum_{i=1}^{\textcolor{niblue}{d}} \lambda_i^{(l)}{\bf I}_d, \; P  \right\rangle,
    \]
    \[
    P^{\rm fair} \in \argmax_{P\in \Pc^k} \min_{\omega\in \Delta^L}\sum_{l=1}^L\omega_l\cdot \left\langle \Sigma^{(l)}-\frac{1}{k}\sum_{i=1}^{\textcolor{nired}{k}} \lambda_i^{(l)}{\bf I}_d, \; P  \right\rangle.
    \]
\end{Proposition}
Together with the StablePCA formulation in \eqref{eq: def diStablePCA}, this result shows that the three robust multi-source PCA methods differ only in the constant matrix subtracted from each $\Sigma^{(l)}$ within the objective. Specifically, SquaredPCA denoted as $P^{\rm sq}$ subtracts $\frac{1}{k}\sum_{i=1}^{\textcolor{niblue}{d}} \lambda_i^{(l)}{\bf I}_d$, and FairPCA denoted as $P^{\rm fair}$ subtracts $\frac{1}{k}\sum_{i=1}^{\textcolor{nired}{k}} \lambda_i^{(l)}{\bf I}_d$, while the proposed StablePCA denoted as $P^*$ subtracts no term at all.

Based on this reformulation, SquaredPCA and FairPCA can be solved using the same algorithmic framework developed in Section~\ref{sec: algo}. In particular, one may first apply the Fantope relaxation by replacing the nonconvex constraint $P\in \Pc^k$ with its convex hull $M\in \Fc^k$, and then solve the resulting minimax problem using the Mirror-Prox Algorithm~\ref{algo: mp}. Due to space limitations, we provide the extended algorithm in Appendix~\ref{subsec: dual appendix - fantope}.

We remark on the computational complexity of the extended algorithm. The extended Mirror-Prox algorithm solves the SquaredPCA or FairPCA iteratively, where each iteration conducts an eigen-decomposition with a cost $\mathcal{O}(d^3)$, leading to a total runtime of $\mathcal{O}(d^3 T)$ over $T$ iterations. 
In contrast, the original FairPCA work \citep{samadi2018price} adopts semidefinite programming (SDP) with a runtime on the order of $\mathcal{O}(d^{6.5})$, which is computationally prohibitive in high-dimensional settings.

In addition, we compare the computational costs of different optimization methods using an experiment with varying ambient dimensions $d\in\{30,50,70,100,200,300\}$. The exact experimental setup is provided in Appendix \ref{appendix: exp setups}. We solve FairPCA using the extended Mirror-Prox algorithm with $T=500$ iterations, and we also implement the SDP-based method following the procedure described in \citet{samadi2018price}. All experiments are repeated 100 times, and we report the average runtime in seconds, together with standard deviations. For ease of comparison, we report the runtime ratio of the SDP method divided by Mirror-Prox.

The results in Table~\ref{tab:compare-sdp} show that the SDP-based approach is competitive only in small ambient dimensions (e.g., $d=30$), since its runtime grows rapidly with the ambient dimension, making it impractical in high-dimensional settings. For example, when $d=300$, our proposal is approximately $40$ times faster than the SDP-based method.

\begin{table}[!htbp]
\centering
\renewcommand{\arraystretch}{1.15}  %
\setlength{\tabcolsep}{8pt}         %
\resizebox{0.95\linewidth}{!}{%
\begin{tabular}{c|cccccc}
\hline
\textbf{Method} & $d=30$ & $d=50$ & $d=70$ & $d=100$ & $d=200$ & $d=300$ \\
\hline
MP & 0.263±0.023 & 0.324±0.035 & 0.511± 0.035 & 0.706±0.067 & 1.86±0.04 & 4.49±0.07 \\
SDP & 0.057±0.007 & 0.281±0.135 & 0.519±0.057 & 9.99±5.68 & 42.96±22.66 & 173.6±70.2 \\
SDP/MP & 0.22 & 0.89 & 1.02 & 14.25 & 23.07  & 38.79 \\
\hline
\end{tabular}
}
\caption{Comparison of the proposed Mirror-Prox (MP) algorithm and the SDP-based method \citep{samadi2018price} for solving FairPCA.
We report the average runtime (in seconds) and standard deviations for each method in 100 independent replications, across varying ambient dimensions $d\in \{30,50,70,100,200,300\}$. The ratio SDP/MP is defined as the averaged runtime of the SDP method divided by that of Mirror-Prox.}
\label{tab:compare-sdp}
\end{table}

\section{Simulations}
\label{sec: simus}
Throughout this section, we investigate the empirical performance of {StablePCA} through simulated experiments, by demonstrating that {StablePCA} effectively captures the shared latent structure in multi-source data and generalizes beyond the observed sources.
Additional experiments, regarding the finite-sample convergence of Algorithm~\ref{algo: mp}, are provided in Appendix~\ref{appendix: add simus}.

We generate $L$ independent sources from the shared latent factor model \eqref{eq: batch model revision}--\eqref{eq: orthogonality revision} of Section~\ref{subsec: interpretation revision}, with $k = 3$ shared programs, $q_l = 5$ artifacts in every source, and isotropic measurement noise. First, we draw an orthonormal matrix $V\in\mathcal{O}^{d\times 3}$ of shared programs, common to all sources. Next, for each source $l\in[L]$, we independently draw an orthonormal matrix $\Gamma^{(l)}\in\mathcal{O}^{d\times 5}$ of artifact directions orthogonal to $V$, so that $\Gamma^{(l)\intercal}V = 0$ as in \eqref{eq: orthogonality revision} and the artifact directions differ across sources. For each source $l\in[L]$ and each sample $i\in[n]$, the data are generated as
\begin{equation}
\begin{aligned}
X^{(l)}_i &= V Z^{(l)}_i + \Gamma^{(l)}U^{(l)}_i + \varepsilon^{(l)}_i,\\
Z^{(l)}_i&\sim\mathcal{N}(0_3, {\bf I}_3),\quad
U^{(l)}_i\sim\mathcal{N}\big(0_5, [\alpha^{(l)}]^2{\bf I}_5\big),\quad
\varepsilon^{(l)}_i&\sim\mathcal{N}\big(0_d, \tfrac14{\bf I}_d\big),
\end{aligned}
\label{eq: simus}
\end{equation}
drawn independently of each other, where $Z^{(l)}_i$ records the activities of the shared programs, $U^{(l)}_i$ the magnitudes of the artifacts, $\varepsilon^{(l)}_i$ is the measurement noise, and $\alpha^{(l)}\stackrel{i.i.d.}{\sim}{\rm Unif}(0.2, 3)$ controls the strength of the artifacts in source $l$. 
Unless otherwise specified, we set $d = 40$ and generate $n = 2000$ samples for each source.

\begin{figure}[!ht]
    \centering
    \includegraphics[width=\linewidth]{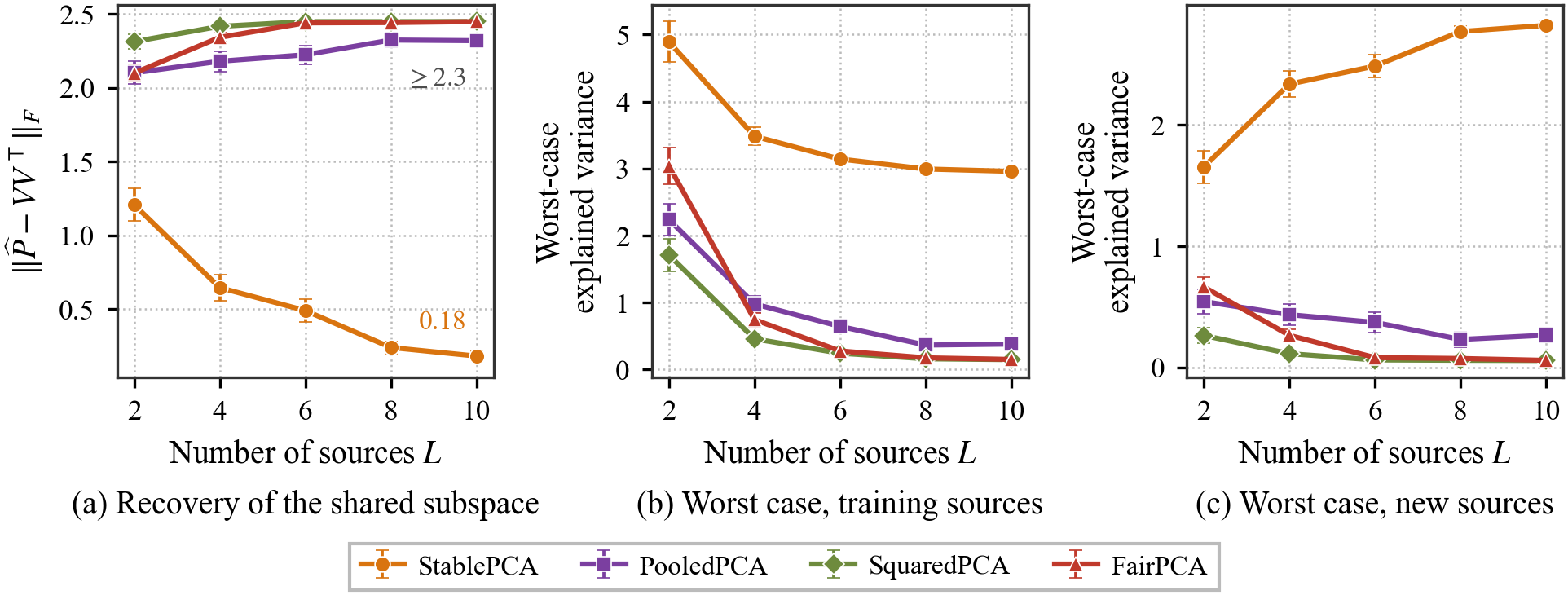}
    \caption{Comparison of {StablePCA}, {PooledPCA}, {SquaredPCA}, and {FairPCA} across different numbers of source domains $L\in\{2,4,6,8,10\}$.
    (a) Recovery error $\|\widehat{P}-VV^\intercal\|_F$ of the shared subspace.
    (b) Worst-case explained variance among the training (in-distribution) sources.
    (c) Worst-case explained variance among new (out-of-distribution) sources.
    Results are averaged over 100 replications, with error bars representing one standard error.}
    \label{fig:worst_case}
\end{figure}

We evaluate the generalization performance of {StablePCA}, implemented via Algorithm~\ref{algo: mp}, and compare it with alternative multi-source PCA methods as the number of sources varies over $L\in\{2,4,6,8,10\}$.
Competing approaches include SquaredPCA, {FairPCA}, and {PooledPCA}.
All methods are fitted with the subspace dimension $k=3$, while the results for other choices of $k\in \{5,10\}$ are provided in Appendix \ref{appendix: simus}.
After training on the observed multi-source data, we denote the estimated projection matrix by $\widehat{P}$.
The results, averaged over 100 independent replications, are summarized in Figure~\ref{fig:worst_case}. %

In panel (a) of Figure~\ref{fig:worst_case}, we first assess whether the learned $\widehat{P}$ recovers the shared subspace ${\rm col}(V)$, evaluated via the metric $\|\widehat{P}-VV^\intercal\|_F$, named as ``Shared Subspace Recovery Error''.  
Smaller values indicate closer alignment between the learned subspace of $\widehat{P}$ and the shared subspace ${\rm col}(V)$. We find that {StablePCA} achieves progressively smaller error as the number of sources $L$ increases, indicating improved recovery of the shared subspace with more source domains. 
In contrast, other competing approaches have substantially larger $\|\widehat{P}-VV^\intercal\|_F$ than {StablePCA}. For example, when $L=10$, the averaged recovery error of StablePCA is approximately $0.18$, whereas the corresponding errors for the other methods exceed $2.3$.

We then compare the methods in terms of the worst-case explained variance, evaluated both in-distribution and out-of-distribution. Larger values indicate better generalization performance.
Specifically, we consider the following two metrics:
\begin{itemize}
\setlength\itemsep{1pt}\setlength\parskip{0pt}
    \item \emph{In-distribution}:
    We report $\min_{l\in [L]}\langle\widehat{\Sigma}^{(l)}, \widehat{P}\rangle$, the worst-case explained variance over the $L$ training sources.
    \item \emph{Out-of-distribution}:
    We generate $L_{\rm out}=100$ new test distributions from \eqref{eq: simus}, each sharing the same programs $V$ but with independently drawn artifact directions $\Gamma^{(l')}$ and artifact strengths $\alpha^{(l')}$.
    The out-of-distribution performance is measured by $\min_{l'\in [L_{\rm out}]} \langle\widehat{\Sigma}^{(l')}, \widehat{P}\rangle$.
\end{itemize}
As shown in panels (b) and (c) of Figure~\ref{fig:worst_case}, {StablePCA} consistently achieves the largest worst-case explained variance under both in-distribution and out-of-distribution evaluations, outperforming the competing methods. This observation is consistent with the formulation of StablePCA, which explicitly maximizes worst-case explained variance.

\section{Real Application}
\label{sec: real}

We evaluate the performance of StablePCA on a real single-cell RNA-sequencing (scRNA-seq) dataset. Our analysis uses the human bone marrow dataset from \citet{luecken2021sandbox} containing gene expression (RNA) of cells from 12 experimental batches, across which variation persists due to differences in sequencing depth, experimental protocols, and laboratory-specific effects. Our goal is to demonstrate that StablePCA learns a low-rank transformation that represents the high-dimensional expression data well across batches, including batches unseen at training, while suppressing batch-specific variation.

We preprocess the RNA expression matrix using standard filtering procedures: genes expressed in fewer than three cells and cells with fewer than 200 detected genes are removed. Each cell is normalized to a fixed total count ($10^4$), followed by log-transformation. Since the raw RNA dimensionality exceeds ten thousand genes and would cause heavy computational burdens, we reduce the initial feature space by selecting the top 1000 highly variable genes using Scanpy \citep{wolf2018scanpy}, a commonly used pipeline.

After preprocessing the data and reducing the original features to $d=1000$ highly variable genes, we apply {StablePCA} with a subspace dimension of $k=50$. This choice provides a moderate dimension suitable for downstream visualization methods such as t-SNE \citep{maaten2008visualizing} and UMAP \citep{healy2024uniform}. Results for additional choices of $k\in \{100,150\}$ are reported in Appendix~\ref{appendix: real data}, and exhibit  similar patterns.

To assess generalization performance, we compare {StablePCA} against three alternative methods: 
{SquaredPCA}, {FairPCA}, and {PooledPCA}. %
We randomly select 8 of the 12 batches as training sources and fit all four methods to these batches. The learned projection matrix $\widehat{P}$ is then evaluated on the remaining four held-out batches. For each fitted projection matrix, we compute the worst-case explained variance over the hold-out samples from the training batches (``in-dist'') and over the test batches (``out-dist''):
\[
\min_{l\in \textrm{Training Batches}}{\langle \widehat{\Sigma}^{(l)}, \widehat{P}\rangle}, \quad \textrm{and}\quad \min_{l'\in \textrm{Test Batches}}{\langle \widehat{\Sigma}^{(l')}, \widehat{P}\rangle}.
\]
A higher worst-case explained variance indicates that the corresponding projection matrix has better generalization performance. We repeat this over 200 random train-test splits.

\begin{figure}[!ht]
    \centering
    \includegraphics[width=0.9\linewidth]{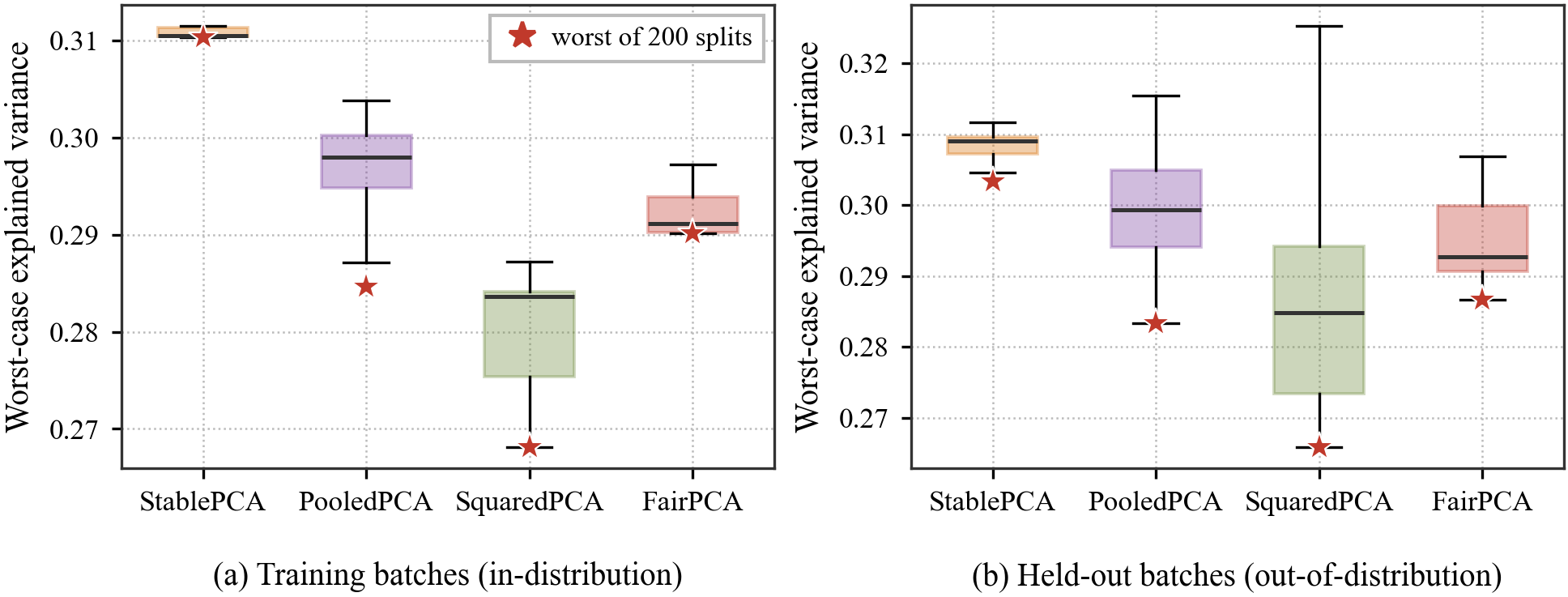}
    \caption{Worst-case explained variance of methods on the single-cell RNA dataset. (a) Worst-case explained variance among the training batches (in-distribution). 
    (b) Worst-case explained variance among the held-out batches (out-of-distribution). 
    Each boxplot summarizes the results across 200 random train-test splits, and the red stars mark the lowest worst-case explained variance observed among the 200 splits.}
    \label{fig:singlecell}
\end{figure}

Figure \ref{fig:singlecell} summarizes the results across 200 random train-test splits, where we normalize the worst-case explained variance by dividing by the maximum scale $\max_{l\in [12]} \|\widehat{\Sigma}^{(l)}\|_{\rm op}$ among 12 experimental batches. Overall, {StablePCA} consistently achieves the highest worst-case explained variance on both training batches (panel (a)) and test batches (panel (b)), indicating that its extracted representations generalize not only to the training batches but also remain stable to unseen batches.

We further examine the performance under the most challenging configurations among the 200 train-test splits. The red stars in Figure \ref{fig:singlecell} represent the lowest worst-case explained variances across all 200 splits. In particular, on the held-out batches (panel (b)), StablePCA achieves approximately 7.0\% higher worst-case explained variance than PooledPCA, 14.1\% higher than SquaredPCA, and 5.8\% higher than FairPCA. These results suggest that StablePCA maintains superior performance under the most challenging settings.

Next, we perform downstream visualizations based on the low-dimensional representations extracted by {StablePCA}. For multi-batch single-cell data, a desirable visualization on their pooled dataset should separate cells by biological cell types (not used for training) but not by the batch identities. In other words, the representations should preserve biological structure while reducing batch effects.
We apply t-SNE \citep{maaten2008visualizing} and UMAP \citep{healy2024uniform} to the representations learned by StablePCA with $k=50$, while the results for other $k$'s are deferred to Appendix~\ref{appendix: real data}.
For clarity, we focus on four major immune cell types: B cells, NK cells, Monocytes, and T cells.
Other annotated cell populations are excluded either because they contain very small numbers of cells or correspond
to transitional, progenitor, or rare cell states (e.g., erythroid precursors, cycling progenitors, or dendritic
subtypes).

\begin{figure}[!ht]
    \centering
    \includegraphics[width=0.8\linewidth]{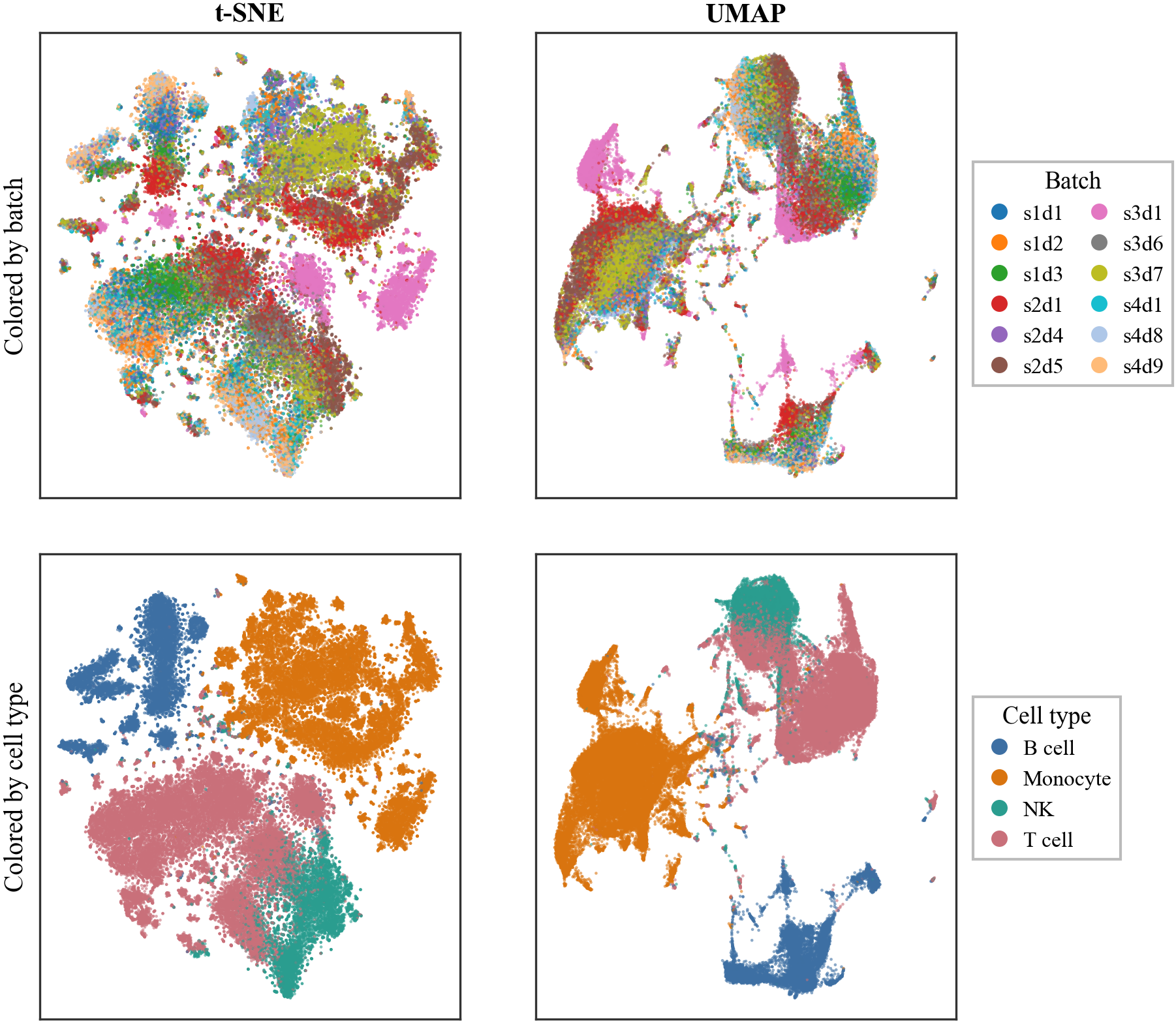}
    \caption{t-SNE and UMAP visualizations of the StablePCA embedding with subspace dimension $k=50$ on the single-cell RNA dataset.
    Top row: cells colored by 12 experimental batches.
    Bottom row: cells colored by cell types (B cell, Monocyte, NK cell, and T cell).}
    \label{fig:singlecell_tsne_umap}
\end{figure}

Figure~\ref{fig:singlecell_tsne_umap} shows the sample points visualized by t-SNE and UMAP, colored by batch identity and cell type.
When colored by batch (top row), cells from 12 different experimental batches are well mixed, indicating that the representations learned by {StablePCA} effectively suppress batch effects.
When colored by cell type (bottom row), the four major cell type samples form coherent clusters for both t-SNE and UMAP.
We note the partial overlap between NK cells and T cells, which is expected, since NK cells and cytotoxic T cells share transcriptional programs and are difficult to separate perfectly using RNA expression alone. The other major cell types are well separated. In conclusion, these results demonstrate that {StablePCA} effectively removes batch-specific variations, while preserving biologically meaningful structures to distinguish the major cell types.

\section{Conclusions and Discussions}
This work proposes StablePCA, a distributionally robust framework for extracting stable low-dimensional representations from multi-source feature data. By maximizing the worst-case explained variance over all mixtures of the sources, StablePCA recovers the structure shared by the sources. To address the nonconvex rank constraint, we reformulate the problem via a convex Fantope relaxation and develop a computationally efficient Mirror-Prox algorithm with closed-form updates. Since the relaxed problem may differ from the original one, we provide a data-dependent certificate to assess how well the algorithm solves the original nonconvex problem, and we show that the relaxation is tight under an eigengap condition.

As an extension, in the high-dimensional setting, one may consider a sparse StablePCA by introducing an $\ell_1$-type regularization, inspired by the sparse PCA formulation \citep{vu2013fantope}. 
Developing efficient algorithms and establishing corresponding theoretical guarantees for this sparse variant remain interesting directions for future research.

The proposed framework can be adapted to a stable Canonical Correlation Analysis (CCA) setting, building on the classical CCA formulation \citep{hotelling1992relations}, where the goal is to seek low-dimensional representations of two feature sets whose correlations remain stable across multiple sources. As in StablePCA, the presence of rank constraints leads to a nonconvex minimax formulation. A natural extension of our approach is to design a computationally efficient Mirror-Prox algorithm for a convex relaxation of the stable CCA minimax problem. We leave the detailed development of this extension to future work.

\clearpage
\bibliography{ref.bib}

\clearpage
\appendix

\addcontentsline{toc}{section}{Appendix} %
\begin{center}
    \textbf{\large Supplements to ``StablePCA''.}
\end{center}
\vspace{0.5em}
We provide a detailed Table of Contents for the Appendix below to facilitate navigation of the supplementary materials.

\setcounter{tocdepth}{2}  %
\startcontents[appendix]
\printcontents[appendix]{l}{1}{\setcounter{tocdepth}{2}}

\section{Additional Theoretical Results}
\label{appendix: additional theoretical results}

\subsection{Incorporation of Target Information to StablePCA}
\label{subsec: prior}
    In Section \ref{subsec: stablepca}, we use $\mathcal{C}$ to account for the uncertainty about the unknown target distribution, which comprises all mixtures of the source distributions.
    In practice, one may not need robustness against all mixtures of source distributions but may wish to perform well on a specific target distribution. Such a preference can be incorporated into the StablePCA formulation. For example, suppose domain expertise suggests that the target distribution closely resembles a particular mixture of the sources with a prior weight vector $\omega^{\rm prior}\in \Delta^L$. Alternatively, one may have access to a small amount of target data, which can be used to form a crude estimate of the mixture weights.
    In either case, rather than optimizing over the entire simplex $\Delta^L$, we may restrict the mixture weights to lie in a neighborhood of $\omega^{\rm prior}$, denoted as $\mathcal{H} = \{\omega\in \Delta^L\mid\|\omega - \omega^{\rm prior}\|_2\leq \rho\}$, where $\rho>0$ encodes the level of confidence in the prior or in the estimated weights. We then refine the uncertainty set to
$
    \Cc_\mathcal{H}:= \left\{\QQ:\; \QQ= \sum_{l=1}^L \omega_l \cdot \TTl{l} \; \text{with}\; \omega\in \mathcal{H} \subseteq\Delta^L\right\}.
$
We may extend the definition in \eqref{eq: stablepca-original} by incorporating prior information as follows:
\begin{equation*}
    \argmax_{P\in \Pc^k} \min_{\QQ\in \Cc_\mathcal{H}}\EE_{X\sim \QQ}(\|X\|_2^2-\|X-PX\|_2^2) = \argmax_{P\in \Pc^k} \min_{\omega\in \mathcal{H}} \sum_{l=1}^L \omega_l \cdot \langle \Sigma^{(l)}, P\rangle.
\end{equation*}

\subsection{Computing the Least Favorable Mixture}
\label{appendix: alter discussion}
This subsection supplements Section~\ref{sec: dual-form}: it establishes the convexity of the function $\phi$ in \eqref{eq: omega_star}, gives an explicit subgradient, and describes the subgradient method with entropic mirror updates that computes a least favorable mixture $\omega^*$.

\subsubsection{Properties of the Function $\phi(\omega)$}
\label{appendix: properties of phi}
The following lemma establishes that $\phi(\omega)$ is convex. The proof is provided in Appendix \ref{proof of lemma: eigenvalue function}.
\begin{Lemma}[Convexity of $\phi$]
\label{lemma: eigenvalue function}
    The function $\phi(\omega)$ defined in \eqref{eq: omega_star} is continuous and convex over $\Delta^L$, and it is differentiable at any $\omega$ for which $\lambda_k(\Sigma(\omega))> \lambda_{k+1}(\Sigma(\omega))$, where $k>0$ is the prescribed subspace dimension.
\end{Lemma}

By Lemma \ref{lemma: eigenvalue function}, the function $\phi(\omega)$ is convex but need not be differentiable everywhere.
Nevertheless, it admits an explicit subgradient characterization, which enables efficient first-order algorithms to solve $\min_{\omega}\phi(\omega)$. The following lemma provides the expression for a subgradient of $\phi(\omega)$, whose proof is deferred to Appendix \ref{proof of lem: subgrad_phi}.
\begin{Lemma}[Subgradient of $\phi$]
\label{lem: subgrad_phi}
For any $\omega\in \Delta^L$, let $V(\omega)\in\mathcal{O}^{d\times k}$ be the top-$k$ eigenvectors of $\Sigma(\omega)$.
Then the vector $g(\omega) = (g_1(\omega),..., g_L(\omega))\in\mathbb{R}^L$ with coordinates
\[
g_l(\omega):=\big\langle \Sigma^{(l)},\, V(\omega) (V(\omega))^\intercal\big\rangle, \qquad l=1,\dots,L,
\]
is a subgradient of $\phi$ at $\omega$, i.e.,
\[
\phi(\omega') \ge \phi(\omega) + \langle g(\omega),\, \omega'-\omega\rangle,
\qquad \forall\,\omega'\in\Delta^L.
\] 
\end{Lemma}

\subsubsection{Algorithm to minimize $\phi(\omega)$}
\label{sec: alg dual}
This subsection details the algorithm that computes a least favorable mixture $\omega^*\in\argmin_\omega \phi(\omega)$ in \eqref{eq: omega_star}. We describe it at the population level; the same procedure applies to the empirical second-moment matrices. As in Algorithm~\ref{algo: mp}, we employ mirror updates to respect the geometry of the simplex. Since $\phi$ is convex and may be nondifferentiable (Lemma~\ref{lemma: eigenvalue function}), the updates use the subgradients of Lemma~\ref{lem: subgrad_phi}.

We initialize $\omega^0 = \frac{1}{L}{\bf 1}_L\in \Delta^L$. 
For each iteration $t=0,1...,T-1$, given the weight vector $\omega^t$, we compute $V^t\in \mathcal{O}^{d\times k}$, the top-$k$ eigenvectors of $\Sigma(\omega^{t})$. By Lemma \ref{lem: subgrad_phi}, this yields a subgradient $g^t$ of $\phi(\omega^t)$ as follows:
\begin{equation}
    \label{eq: subgrad of phi}
    g_l^t = \big\langle \Sigma^{(l)}, V^t (V^t)^\intercal \big\rangle,
    \quad l=1,\dots,L.
\end{equation}
Given the step size $\eta>0$, we perform a mirror update over the constraint $\omega\in \Delta^L$:
\[
\omega^{t+1} = \argmin_{\omega\in \Delta^L}\left\{\eta \langle g^t, \omega\rangle + D_{\psi_2}(\omega, \omega^t)\right\},
\]
where $D_{\psi_2}(\omega, \omega^t)=\sum_{l=1}^L \omega_l \log \frac{\omega_l}{\omega^t_l}$ is the Bregman divergence associated with the simplex $\Delta^L$, defined in \eqref{eq: Breg separate}. 
This update admits a closed-form solution:
    \begin{equation}
        \label{eq: update on omega}
        \omega^{t+1}_l 
    = \frac{\omega^{t}_l \exp\{-\eta g_l^{t}\}}{\sum_{r=1}^L \omega^{t}_r \exp\{-\eta g_r^{t}\}}, 
    \quad l=1,\dots,L.
    \end{equation}
The derivation of \eqref{eq: update on omega} follows from standard mirror descent updates on the simplex; see Chapter~4 of \citet{bubeck2015convex} for details.

After $T$ iterations, we form the averaged iterate
\[
    \widehat{\omega}_T := \frac{1}{T}\sum_{t=0}^{T-1} \omega^t,
\]
as the output to approximately solves the optimization problem in \eqref{eq: omega_star}.
We summarize the overall procedure in Algorithm~\ref{algo: dual}.

\begin{algorithm}[!ht]
\DontPrintSemicolon
\SetAlgoLined
\SetNoFillComment
\LinesNotNumbered 
\caption{Algorithm to minimize $\phi(\omega)$ for $\omega\in \Delta^L$}
\label{algo: dual}
\KwData{Second-moment Matrices $\{{\Sigma}^{(l)}\}_{l=1}^L$; subspace dimension $k$; step size $\eta$; iterations $T$}

  Initialize $\omega^0 = \frac{1}{L}{\bf 1}_L\in \Delta^L$;

\For{$t = 0,\dots,T-1$}{

  Compute the top-$k$ eigenvectors $V^t$ of ${\Sigma}(\omega^t)=\sum_{l=1}^L \omega_l^t {\Sigma}^{(l)}$;

    Compute the subgradient $g^t$ as in \eqref{eq: subgrad of phi};
  
  Update $\omega^{t+1}$ by mirror update in \eqref{eq: update on omega};
}

Set $\widehat{\omega}_T= \frac{1}{T}\sum_{t=0}^{T-1} \omega^{t}$.\;

\Return $\widehat{\omega}_T$ 
\end{algorithm}

\subsection{Convergence under the Eigengap Condition}
\label{appendix: eigengap conv}
This subsection supplements Section~\ref{sec: dual-form}. Under the eigengap Condition~\ref{cond: eigengap}, Theorem~\ref{thm: fantope tight} shows that the relaxed and the original StablePCA have the same unique solution $P^*$. We show that the outputs of Algorithm~\ref{algo: mp} converge to $P^*$ itself, which strengthens the objective-value guarantees of Section~\ref{sec: theory}. In particular, we establish that (i) the relaxed output $\widehat{M}_T$ converges to $P^*$, (ii) the certificate $\tau$ vanishes, and (iii) the projected output $\widehat{P}_T$ solves the original problem and converges to $P^*$ as well. We state the three results in turn. The proofs are given in Appendix~\ref{proof of thm: conv for tau}.

\begin{Theorem}[Convergence to the relaxed StablePCA under the eigengap]
\label{thm: conv relaxed eigengap}
Suppose that Conditions~\ref{cond: eigengap} and \ref{cond: subgauss} hold, that $n\geq\mathbf{r}_{\max}$, and that the step sizes are specified as in \eqref{eq: step sizes}. For $t\in[0, n - \mathbf{r}_{\max}]$, define
\begin{equation}
    \mathcal{E}_{n,T} := C\rho_{\max}\cdot k\left(\sqrt{\frac{t}{n}} + \min\left\{\frac{\rho_{\max}}{\delta^*}\cdot\frac{\mathbf{r}_{\max} + t}{n},\ \sqrt{\frac{\mathbf{r}_{\max} + t}{n}}\right\} + \frac{\sqrt{k\log(d/k)\log L}}{T}\right),
    \label{eq: eigengap rate}
\end{equation}
where $C > 0$ denotes a constant depending only on the sub-Gaussian parameter $\sigma$. Then, with probability at least $1 - 4Le^{-t}$, the output $\widehat{M}_T$ of Algorithm~\ref{algo: mp} satisfies
\[
\max_{M\in\Fc^k}\left[\min_{\omega\in\Delta^L}\langle \Sigma(\omega), M\rangle\right] - \min_{\omega\in\Delta^L}\langle \Sigma(\omega), \widehat{M}_T\rangle \leq \mathcal{E}_{n,T}
\qquad\text{and}\qquad
\|\widehat{M}_T - P^*\|_F^2\leq \frac{2\,\mathcal{E}_{n,T}}{\delta^*}.
\]
\end{Theorem}

Compared with Theorem~\ref{thm: new conv}, the current theorem improves in two ways under the eigengap condition. We first comment on the objective gap of $\widehat{M}_T$ from solving the relaxed StablePCA. Note that the leading term $\sqrt{t/n}$ does not involve the effective rank $\mathbf{r}_{\max}$, while the effective rank enters only through the minimum term in \eqref{eq: eigengap rate}, which is never slower than that of  Theorem~\ref{thm: new conv}, however small the eigengap $\delta^*$ may be. If $n\gtrsim\rho_{\max}^2\mathbf{r}_{\max}^2/(\delta^{*2}t)$, the finite-sample error is dominated by the leading term $k\rho_{\max}\sqrt{t/n}$.
Second, the second inequality of Theorem~\ref{thm: conv relaxed eigengap} converts the objective gap into a distance, where we establish that the output $\widehat{M}_T$ converges to the solution $P^*$ itself.

\begin{Theorem}[Convergence of the certificate]
\label{thm: conv for tau}
Under the conditions of Theorem~\ref{thm: conv relaxed eigengap}, with probability at least $1 - 4Le^{-t}$, the certificate $\tau$ in \eqref{eq: certificate} satisfies
\[
\left|\tau\right|^2 \leq C\rho_{\max}^2\cdot\frac{k}{\delta^*}\,\mathcal{E}_{n,T}.
\]
\end{Theorem}
Theorem~\ref{thm: conv for tau} confirms that, under the eigengap and the growth conditions stated below, the certificate vanishes as $n$ and $T$ grow, so that the projection step of Algorithm~\ref{algo: mp} loses nothing in the limit. Combining it with Theorem~\ref{thm: conv relaxed eigengap} yields the convergence of the projected output $\widehat{P}_T$ for the original StablePCA.

\begin{Theorem}[Convergence to the original StablePCA under the eigengap]
\label{thm: conv original eigengap}
Under the conditions of Theorem~\ref{thm: conv relaxed eigengap}, with probability at least $1 - 4Le^{-t}$, the projected output $\widehat{P}_T$ of Algorithm~\ref{algo: mp} satisfies
\[
\max_{P\in\Pc^k}\left[\min_{\omega\in\Delta^L}\langle \Sigma(\omega), P\rangle\right] - \min_{\omega\in\Delta^L}\langle \Sigma(\omega), \widehat{P}_T\rangle \leq \mathcal{E}_{n,T} + 3\rho_{\max}\sqrt{\frac{k\,\mathcal{E}_{n,T}}{\delta^*}}
\qquad\text{and}\qquad
\|\widehat{P}_T - P^*\|_F^2\leq \frac{8\,\mathcal{E}_{n,T}}{\delta^*}.
\]
\end{Theorem}
Compared with Theorem~\ref{thm: global conv rates alg}, the data-dependent certificate $\tau$ is replaced by the explicit term $3\rho_{\max}\sqrt{k\mathcal{E}_{n,T}/\delta^*}$, which bounds the loss of the projection step. The second inequality says that $\widehat{P}_T$ converges to the solution $P^*$ itself, not merely in objective value.

Treating the subspace dimension $k$, the number of sources $L$, the operator norm $\rho_{\max}$, the sub-Gaussian parameter $\sigma$, and the eigengap $\delta^*$ as constants, the bounds of Theorems~\ref{thm: conv relaxed eigengap} and \ref{thm: conv original eigengap} yield
\begin{equation}
    \begin{aligned}
    \max\left\{\|\widehat{M}_T - P^*\|_F^2, \ \|\widehat{P}_T - P^*\|_F^2 \right\}\ &\lesssim \ \sqrt{\frac{t}{n}} + \frac{\mathbf{r}_{\max}+t}{n} + \frac{\sqrt{\log d}}{T}.
    \end{aligned}
\end{equation}
Consequently, under the eigengap, both outputs converge to $P^*$ whenever $(\mathbf{r}_{\max}+t)/n\to0$ and $\sqrt{\log d}/T\to0$, 
at a rate where the effective rank $\mathbf{r}_{\max}$ enters at the order of $n^{-1}$ and the ambient dimension $d$ enters via a logarithmic factor. Taking $t-\log(4L)\to\infty$ while maintaining $t\leq n-\mathbf{r}_{\max}$ makes the probability tend to one.

\subsection{Additional Discussions on Alternative Robust Multi-source PCA}
\label{subsec: dual appendix}
This subsection supplements Section~\ref{subsec: alternative}. We first show that our FairPCA formulation \eqref{eq: fairPCA}, defined through the uncertainty class $\Cc$, is equivalent to the worst-case regret over the individual sources used in the FairPCA literature, and we then extend Algorithm~\ref{algo: mp} to SquaredPCA and FairPCA.

\subsubsection{Equivalence between FairPCA Formulations}
\label{subsec: dual appendix - connection}
The following lemma shows that the FairPCA objective, originally defined via a worst-case distribution over the uncertainty class $\Cc$, admits an equivalent formulation as a worst-case regret over the individual source distributions. The proof is provided in Appendix \ref{proof of lemma: fairpca equiv}.
\begin{Lemma}
\label{lemma: fairpca equiv}
    Let $\Cc$ be the uncertainty class defined in \eqref{eq: uncertainty class}, and let $P^{\rm fair}$ be defined in \eqref{eq: fairPCA}. Then
    \[
    P^{\rm fair}\in \argmin_{P\in \Pc^k}\max_{\QQ\in \mathcal{C}}{\rm Regret}_\QQ(P) = \argmin_{P\in \Pc^k} \max_{l\in [L]}{\rm Regret}_{\TT^{(l)}}(P).
    \]
\end{Lemma}
As an immediate corollary, when there are only two source distributions $\{\TT^{(1)}, \TT^{(2)}\}$ with $L=2$, the FairPCA solution satisfies
\[
P^{\rm fair} \in \argmin_{P\in \Pc^k}\max\left\{{\rm Regret}_{\TT^{(1)}}(P), {\rm Regret}_{\TT^{(2)}}(P)\right\}.
\]
This coincides exactly with the original FairPCA formulation studied in the prior works \citep{samadi2018price, shen2025hidden}.

\subsubsection{Fantope-relaxation of SquaredPCA and FairPCA}
\label{subsec: dual appendix - fantope}

Analogous to StablePCA, we apply the Fantope relaxation to SquaredPCA and FairPCA by replacing $P\in \Pc^k$ with its convex hull $M\in \Fc^k$.
For any $\omega\in \Delta^L$, we define the mixtures of the shifted source matrices of Proposition~\ref{prop: alternative reformulation},
\begin{equation}
    \Sigma^{\rm sq}(\omega) := \sum_{l=1}^L\omega_l\Big(\Sigma^{(l)}-\frac{1}{k}\sum_{i=1}^{\textcolor{niblue}{d}} \lambda_i^{(l)}{\bf I}_d\Big), \quad \textrm{and}\quad
\Sigma^{\rm fair}(\omega) := \sum_{l=1}^L\omega_l\Big(\Sigma^{(l)}-\frac{1}{k}\sum_{i=1}^{\textcolor{nired}{k}} \lambda_i^{(l)}{\bf I}_d\Big),
\label{eq: weight covariance alter}
\end{equation}
where $\lambda_i^{(l)}$ denotes the $i$-th largest eigenvalue of $\Sigma^{(l)}$, for each $l\in [L]$.

It follows from Proposition \ref{prop: alternative reformulation} that the Fantope-relaxed SquaredPCA and FairPCA are defined as follows:
\begin{equation}
    M^{\rm sq}\in \argmax_{M\in \Fc^k} \min_{\omega\in \Delta^L}\left\langle \Sigma^{\rm sq}(\omega), \; M  \right\rangle,
    \label{eq: squaredpca - fantope}
\end{equation}
and
\begin{equation}
        M^{\rm fair} \in \argmax_{M\in \Fc^k} \min_{\omega\in \Delta^L}\left\langle  \Sigma^{\rm fair}(\omega),\; M  \right\rangle.
    \label{eq: fairpca - fantope}
\end{equation}
As discussed in Section~\ref{subsec: alternative}, these problems differ from the relaxed StablePCA \eqref{eq: fantope MPCA} only through the matrices subtracted from each source. Algorithm~\ref{algo: mp} therefore applies verbatim to the shifted inputs, as summarized in Algorithm~\ref{algo: mp-alter}, with $\hat\rho_{\max}$ in the step sizes \eqref{eq: step sizes} computed from the shifted matrices; the projection step and the certificate of Section~\ref{sec: algo} apply as well.
\begin{algorithm}[!ht]
\DontPrintSemicolon
\SetAlgoLined
\SetNoFillComment
\LinesNotNumbered 
\caption{Mirror-Prox Algorithm for SquaredPCA and FairPCA}
\label{algo: mp-alter}
\KwData{Empirical matrices $\{\widehat{\Sigma}^{(l)}\}_{l=1}^L$; subspace dimension $k$; step sizes $\eta_M,\eta_\omega$; iterations $T$}

\For{$l=1,...,L$}{
Compute the eigenvalues of $\widehat{\Sigma}^{(l)}$: $\widehat{\lambda}^{(l)}_1\geq \cdots\geq \widehat{\lambda}^{(l)}_d$;
}
\If{\textnormal{SquaredPCA}}{
    Run Algorithm \ref{algo: mp} with the input $\widehat{\Sigma}^{(l)}$ replaced by $\widehat{\Sigma}^{(l)} - \frac{1}{k}\sum_{i=1}^{\textcolor{niblue}{d}} \widehat{\lambda}^{(l)}_i {\bf I}_d$;
}
\ElseIf{\textnormal{FairPCA}}{
    Run Algorithm \ref{algo: mp} with the input $\widehat{\Sigma}^{(l)}$ replaced by $\widehat{\Sigma}^{(l)} - \frac{1}{k}\sum_{i=1}^{\textcolor{nired}{k}} \widehat{\lambda}^{(l)}_i {\bf I}_d$.
}
\end{algorithm}

\subsection{Mirror Updates and Bregman Divergences}
\label{appendix: defs}

This subsection briefly reviews Bregman divergences and mirror-based updates,
which form the algorithmic foundation of our optimization procedure in Section \ref{sec: algo}.
A comprehensive introduction can be found in Chapter~4 of
\citet{bubeck2015convex}.

Mirror-based updates generalize projected gradient descent to \emph{non-Euclidean geometries}. Rather than taking a gradient step followed by a possibly expensive Euclidean projection, a mirror update linearizes the objective and measures the proximity between iterates with a Bregman divergence induced by a strictly convex mirror map. When the mirror map is matched to the constraint set, feasibility is enforced implicitly and the update admits a closed form, as on the simplex and on the Fantope in Proposition~\ref{prop: closed form revised}.

\begin{Definition}[Mirror Map and Bregman Divergence]
\label{def: bregman}
Let $\mathcal{D}\subseteq\mathbb{R}^m$ be a convex set.
A function $\psi:\mathcal{D}\to\mathbb{R}$ is called a {\textbf{mirror map}} if it is
strictly convex and continuous on $\mathcal{D}$, and continuously differentiable on its relative interior. Gradients are taken in the affine hull of $\mathcal{D}$, or from a differentiable ambient extension.
The {\textbf{Bregman divergence}} induced by $\psi$ is defined as
\begin{equation}
D_{\psi}(x,y)
:= \psi(x) - \psi(y) - \langle \nabla \psi(y),\, x-y\rangle,
\label{eq: bregman}
\end{equation}
for $x\in\mathcal{D}$ and $y$ in its relative interior, with boundary values in the first argument defined by continuity.
\end{Definition}
Unlike the squared Euclidean distance $\|x-y\|_2^2$, the Bregman divergence is
generally asymmetric and adapts to the geometry of the constraint set.
In this work, we choose mirror maps $\psi_1(M) = {\rm Tr}(M\log M)$ for Fantope $M\in\Fc^k$ and $\psi_2(\omega)=\sum_{l=1}^L \omega_l \log \omega_l$ for simplex $\omega\in \Delta^L$. We subsequently derive their associated Bregman divergences as specified in \eqref{eq: Breg separate}.

\noindent Derivation of $D_{\psi_1}(M,M')$.

For $M,M'\in\Fc^k$ with $M'\succ0$, recall
\[
D_{\psi_1}(M,M')
=
\psi_1(M)-\psi_1(M')-\langle \nabla\psi_1(M'),\, M-M'\rangle,
\qquad
\nabla\psi_1(M')=\log M' + I.
\]
Using $\Tr(M)=\Tr(M')=k$, we have $\langle I, M-M'\rangle = \Tr(M-M')=0$, hence
\begin{equation*}
D_{\psi_1}(M,M')
=
\Tr(M\log M)-\Tr(M\log M') - \Tr(M') + \Tr(M')
=
\Tr(M\log M)-\Tr(M\log M').
\end{equation*}

\noindent Derivation of $D_{\psi_2}(\omega,\omega')$.

For $\omega,\omega'\in\Delta^L$ with $\omega'_l>0$ for every $l$, it holds
\[
\begin{aligned}
    D_{\psi_2}(\omega,\omega') &= \psi_2(\omega) - \psi_2(\omega') - \langle\nabla\psi_2(\omega'),\;\omega - \omega'\rangle \\
&= \sum_{l=1}^L\omega_l\log \omega_l - \sum_{l=1}^L\omega_l'\log \omega_l' - \sum_{l=1}^L (\log \omega_l' + 1)(\omega_l-\omega_l') \\
&= \sum_{l=1}^L \omega_l \log \frac{\omega_l}{\omega_l'} =: {\rm KL}(\omega\|\omega').
\end{aligned}
\]

In the remainder of this subsection, we consider a convex function $f:\mathcal{D}\to\mathbb{R}$, with a subgradient $\partial f(\cdot)$, and we aim to minimize it with $\min_{x\in \mathcal{D}} f(x)$. 
We now introduce the mirror descent update to solve it.
\begin{Definition}[Mirror Descent Update]
\label{def: mirror descent}
Let $x^t$ be a current iterate in the differentiability domain of $\psi$.
The mirror descent update is defined as
\begin{equation}
x^{t+1}
= \argmin_{x\in\mathcal{D}}
\left\{
\eta \langle \partial f(x^t), x\rangle
+ D_{\psi}(x,x^t)
\right\}.
\label{eq: mirror update}
\end{equation}
\end{Definition}
The update in \eqref{eq: mirror update} trades off descent in the linearized
objective against proximity to the previous iterate, where proximity is
measured in the geometry induced by the mirror map $\psi$.
Different choices of $\psi$ yield different algorithms that are tailored to
the structure of $\mathcal{D}$.

Two choices of the mirror map are relevant to this paper. With the entropy $\psi(x)=\sum_{j=1}^m x_j\log x_j$ on the simplex $\mathcal{D} = \Delta^m$, the Bregman divergence is the Kullback--Leibler divergence $D_\psi(x,y) = \sum_j x_j\log(x_j/y_j)$, and \eqref{eq: mirror update} has the multiplicative closed form
\[
x_j^{t+1} = \frac{x_j^t\exp(-\eta[\partial f(x^t)]_j)}{\sum_{s=1}^m x_s^t\exp(-\eta[\partial f(x^t)]_s)},\qquad j = 1,\dots,m,
\]
which preserves feasibility without any projection; this is the form of the $\omega$-updates in Proposition~\ref{prop: closed form revised}, and its matrix analogue with $\psi_1(M) = {\rm Tr}(M\log M)$ gives the $M$-updates. With the squared Euclidean norm $\psi(x) = \frac12\|x\|_2^2$, the divergence is $\frac12\|x-y\|_2^2$ and \eqref{eq: mirror update} reduces to projected (sub)gradient descent $x^{t+1} = \Pi_{\mathcal{D}}(x^t - \eta\partial f(x^t))$, and to (sub)gradient descent when $\mathcal{D} = \RR^m$.

\subsection{Mirror-Prox for General Convex-Concave Minimax Problems}
\label{sec: general}

This subsection reviews the Mirror-Prox algorithm for general smooth convex--concave minimax problems and states a standard $\mathcal{O}(1/T)$ convergence guarantee. Algorithm~\ref{algo: mp} is obtained by specializing this scheme to the StablePCA objective in \eqref{eq: finite sample}, and the optimization-error bound of Theorem~\ref{thm: conv rates alg} follows from the general guarantee below. The presentation follows Section~5.2.3 of \citet{bubeck2015convex}.

We start with the definition of a saddle point for a minimax problem.
\begin{Definition}[Saddle Point]
\label{def: saddle point}
    Let $f: \mathcal{X}\times \mathcal{Y}\to \mathbb{R}$ be a function. A point $(x^*, y^*)$ is a saddle point of the minimax problem $\min_{x\in \mathcal{X}}\max_{y\in \mathcal{Y}} f(x,y)$, if
    \[
    f(x^*, y^*) = \max_{y\in \mathcal{Y}}f(x^*, y) = \min_{x\in \mathcal{X}} f(x,y^*).
    \]
    Moreover, a point $(\hat{x}, \hat{y})$ is an $\varepsilon$-nearly saddle point of $f(\cdot,\cdot)$ if
    \[
    0\leq \max_{y\in \mathcal{Y}} f(\hat{x}, y) - \min_{x\in \mathcal{X}} f(x, \hat{y}) \leq \varepsilon.
    \]
\end{Definition}

Throughout this subsection, we aim to identify a (nearly) saddle point of the convex--concave minimax problem
\[
\min_{x\in \mathcal{X}}\; \max_{y\in \mathcal{Y}} f(x,y),
\]
where $\mathcal{X}$ and $\mathcal{Y}$ are compact convex sets, and
\begin{itemize}
    \item for every fixed $y\in\mathcal{Y}$ the map $x\mapsto f(x,y)$ is convex on $\mathcal{X}$, and
    \item for every fixed $x\in\mathcal{X}$ the map $y\mapsto f(x,y)$ is concave on $\mathcal{Y}$.
\end{itemize}

Let $\Xc, \Yc$ be equipped with mirror maps: $\psi_\Xc: \Dc_\Xc\to \RR$ and $\psi_\Yc: \Dc_\Yc\to \RR$, respectively, with induced Bregman divergences $D_{\psi_{\mathcal{X}}}$ and $D_{\psi_{\mathcal{Y}}}$. Here, we use $\Dc_\Xc$ and $\Dc_\Yc$ to denote the domain for mirror maps, which could be different from $\Xc, \Yc$.
We work on the product space $\mathcal{Z}\coloneqq \mathcal{X}\times \mathcal{Y}$ with mirror map
\[
\psi(z)\coloneqq a\,\psi_{\mathcal{X}}(x)+b\,\psi_{\mathcal{Y}}(y),
\qquad z=(x,y)\in\mathcal{Z},
\]
that is defined on $\Dc_\Zc =\Dc_\Xc\times \Dc_\Yc$,
for fixed constants $a,b>0$. We equip the product space with
\[
\|(x,y)\|_{\mathcal Z}:=\sqrt{a\|x\|_{\mathcal X}^2+b\|y\|_{\mathcal Y}^2}.
\]
We assume that the mirror maps extend continuously to the feasible sets, that their differentiability domains contain the relative interiors of those sets, and that the initial minimizer and all prox minimizers below exist in these domains. Bregman comparisons with boundary points are understood by continuity.

The associated Bregman divergence decomposes as
\[
D_{\psi}(z,z')
= a\,D_{\psi_{\mathcal{X}}}(x,x') + b\,D_{\psi_{\mathcal{Y}}}(y,y'),
\qquad z=(x,y),\; z'=(x',y').
\]
Additionally, we define the monotone saddle-point operator
\begin{equation}
g(z)\coloneqq \big(\nabla_x f(x,y),\; -\nabla_y f(x,y)\big), \qquad z=(x,y)\in\mathcal{Z}.
\label{eq: g(z) def}
\end{equation}

We now describe the Mirror-Prox algorithm.
We initialize at $z_0\in \argmin_{z\in \Zc\cap \Dc_\Zc} \psi(z)$. Given a step-size $\eta>0$, for each iteration $t=0,1...,T-1$, starting from the current iterate $z_t=(x_t, y_t)$, the first step computes the midpoint iterate $z_{t+\frac12}=(x_{t+\frac12}, y_{t+\frac{1}{2}})$ as follows:
\begin{equation}
    z_{t+\frac12} = \argmin_{z\in \mathcal{Z}\cap \Dc_\Zc} \eta \left\langle g(z_t), z\right\rangle + D_{\psi}(z, z_t).
    \label{eq: MP general first}
\end{equation}
The second step then updates the iterate using gradients evaluated at the
midpoint $z_{t+\frac12}$:
\begin{equation}
    z_{t+1}= \argmin_{z\in \mathcal{Z}\cap \Dc_\Zc} \eta \left\langle g(z_{t+\frac12}), z\right\rangle + D_{\psi}(z, z_t).
    \label{eq: MP general second}
\end{equation}
After completing $T$ iterations, Mirror-Prox averages the midpoints $\{(x_{t+\frac12}, y_{t+\frac12})\}_{t=0}^{T-1}$ as the final output:
\begin{equation}
    (\bar{x}_T, \bar{y}_T) := \left(\frac{1}{T}\sum_{t=0}^{T-1} x_{t+\frac12},\; \frac{1}{T}\sum_{t=0}^{T-1} y_{t+\frac12}\right).
    \label{eq: MP general}
\end{equation}

We next present the convergence analysis of the Mirror-Prox algorithm. We impose the following necessary conditions.
\begin{Condition}
    \label{cond: mirror and smooth}
    Suppose that the mirror maps $\psi_\Xc(x), \psi_\Yc(y)$ are strongly convex such that
    \begin{itemize}
        \item $\psi_\Xc$ is $\mu_\Xc$-strongly convex on $\Xc\cap \Dc_\Xc$ w.r.t. $\|\cdot\|_\Xc$. Define $R^2_\Xc = \sup_{x\in \Xc\cap \Dc_\Xc} \psi_\Xc(x)-\inf_{x\in \Xc\cap \Dc_\Xc}\psi_\Xc(x)$.
        \item $\psi_\Yc$ is $\mu_\Yc$-strongly convex on $\Yc\cap \Dc_\Yc$ w.r.t. $\|\cdot\|_{\Yc}$. Define $R^2_\Yc = \sup_{y\in \Yc\cap \Dc_\Yc} \psi_\Yc(y) - \inf_{y\in \Yc\cap \Dc_\Yc}\psi_\Yc(y)$.
    \end{itemize}
    Moreover, the convex--concave function $f(x,y)$ is $(\beta_{11},\beta_{12},\beta_{21},\beta_{22})$-smooth, such that for any $x,x'\in \Xc$ and $y,y'\in \Yc$,
    \[
    \begin{aligned}
    \|\nabla_x f(x,y) - \nabla_x f(x',y)\|_{\mathcal{X},*}
    &\le \beta_{11}\|x-x'\|_{\mathcal{X}},\\
    \|\nabla_x f(x,y) - \nabla_x f(x,y')\|_{\mathcal{X},*}
    &\le \beta_{12}\|y-y'\|_{\mathcal{Y}},\\
    \|\nabla_y f(x,y) - \nabla_y f(x',y)\|_{\mathcal{Y},*}
    &\le \beta_{21}\|x-x'\|_{\mathcal{X}},\\
    \|\nabla_y f(x,y) - \nabla_y f(x,y')\|_{\mathcal{Y},*}
    &\le \beta_{22}\|y-y'\|_{\mathcal{Y}}.
    \end{aligned}
    \]
    Here $\|\cdot\|_{\mathcal{X},*}$ and $\|\cdot\|_{\mathcal{Y},*}$ denote the dual
    norms of $\|\cdot\|_{\mathcal{X}}$ and $\|\cdot\|_{\mathcal{Y}}$, respectively,
    defined by
    \[
    \|u\|_{\mathcal{X},*}
    := \sup_{\|x\|_{\mathcal{X}}\le 1} \langle u,x\rangle,
    \qquad
    \|v\|_{\mathcal{Y},*}
    := \sup_{\|y\|_{\mathcal{Y}}\le 1} \langle v,y\rangle.
    \]
\end{Condition}

The following theorem establishes the convergence guarantee of the Mirror-Prox
algorithm. The result is adapted from Theorem~5.2 of
\citet{bubeck2015convex}. For completeness, we provide the proof in
Section~\ref{proof: theorem MP general}, which is omitted in
\citet{bubeck2015convex}.

\begin{Lemma}
\label{thm: MP conv general}
Suppose Condition~\ref{cond: mirror and smooth} holds and at least one of the smoothness bounds $\beta_{ij}$ is positive.
With the step size
\[
\eta
= \frac{\mu_{\mathcal{X}}\wedge\mu_{\mathcal{Y}}}
{2\max\left\{
\frac{\beta_{11}}{a},
\frac{\beta_{22}}{b},
\frac{\beta_{12}}{\sqrt{ab}},
\frac{\beta_{21}}{\sqrt{ab}}
\right\}},
\]
the output $(\bar{x}_T, \bar{y}_T)$ of Mirror-Prox algorithm satisfies:
\[
\max_{y\in\mathcal{Y}}
f\!\left(\bar{x}_T,\, y\right)
-
\min_{x\in\mathcal{X}}
f\!\left(x,\, \bar{y}_T\right)
\le
2\max\left\{
\frac{\beta_{11}}{a},
\frac{\beta_{22}}{b},
\frac{\beta_{12}}{\sqrt{ab}},
\frac{\beta_{21}}{\sqrt{ab}}
\right\}
\frac{aR_{\mathcal{X}}^{2}+bR_{\mathcal{Y}}^{2}}
{\mu_{\mathcal{X}}\wedge\mu_{\mathcal{Y}}}
\cdot \frac{1}{T}.
\]
\end{Lemma}
Lemma~\ref{thm: MP conv general} establishes an $\mathcal{O}(1/T)$ convergence rate
for the Mirror-Prox algorithm in terms of the primal–dual gap, indicating that Mirror-Prox identifies an $\mathcal{O}(T^{-1})$-nearly saddle point of the smooth convex-concave function $f(x,y)$, as defined in Definition \ref{def: saddle point}.

We emphasize that the guarantee holds for the ergodic averages of the \emph{midpoint iterates}
\[
\{(x_{t+\frac12},y_{t+\frac12})\}_{t=0}^{T-1}.
\]
Such averaging is standard in convex–concave saddle-point problems and
ensures stability in the presence of the rotational behavior of the
minimax gradient field.

\clearpage
\section{Additional Numerical Experiments}
\label{appendix: additional numerical experiments}

\subsection{Three-dimensional Illustration of Shared-Subspace Recovery}
\label{appendix: shared recovery revision illustration}

This example supplements Section~\ref{subsec: interpretation revision} and shows explicitly that the source-specific directions need not coincide. Let $k=1$, $V=e_1$, $\Gamma^{(1)}=e_2$, $\Gamma^{(2)}=e_3$, and $\Gamma^{(3)}=(e_2+e_3)/\sqrt{2}$. The shared latent factor model is
\begin{equation}
 X^{(1)}=e_1Z^{(1)}+e_2U^{(1)},\quad
 X^{(2)}=e_1Z^{(2)}+e_3U^{(2)},\quad
 X^{(3)}=e_1Z^{(3)}+\frac{e_2+e_3}{\sqrt{2}}U^{(3)},
 \label{eq: example model revision}
\end{equation}
with centered, uncorrelated factors, $\EE[(Z^{(l)})^2]=3$ for all $l$, and $(\EE[(U^{(1)})^2],\EE[(U^{(2)})^2],\EE[(U^{(3)})^2])\allowbreak=(1,9,4)$. All sources have the same variance along the shared direction, while their source-specific directions and variances differ. 

\begin{figure}[!ht]
    \centering
    \includegraphics[width=.66\linewidth]{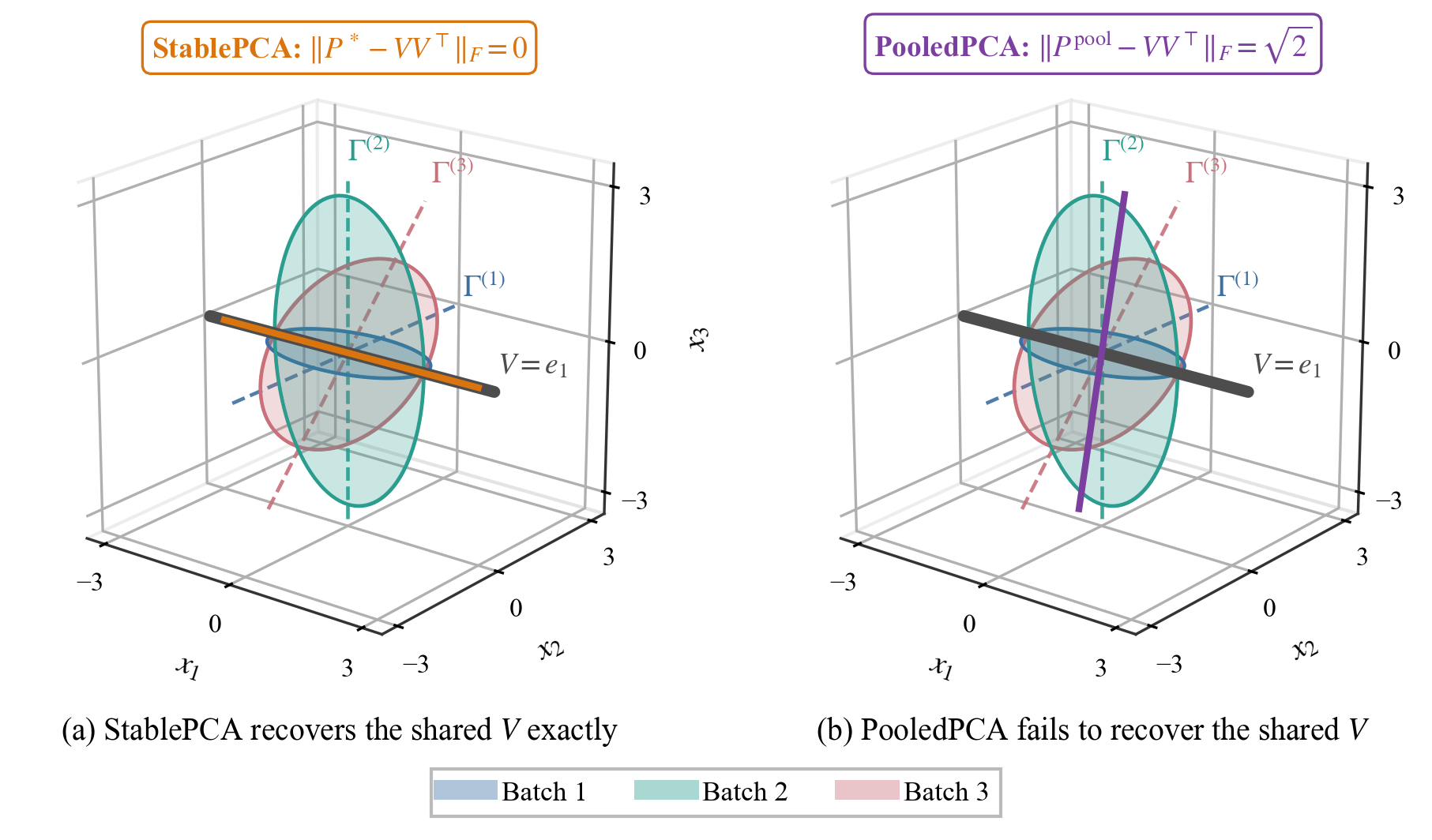}
 \caption{StablePCA and PooledPCA under model \eqref{eq: example model revision}. The ellipses show the three sources. StablePCA recovers the shared subspace ${\rm span}(e_1)$, whereas PooledPCA selects an orthogonal direction.}
 \label{fig: prop1 illus revision}
    \par\vspace{8pt}
    \includegraphics[width=.66\linewidth]{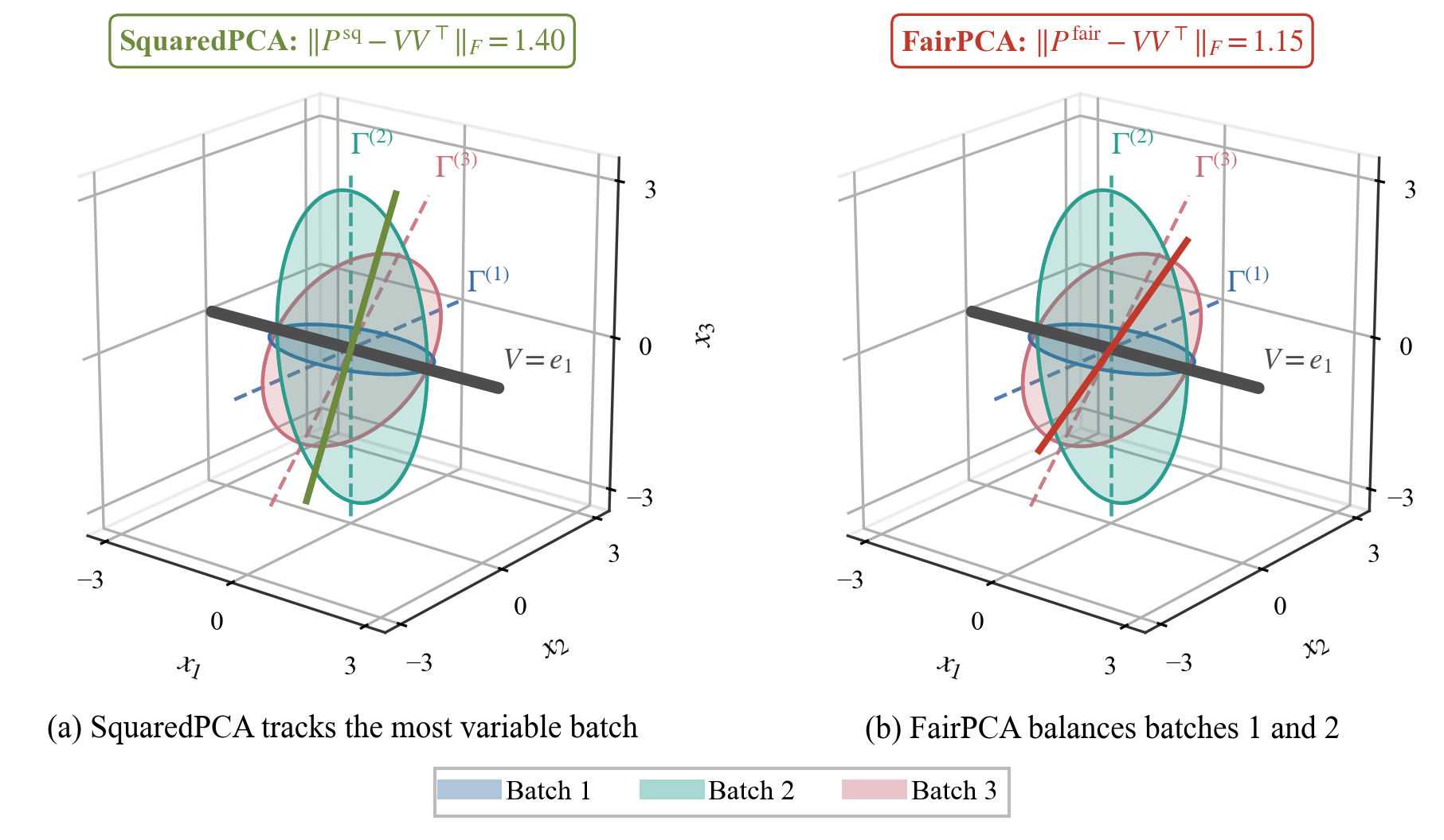}
    \caption{SquaredPCA and FairPCA under the same model as Figure~\ref{fig: prop1 illus revision}. Both select a subspace different from the shared subspace ${\rm span}(e_1)$.}
    \label{fig: alt 3d}
\end{figure}

As shown in Figures \ref{fig: prop1 illus revision} and \ref{fig: alt 3d}, StablePCA uniquely recovers the shared subspace $e_1 e_1^\intercal$, while all other methods, PooledPCA, SquaredPCA, and FairPCA, select a subspace different from ${\rm span}(e_1)$.

\FloatBarrier
\subsection{Additional Simulated Experiments}
\label{appendix: add simus}
\subsubsection{Finite-sample Convergence of Algorithm \ref{algo: mp}}

\begin{figure}[!htbp]
    \centering
    \includegraphics[width=.88\linewidth]{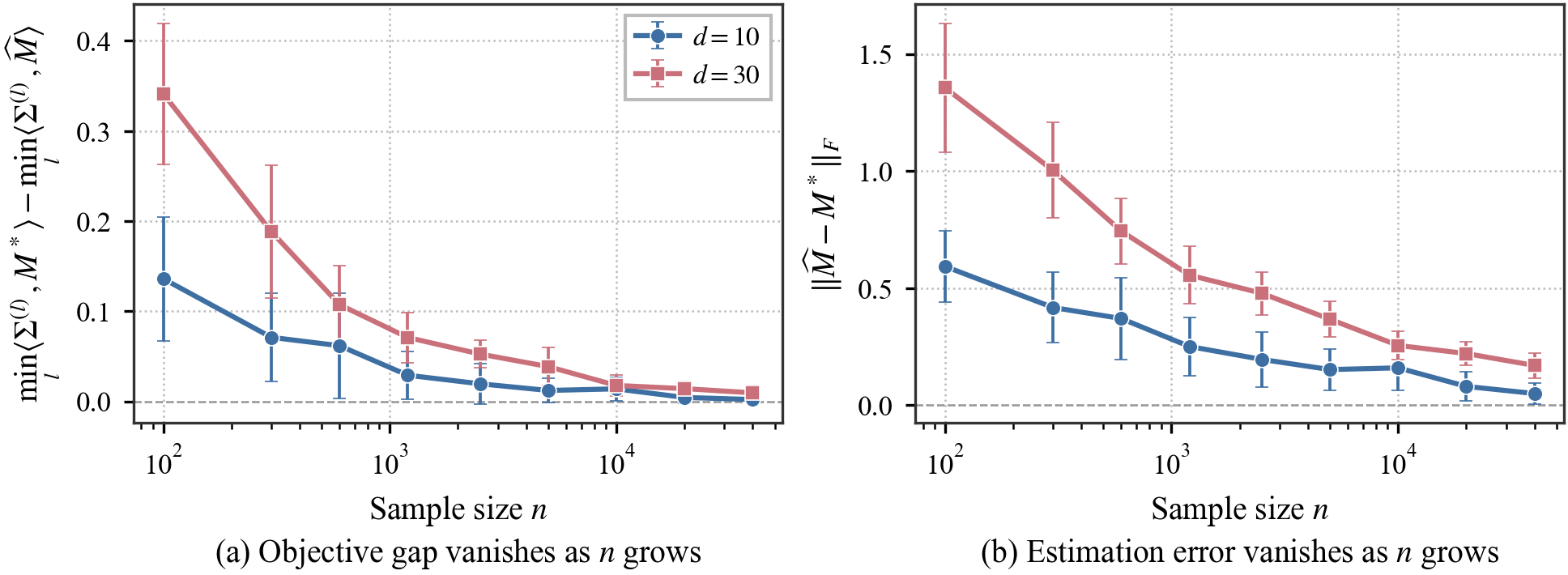}
    \caption{Finite-sample convergence behavior of Fantope-relaxed StablePCA across varying sample sizes $n$ over $[100,40000]$ and ambient dimensions $d\in \{10,30\}$.
(a) Objective gap $\min_{l}\langle\Sigma^{(l)}, M^*\rangle - \min_{l}\langle\Sigma^{(l)}, \widehat{M}_T\rangle$.
(b) Estimation error $\|\widehat{M}-M^*\|_F$.
Results are averaged over 20 replications, with error bars representing one standard deviation.
}
    \label{fig:finite_convergence}
\end{figure}

We examine the finite-sample convergence of {StablePCA}, solved using the proposed Algorithm \ref{algo: mp}. We fix the number of sources at $L=4$, while varying the ambient dimension $d\in \{10, 30\}$, and the sample size $n$ in the range of $[100,40000]$. 
For each configuration $(n,d)$, we run Algorithm \ref{algo: mp} with $T=500$ iterations to obtain the empirical estimator $\widehat{M}$. We compare $\widehat{M}$ to a numerical population reference $M^*$ obtained by running the same algorithm with the population second-moment matrices $\{\Sigma^{(l)}\}_{l=1}^L$. The reference is a numerical approximation to a population optimizer; the comparisons below include its optimization error. Figure \ref{fig:finite_convergence} summarizes the following two metrics on the left and right panels, respectively:
\begin{itemize}
\setlength\itemsep{1pt}\setlength\parskip{0pt}
    \item \emph{Objective Gap (relative to the numerical reference).} We compute $\min_l\langle\Sigma^{(l)}, M^*\rangle - \min_l\langle \Sigma^{(l)}, \widehat{M}\rangle$, which compares its objective value with that of the population reference. This difference need not be nonnegative when the reference is approximate.
    \item \emph{Estimation Error (relative to the numerical reference).} We compute $\|\widehat{M}-M^*\|_F$, which measures the distance from the empirical estimator to the numerical reference.
\end{itemize}

As shown in Figure \ref{fig:finite_convergence}, the objective difference and distance to the population reference decrease with increasing sample size, providing numerical evidence consistent with the convergence theory.

\subsubsection{Generalization Performance}
\label{appendix: simus}

\begin{figure}[!htbp]
    \centering
    \includegraphics[width=.88\linewidth]{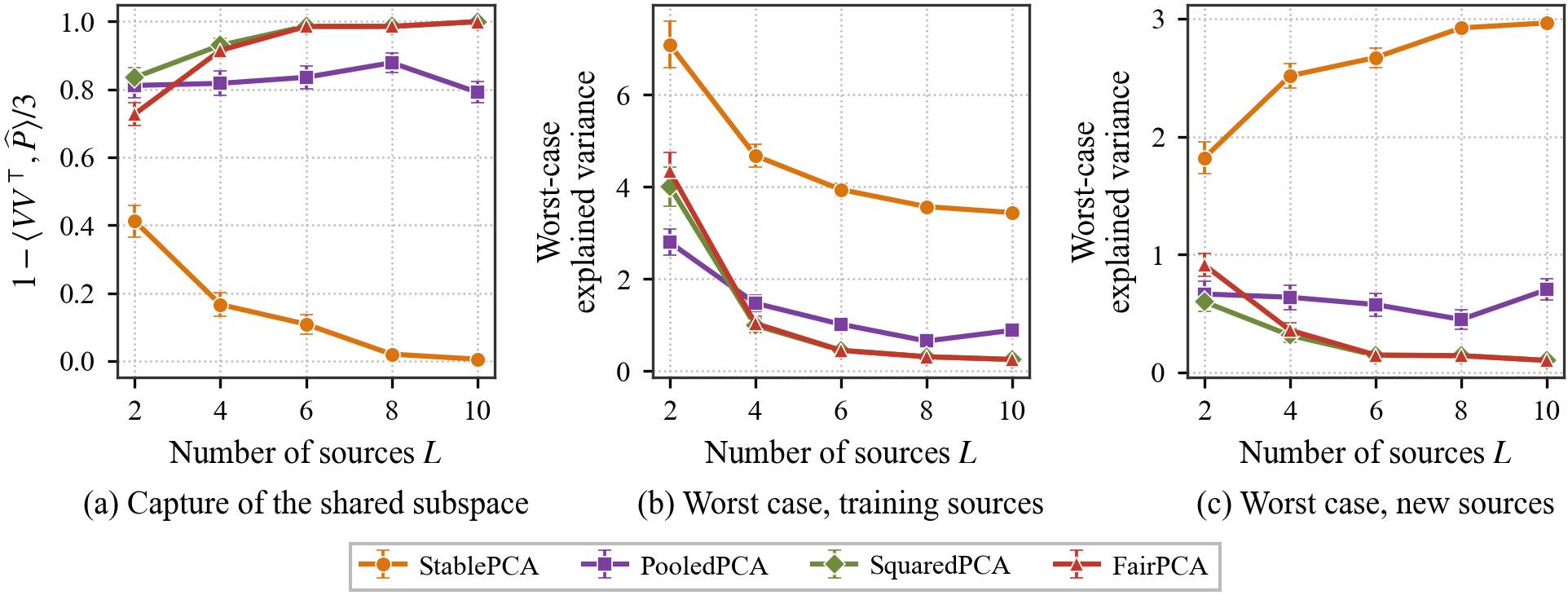}
    \caption{Comparison of {StablePCA}, {PooledPCA}, {SquaredPCA}, and {FairPCA} across different numbers of source domains $L\in\{2,4,6,8,10\}$, when the methods are fitted with $k=5$. (a) Subspace capture error $1 - \langle VV^\intercal, \widehat{P}\rangle/3$ of the shared biological subspace, which replaces the recovery error of Figure~\ref{fig:worst_case} since ${\rm rank}(\widehat{P}) = 5 > 3$; (b) and (c) as in Figure~\ref{fig:worst_case}.}
    \label{fig:worst_case-k5}
\end{figure}

\begin{figure}[!htbp]
    \centering
    \includegraphics[width=.88\linewidth]{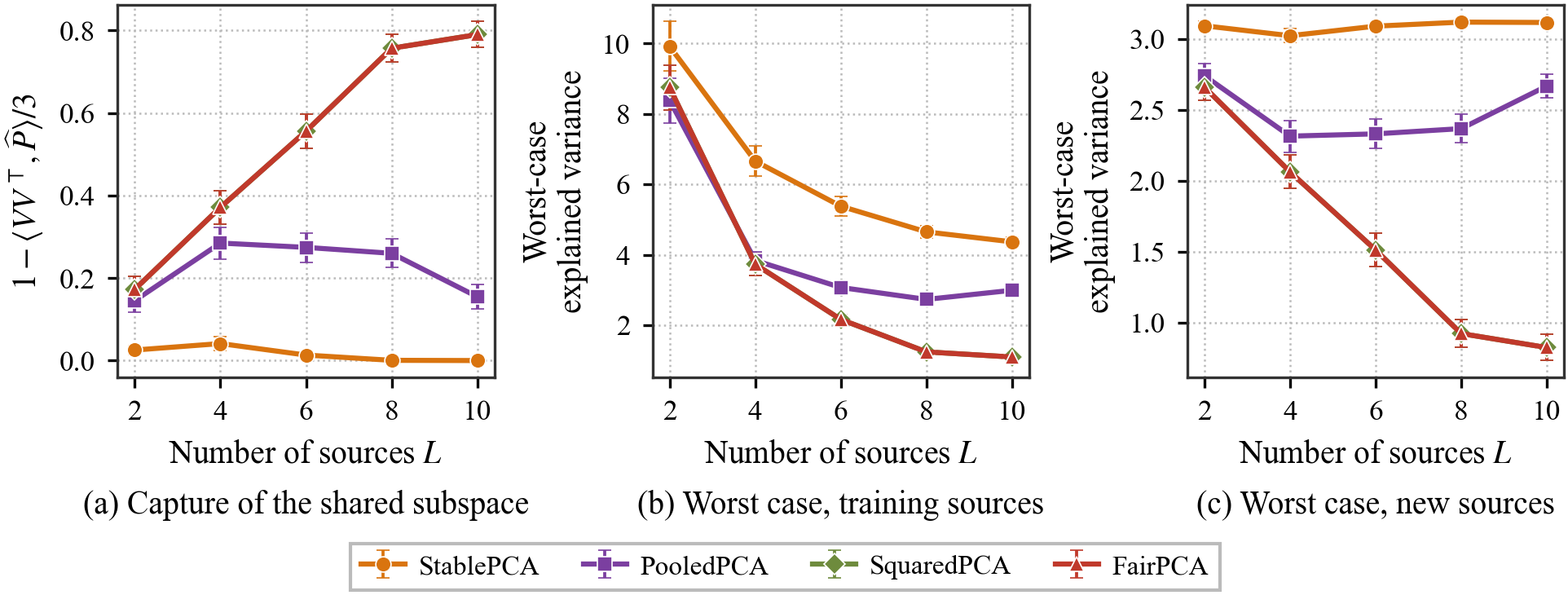}
    \caption{Comparison of {StablePCA}, {PooledPCA}, {SquaredPCA}, and {FairPCA} across different numbers of source domains $L\in\{2,4,6,8,10\}$, when the methods are fitted with $k=10$. (a) Subspace capture error $1 - \langle VV^\intercal, \widehat{P}\rangle/3$ of the shared biological subspace, which replaces the recovery error of Figure~\ref{fig:worst_case} since ${\rm rank}(\widehat{P}) = 10 > 3$; (b) and (c) as in Figure~\ref{fig:worst_case}.}
    \label{fig:worst_case-k10}
\end{figure}

In addition to the results of $k=3$ reported in  Figure \ref{fig:worst_case}, we now present the results for $k=5$ and $k=10$ in Figure \ref{fig:worst_case-k5} and Figure \ref{fig:worst_case-k10}, respectively.

Note that in Figure~\ref{fig:worst_case}, we adopt the metric ``Shared Subspace Recovery Error'' $\|\widehat{P}-VV^\intercal\|_F$ to evaluate how well the estimated projection matrix $\widehat{P}$ recovers the shared subspace ${\rm col}(V)$. This metric is meaningful only when ${\rm rank}(\widehat{P}) = {\rm rank}(V)$: when $k = {\rm rank}(\widehat{P}) > {\rm rank}(V) = 3$, one has $\|\widehat{P}-VV^\intercal\|_F^2 = k + 3 - 2\langle VV^\intercal, \widehat{P}\rangle\geq k - 3$ even if ${\rm col}(\widehat{P})$ contains the shared subspace. For $k = 5$ and $k = 10$ we therefore adopt the metric
$$1 - \frac{1}{{\rm rank}(V)} \langle VV^\intercal, \widehat{P}\rangle,$$
named ``Subspace Capture Error'', which measures the fraction of the shared subspace that is \emph{not} captured by $\widehat{P}$: it lies in $[0,1]$, equals $0$ if and only if ${\rm col}(V)\subseteq{\rm col}(\widehat{P})$, and remains valid when the rank of $\widehat{P}$ exceeds ${\rm rank}(V)$.   

As shown in Figure \ref{fig:worst_case-k5} and Figure \ref{fig:worst_case-k10}, {StablePCA} achieves a substantially smaller capture error and a higher worst-explained variance for both in distribution and out-of-distribution scenarios than other competing methods. This phenomenon aligns with what we have discussed in the main text for Figure \ref{fig:worst_case}.

\subsection{Additional Results for Real Applications}
\label{appendix: real data}

\begin{figure}[!htbp]
    \centering
    \includegraphics[width=.82\linewidth]{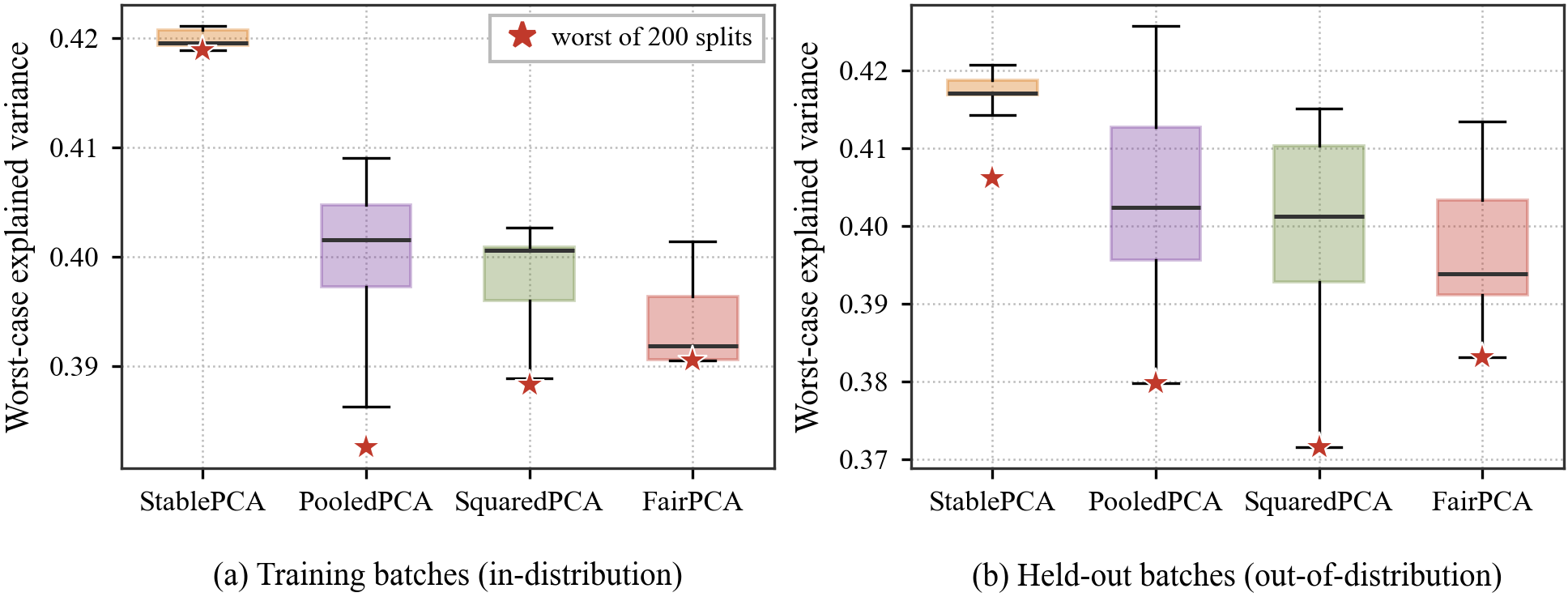}
    \caption{Worst-case explained variance ratio achieved by {StablePCA}, {PooledPCA}, {SquaredPCA}, and {FairPCA} on the single-cell RNA dataset with $k=100$.}
    \label{fig:singlecell-k100}
    \par\vspace{8pt}
    \includegraphics[width=.82\linewidth]{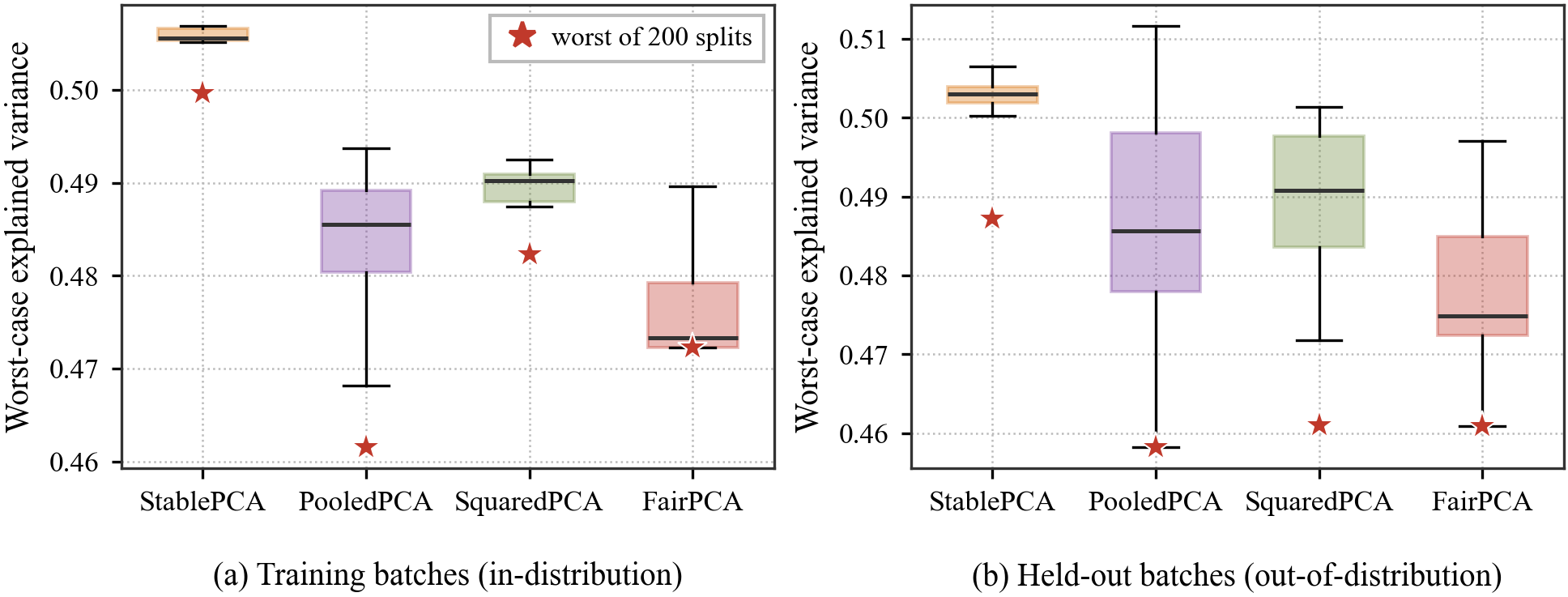}
    \caption{Worst-case explained variance ratio achieved by {StablePCA}, {PooledPCA}, {SquaredPCA}, and {FairPCA} on the single-cell RNA dataset with $k=150$.}
    \label{fig:singlecell-k150}
\end{figure}

In addition to Figure \ref{fig:singlecell} for $k=50$, we present the worst-case explained variance when we set $k=100$ and $k=150$ in Figure \ref{fig:singlecell-k100} and Figure \ref{fig:singlecell-k150}, respectively. Analogous to $k=50$, we find that in both $k=100$ and $k=150$, {StablePCA} consistently achieves the highest worst-case explained variance on both training batches (panel (a)) and test batches (panel (b)), indicating that the low-rank representations it extracts generalize not only to the training batches but also reliably to unseen batches. 

\begin{figure}[!htbp]
    \centering
    \includegraphics[width=.66\linewidth]{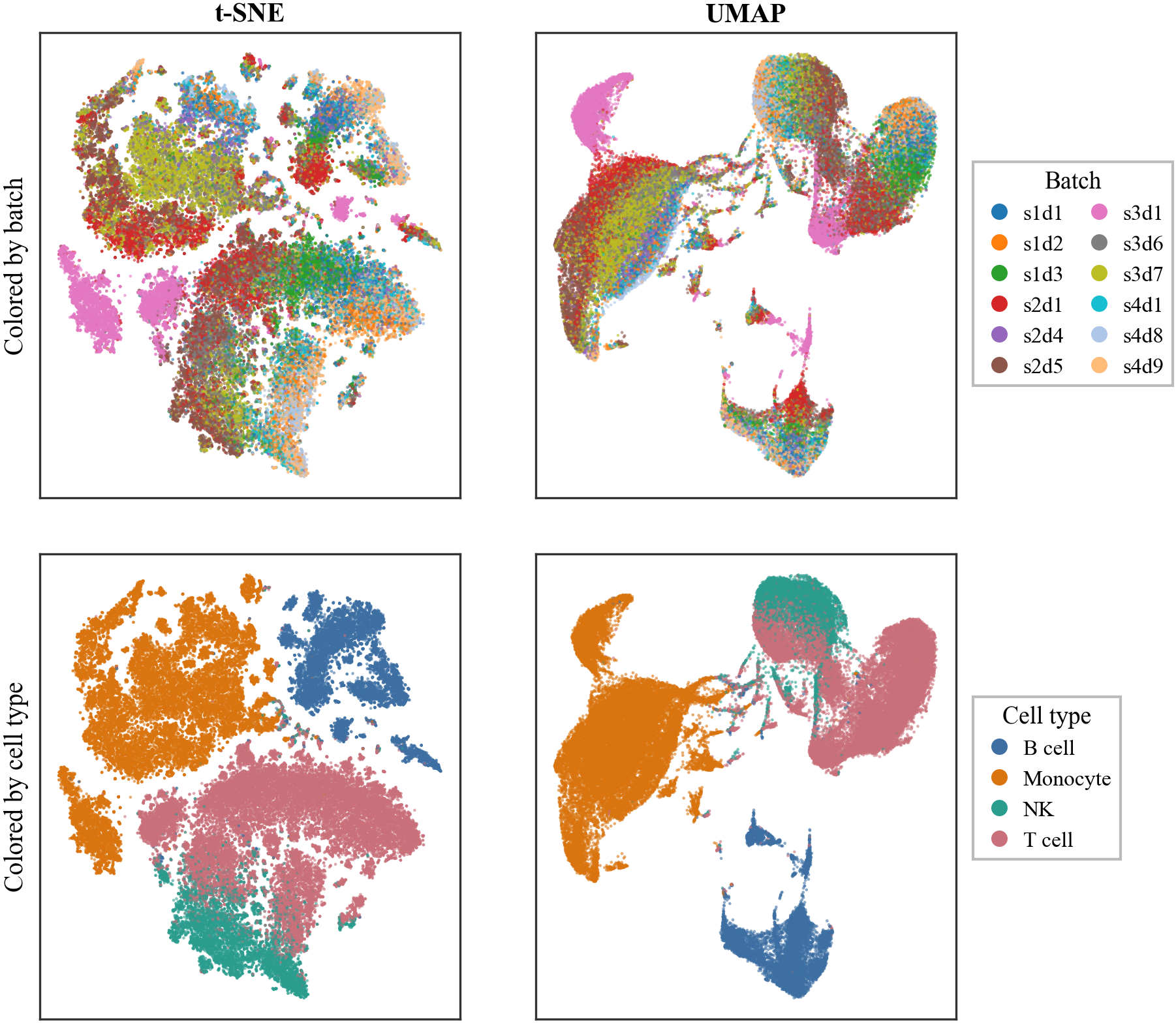}
    \caption{t-SNE and UMAP visualizations of the StablePCA embedding with subspace dimension $k=25$ on the single-cell RNA dataset.
    Top row: cells colored by 12 experimental batches.
    Bottom row: cells colored by cell types (B cell, Monocyte, NK cell, and T cell).}
    \label{fig:singlecell_tsne_umap-k25}
\end{figure}

\begin{figure}[!htbp]
    \centering
    \includegraphics[width=.66\linewidth]{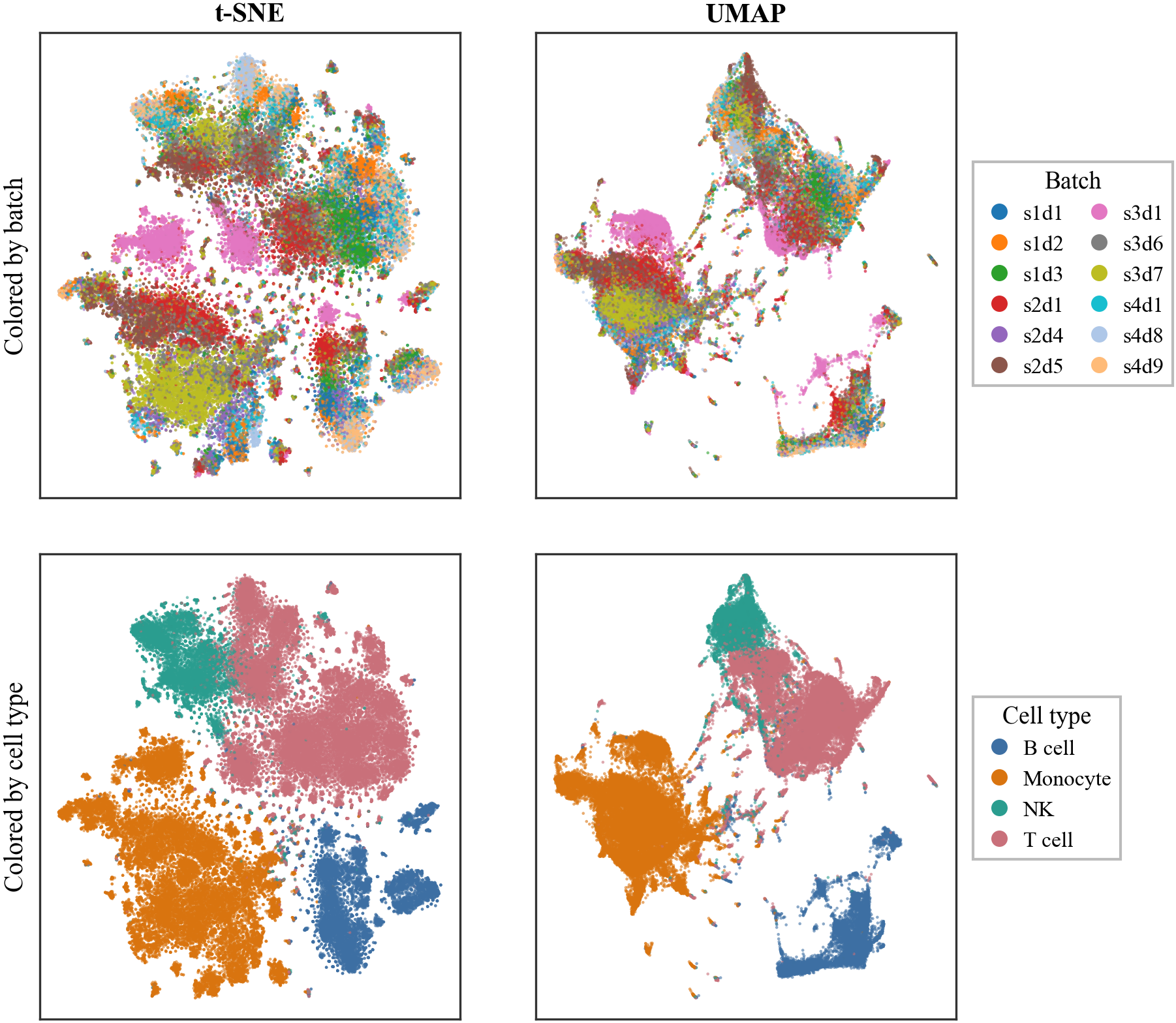}
    \caption{t-SNE and UMAP visualizations of the StablePCA embedding with subspace dimension $k=75$ on the single-cell RNA dataset.
    Top row: cells colored by 12 experimental batches.
    Bottom row: cells colored by cell types (B cell, Monocyte, NK cell, and T cell).}
    \label{fig:singlecell_tsne_umap-k75}
\end{figure}

Analogous to Figure \ref{fig:singlecell_tsne_umap} for $k=50$, we now present the visualizations of the StablePCA embedding with $k=25$ and $k=75$ in Figure \ref{fig:singlecell_tsne_umap-k25} and Figure \ref{fig:singlecell_tsne_umap-k75}, respectively. Analogous to Figure \ref{fig:singlecell_tsne_umap} for $k=50$, when $k=25$ and $k=75$, cells from 12 different experimental batches are well mixed when colored by batch (top row), and the four major cell-type populations form coherent clusters when colored by cell type (bottom row). These results demonstrate that {StablePCA} effectively removes batch-specific variations, while preserving biologically meaningful structure to distinguish major cell types.

\subsection{Omitted Experimental Setups}
\label{appendix: exp setups}

\noindent\textbf{Setup of Table \ref{tab:compare-sdp}.}
We adopt the simulation model \eqref{eq: simus} of Section~\ref{sec: simus} with $L = 4$ sources and $n=10{,}000$ samples per source, while varying the ambient dimension over $d\in\{30,50,70,100,200,300\}$. For each ambient dimension, we solve FairPCA using the proposed Mirror-Prox Algorithm~\ref{algo: mp-alter} with $T=500$ iterations. We also implement the SDP-based method following the procedure described in \citet{samadi2018price}. All experiments are repeated 100 times, independently.

\clearpage
\section{Proofs of Main Results}

\subsection{Proof of Proposition~\ref{prop: shared recovery revision}}
\label{appendix: general recovery}

Throughout the proof, we abbreviate the two second-moment matrices that appear in the margin $\Delta_l$ as
\[
\Psi_l := \EE[Z^{(l)}Z^{(l)\intercal}]\in\RR^{k\times k}
\qquad\text{and}\qquad
\Phi_l := \EE[U^{(l)}U^{(l)\intercal}]\in\RR^{q_l\times q_l},
\]
so that \eqref{eq: shared model revision} reads $\Sigma^{(l)} = V\Psi_lV^\intercal + \Gamma^{(l)}\Phi_l\Gamma^{(l)\intercal}$, the margin is $\Delta_l = \lambda_{\min}(\Psi_l) - \lambda_{\max}(\Phi_l)$, and ${\rm Tr}(\Psi_l) = \EE\|Z^{(l)}\|_2^2$.

We repeatedly use the following fact. If $A$ is symmetric and $B\succeq 0$, then
\begin{equation}
    \lambda_{\min}(A)\,{\rm Tr}(B)\ \leq\ \langle A, B\rangle\ \leq\ \lambda_{\max}(A)\,{\rm Tr}(B) .
    \label{eq: proof prop trace fact}
\end{equation}
Its proof is as follows. Writing $A = \sum_i\lambda_i(A)\,u_iu_i^\intercal$ with orthonormal $u_i$, we have $\langle A, B\rangle = \sum_i\lambda_i(A)\,u_i^\intercal Bu_i$, where every $u_i^\intercal Bu_i$ is nonnegative and $\sum_iu_i^\intercal Bu_i = {\rm Tr}(B)$.

\medskip
\noindent\textbf{Step 1: Write StablePCA as the worst-batch explained variance.} For a fixed $P$, the function $\omega\mapsto\sum_{l=1}^L\omega_l\langle\Sigma^{(l)}, P\rangle$ is linear in $\omega$, so its minimum over the simplex $\Delta^L$ is attained at a vertex, that is,
\[
\min_{\omega\in\Delta^L}\ \sum_{l=1}^L\omega_l\langle\Sigma^{(l)}, P\rangle = \min_{1\leq l\leq L}\ \langle\Sigma^{(l)}, P\rangle .
\]
Hence $P^*$ solves \eqref{eq: def diStablePCA} if and only if it maximizes $\min_{l}\langle\Sigma^{(l)}, P\rangle$ over $P\in\Pc^k$.

\medskip
\noindent\textbf{Step 2: the shared subspace explains $\EE\|Z^{(l)}\|_2^2$ in batch $l$.} Since $V^\intercal V = {\bf I}_k$ and $V^\intercal\Gamma^{(l)} = 0$ by \eqref{eq: orthogonality revision},
\[
\langle\Sigma^{(l)}, VV^\intercal\rangle = {\rm Tr}\big(V^\intercal\Sigma^{(l)}V\big) = {\rm Tr}\big(V^\intercal V\Psi_lV^\intercal V\big) + {\rm Tr}\big(V^\intercal\Gamma^{(l)}\Phi_l\Gamma^{(l)\intercal}V\big) = {\rm Tr}(\Psi_l) = \EE\|Z^{(l)}\|_2^2 .
\]
Therefore the worst-case explained variance of $VV^\intercal$ is $\min_{m}\EE\|Z^{(m)}\|_2^2$. Since $VV^\intercal\in\Pc^k$ is feasible and $P^*$ is optimal, Step~1 gives
\begin{equation}
    \min_{1\leq m\leq L}\ \langle\Sigma^{(m)}, P^*\rangle\ \geq\ \min_{1\leq m\leq L}\ \EE\|Z^{(m)}\|_2^2 .
    \label{eq: proof prop lower}
\end{equation}

\medskip
\noindent\textbf{Step 3: the mass of $P^*$ outside the shared subspace.} Define the scalar
\[
t := k - \langle P^*, VV^\intercal\rangle = \langle P^*, {\bf I}_d - VV^\intercal\rangle ,
\]
where the two expressions agree because ${\rm Tr}(P^*) = k$. Thus $t$ is the part of the $k$ units of trace of $P^*$ that lies outside ${\rm col}(V)$. As $P^*$ and $VV^\intercal$ are projection matrices of rank $k$, $\|P^*\|_F^2 = {\rm Tr}\big((P^*)^2\big) = {\rm Tr}(P^*) = k$ and likewise $\|VV^\intercal\|_F^2 = k$, so
\begin{equation}
    \|P^* - VV^\intercal\|_F^2 = \|P^*\|_F^2 + \|VV^\intercal\|_F^2 - 2\langle P^*, VV^\intercal\rangle = 2k - 2(k - t) = 2t \geq0.
    \label{eq: proof prop dist}
\end{equation}

\medskip
\noindent\textbf{Step 4: the explained variance of $P^*$ in a single batch.} Fix a batch $l$. By \eqref{eq: shared model revision} and the cyclic property of the trace,
\begin{equation}
    \langle\Sigma^{(l)}, P^*\rangle = \big\langle\Psi_l,\ V^\intercal P^*V\big\rangle + \big\langle\Phi_l,\ \Gamma^{(l)\intercal}P^*\Gamma^{(l)}\big\rangle .
    \label{eq: proof prop split}
\end{equation}
We bound the two terms separately.

For the first term, the matrix $V^\intercal P^*V$ is symmetric and satisfies $0\preceq V^\intercal P^*V\preceq V^\intercal V = {\bf I}_k$, because $0\preceq P^*\preceq{\bf I}_d$. Hence ${\bf I}_k - V^\intercal P^*V\succeq 0$, and its trace is $k - {\rm Tr}(V^\intercal P^*V) = k - \langle P^*, VV^\intercal\rangle = t$. Applying the left inequality of \eqref{eq: proof prop trace fact} with $A = \Psi_l$ and $B = {\bf I}_k - V^\intercal P^*V$,
\[
\big\langle\Psi_l,\ V^\intercal P^*V\big\rangle = {\rm Tr}(\Psi_l) - \big\langle\Psi_l,\ {\bf I}_k - V^\intercal P^*V\big\rangle\ \leq\ \EE\|Z^{(l)}\|_2^2 - \lambda_{\min}(\Psi_l)\, t .
\]
For the second term, the matrix $\Gamma^{(l)\intercal}P^*\Gamma^{(l)}$ is positive semidefinite because $P^*\succeq 0$, and its trace is $\langle P^*, \Gamma^{(l)}\Gamma^{(l)\intercal}\rangle$. Since $V$ and $\Gamma^{(l)}$ have orthonormal columns and $\Gamma^{(l)\intercal}V = 0$ by \eqref{eq: orthogonality revision}, the matrix $[V, \Gamma^{(l)}]\in\RR^{d\times(k+q_l)}$ has orthonormal columns, so $VV^\intercal + \Gamma^{(l)}\Gamma^{(l)\intercal} = [V, \Gamma^{(l)}][V, \Gamma^{(l)}]^\intercal$ is an orthogonal projection and therefore $\preceq{\bf I}_d$. Hence ${\bf I}_d - VV^\intercal - \Gamma^{(l)}\Gamma^{(l)\intercal}\succeq 0$, and $P^*\succeq 0$ gives $\langle P^*, {\bf I}_d - VV^\intercal - \Gamma^{(l)}\Gamma^{(l)\intercal}\rangle\geq 0$, i.e.,
\[
{\rm Tr}\big(\Gamma^{(l)\intercal}P^*\Gamma^{(l)}\big) = \langle P^*, \Gamma^{(l)}\Gamma^{(l)\intercal}\rangle\ \leq\ \langle P^*, {\bf I}_d - VV^\intercal\rangle = t .
\]
Applying the right inequality of \eqref{eq: proof prop trace fact} with $A = \Phi_l$ and $B = \Gamma^{(l)\intercal}P^*\Gamma^{(l)}$, and then using ${\rm Tr}(B)\leq t$ together with $\lambda_{\max}(\Phi_l)\geq 0$ (as $\Phi_l\succeq 0$),
\[
\big\langle\Phi_l,\ \Gamma^{(l)\intercal}P^*\Gamma^{(l)}\big\rangle\ \leq\ \lambda_{\max}(\Phi_l)\,{\rm Tr}\big(\Gamma^{(l)\intercal}P^*\Gamma^{(l)}\big)\ \leq\ \lambda_{\max}(\Phi_l)\, t .
\]
Adding the two bounds in \eqref{eq: proof prop split} and recalling $\Delta_l = \lambda_{\min}(\Psi_l) - \lambda_{\max}(\Phi_l)$, we obtain
\begin{equation}
    \langle\Sigma^{(l)}, P^*\rangle\ \leq\ \EE\|Z^{(l)}\|_2^2 - \Delta_l\, t\qquad\text{for every batch } l .
    \label{eq: proof prop key}
\end{equation}

\medskip
\noindent\textbf{Step 5: the inequality \eqref{eq: margin equal revision}.} Let $l$ be a batch with $\Delta_l > 0$. The worst-case explained variance of $P^*$ is at most its explained variance in batch $l$, so \eqref{eq: proof prop lower} and \eqref{eq: proof prop key} give
\[
\min_{1\leq m\leq L}\EE\|Z^{(m)}\|_2^2\ \leq\ \min_{1\leq m\leq L}\langle\Sigma^{(m)}, P^*\rangle\ \leq\ \langle\Sigma^{(l)}, P^*\rangle\ \leq\ \EE\|Z^{(l)}\|_2^2 - \Delta_l\, t .
\]
Rearranging, $\Delta_l\, t\leq\EE\|Z^{(l)}\|_2^2 - \min_{m}\EE\|Z^{(m)}\|_2^2$. Dividing by $\Delta_l > 0$ and using \eqref{eq: proof prop dist} yields \eqref{eq: margin equal revision}.

\medskip
\noindent\textbf{Step 6: exact recovery.} Suppose $s_1=\cdots=s_L$ and $\Delta_l>0$ for some source $l$. Applying Step~5 to that source, the right-hand side of \eqref{eq: margin equal revision} equals zero, so $\|P^* - VV^\intercal\|_F = 0$, that is, $P^* = VV^\intercal$. This holds for every solution $P^*$ of \eqref{eq: def diStablePCA}, and a solution exists because the objective $\min_l\langle\Sigma^{(l)}, P\rangle$ is continuous on the compact set $\Pc^k$. Hence $VV^\intercal$ is the unique solution.

\smallskip
\noindent Finally, the same argument applies to the relaxed StablePCA \eqref{eq: fantope MPCA}, with $P^*$ replaced by any solution $M^*\in\Fc^k$. Steps~1, 2, 4, and~5 use only $0\preceq M^*\preceq{\bf I}_d$, ${\rm Tr}(M^*) = k$, and the feasibility of $VV^\intercal\in\Pc^k\subseteq\Fc^k$; Step~6 uses only that $\Fc^k$ is compact; and in Step~3 the identity \eqref{eq: proof prop dist} becomes the inequality $\|M^* - VV^\intercal\|_F^2\leq 2t$, because $\|M^*\|_F^2 = {\rm Tr}\big((M^*)^2\big)\leq{\rm Tr}(M^*) = k$ when $0\preceq M^*\preceq{\bf I}_d$. Hence \eqref{eq: margin equal revision} holds for every solution of \eqref{eq: fantope MPCA}, and when $\Delta_{l^*}>0$ for some $l^*\in\argmin_l s_l$, $VV^\intercal$ is also the unique solution of \eqref{eq: fantope MPCA}. \hfill$\square$

\subsection{Proof of Proposition \ref{prop: closed form revised}}
\label{proof of prop: closed form}

We mainly focus on the proof of the first-step update in \eqref{eq: mirror middle update}, while similar arguments follow for the second-step update in \eqref{eq: mirror final update}.

Recall that in \eqref{eq: finite sample}, we define
\[
f(M, \omega) = -\sum_{l=1}^L\omega_l\cdot \left\langle \widehat{\Sigma}^{(l)}, M\right\rangle.
\]
Its gradients over $M$ and $\omega$ are expressed as:
\[
\nabla_Mf(M, \omega) = -\sum_{l=1}^L \omega_l\widehat{\Sigma}^{(l)} \in \RR^{d\times d}, \quad \textrm{and}\quad \nabla_\omega f(M, \omega) = \Big(
-\langle \widehat\Sigma^{(1)},M\rangle,\ldots,
-\langle \widehat\Sigma^{(L)},M\rangle
\Big)^\intercal \in \RR^{L}.
\]

We write the optimization problems in \eqref{eq: mirror middle update} as follows:
\begin{equation}
\begin{aligned}
    \omega^{t+\frac{1}{2}}&= \argmin_{\omega\in \Delta^{L}} -\eta_{\omega} \langle \nabla_\omega f(M^t, \omega^t), \; \omega\rangle + D_{\psi_2}(\omega, \omega^t)
    \\
    &=\argmin_{\omega\in \Delta^{L}}\eta_\omega\cdot \sum_{l=1}^{L}\omega_l \left\langle \widehat{\Sigma}^{(l)}, {M}^{t} \right\rangle+\sum_{l=1}^{L}\omega_l \log \frac{\omega_l}{{\omega}^{t}_l},
\end{aligned}
\label{eq: mirror middle 1}
\end{equation}
and 
\begin{equation}
\begin{aligned}
M^{t+\frac{1}{2}} &= \argmin_{M\in \mathcal{F}^{k}} \eta_M \langle \nabla_M \widehat{f}(M^t, \omega^t), M\rangle + D_{\psi_1}(M, M^t)\\
&=\argmin_{M\in \mathcal{F}^{k}} -\eta_M \cdot \left\langle \sum_{l=1}^{L}\omega^{t}_l  \widehat{\Sigma}^{(l)}, M\right\rangle+ {\rm Tr}\left(M\left(\log M-\log  M^{t}\right)\right)\\
&=\argmin_{M\in \mathcal{F}^{k}} \left\langle-\eta_M \sum_{l=1}^{L}\omega^{t}_l  \widehat{\Sigma}^{(l)}-\log M^{t},\; M\right\rangle+ {\rm Tr}\left(M\log M\right)%
\end{aligned}
\label{eq: mirror middle 2}
\end{equation}
In the following discussions, we present the closed-form expressions of $\omega$-update and $M$-update in order.

\vspace{0.5em}
\noindent\textbf{Closed-form expressions of ${\omega}^{t+\frac{1}{2}}$ and $\omega^{t+1}$.}
We derive the closed-form expression of $\omega^{t+\frac12}$, while the update for $\omega^{t+1}$ follows by the same argument with $M^t$ replaced by $M^{t+\frac12}$.

For notation convenience, we define 
$$h^{(l)}=\eta_\omega \langle \widehat{\Sigma}^{(l)}, {M}^{t} \rangle\quad \textrm{for $l=1,\dots,L$}.$$
Then the optimization problem in \eqref{eq: mirror middle 1} can be written as
\begin{equation}
\omega^{t+\frac{1}{2}} =\argmin_{\omega \in \Delta^L}
\;\sum_{l=1}^L \omega_l h^{(l)}
+ \sum_{l=1}^L \omega_l \log \frac{\omega_l}{\omega_l^t}.
\label{eq: omega subproblem}
\end{equation}

This is a strictly convex problem over the probability simplex, hence it admits
a unique minimizer. Introducing a Lagrange multiplier $\mu \in \mathbb{R}$ for
the simplex constraint $\sum_{l=1}^L \omega_l = 1$ and multipliers
$\lambda_l \ge 0$ for the nonnegativity constraints $\omega_l \ge 0$, the
Lagrangian is
\[
\mathcal{L}(\omega, \mu, \lambda)
= \sum_{l=1}^L \omega_l h^{(l)}
+ \sum_{l=1}^L \omega_l \log \frac{\omega_l}{\omega_l^t}
+ \mu\!\left(1 - \sum_{l=1}^L \omega_l\right)
- \sum_{l=1}^L \lambda_l \omega_l .
\]
The Karush–Kuhn–Tucker (KKT) conditions are given by
\begin{align}
h^{(l)} + \log \frac{\omega_l^{t+\frac12}}{\omega_l^t} + 1 - \mu - \lambda_l &= 0,
\quad l=1,\dots,L, \label{eq:kkt_stationarity} \\
\sum_{l=1}^L \omega_l^{t+\frac12} &= 1, \label{eq:kkt_simplex} \\
\omega_l^{t+\frac12} \ge 0,\;\; \lambda_l \ge 0,\;\;
\lambda_l \omega_l^{t+\frac12} &= 0, \quad l=1,\dots,L. \label{eq:kkt_complementary}
\end{align}

Since $\omega^t$ has strictly positive entries, the derivative of $\omega_l\log(\omega_l/\omega_l^t)$ tends to $-\infty$ as $\omega_l\downarrow0$. Moving a small amount of mass from a positive coordinate to a zero coordinate therefore decreases the objective, so the minimizer satisfies $\omega_l^{t+\frac12}>0$ for all $l$. Thus $\lambda_l=0$ by \eqref{eq:kkt_complementary}. Absorbing the constant $1$ into $\mu$, \eqref{eq:kkt_stationarity} becomes
\[
\log \frac{\omega_l^{t+\frac12}}{\omega_l^t}
= \mu - h^{(l)}, \quad l=1,\dots,L.
\]
Exponentiating both sides yields
\[
\omega_l^{t+\frac12}
= \omega_l^t \exp(\mu - h^{(l)}), \quad l=1,\dots,L.
\]

Finally, enforcing the normalization condition \eqref{eq:kkt_simplex} gives
\[
1 = \sum_{l=1}^L \omega_l^{t+\frac12}
= \exp(\mu) \sum_{l=1}^L \omega_l^t \exp(-h^{(l)}),
\]
which implies
\[
\exp(\mu)
= \left(\sum_{s=1}^L \omega_s^t
\exp\!\left(-\eta_\omega \langle \widehat{\Sigma}^{(s)}, M^t \rangle\right)
\right)^{-1}.
\]
Substituting this expression back yields
\[
\omega_l^{t+\frac12}
= \frac{\omega_l^t
\exp\!\left(-\eta_\omega \langle \widehat{\Sigma}^{(l)}, M^t \rangle\right)}
{\sum_{s=1}^L \omega_s^t
\exp\!\left(-\eta_\omega \langle \widehat{\Sigma}^{(s)}, M^t \rangle\right)},
\quad l=1,\dots,L,
\]
which proves the first $\omega$-update formula in
\eqref{eq: omega update 1}. The expression for $\omega^{t+1}$ in
\eqref{eq: omega update 2} follows from an identical argument with
$M^t$ replaced by $M^{t+\frac12}$.

\vspace{0.5em}
\noindent\textbf{Closed-form expressions of ${M}^{t+\frac{1}{2}}$ and $M^{t+1}$.}
We derive the closed-form expression of $M^{t+\frac{1}{2}}$, while the update for
$M^{t+1}$ follows by the same argument with $\omega^t$ replaced by
$\omega^{t+\frac{1}{2}}$.

For notation convenience, we define
\[
A^t \coloneqq \log M^t + \eta_M \sum_{l=1}^L \omega_l^t \widehat{\Sigma}^{(l)} .
\]
Then the optimization problem in \eqref{eq: mirror middle 2} is equivalently written as
\begin{equation}
M^{t+\frac{1}{2}}
= \argmin_{M\in \mathcal{F}^k}
\; \langle -A^t, M\rangle + {\rm Tr}(M\log M).
\label{eq:M_simplified}
\end{equation}
The proof proceeds in two steps:
\begin{enumerate}
    \item[(Step-1)] We show that $M^{t+\frac{1}{2}}$ shares the same eigenvectors as $A^t$. Therefore, the optimization problem \eqref{eq:M_simplified} reduces to identifying the eigenvalues of $M^{t+\frac{1}{2}}$;
    \item[(Step-2)] We solve the eigenvalue optimization problem for $M^{t+\frac{1}{2}}$.
\end{enumerate}

We first show that the minimizer $M^{t+\frac12}$ of \eqref{eq:M_simplified} must share the same
eigenvectors as $A^t$.
We recall the following classical trace inequality.

\begin{Lemma}[von Neumann trace inequality]
Let $A,B\in\mathbb{R}^{d\times d}$ be symmetric matrices with eigenvalues
$\lambda_1(\cdot)\ge \cdots \ge \lambda_d(\cdot)$. Then
\[
\sum_{j=1}^d \lambda_j(A)\lambda_{d+1-j}(B)
\;\le\;
\Tr(AB)
\;\le\;
\sum_{j=1}^d \lambda_j(A)\lambda_j(B),
\]
where equality in the upper (lower) bound holds if and only if $A$ and $B$ admit a common orthonormal eigenbasis with their eigenvalues paired in the same (opposite) order.
\label{lemma: trace inequality}
\end{Lemma}

Applying the upper bound with $A=A^t$ and $B=M$ and multiplying by $-1$, we obtain, for any $M\in\mathcal{F}^k$,
\[
\langle -A^t, M\rangle
\;\ge\;
-\sum_{j=1}^d \lambda_j(A^t)\lambda_j(M),
\]
with equality if and only if $M$ and $A^t$ admit a common orthonormal eigenbasis with their eigenvalues paired in the same order.
Moreover, since
\[
\Tr(M\log M) = \sum_{j=1}^d \lambda_j(M)\log\lambda_j(M),
\]
the entropy term depends only on the eigenvalues of $M$.
Therefore, any minimizer of \eqref{eq:M_simplified} must be co-diagonalizable
with $A^t$.

Let the eigen decomposition of $A^t$ be
\[
A^t = U^t \Diag(\lambda^t_1,\dots,\lambda^t_d)(U^t)^\top,
\qquad \lambda^t_1\ge \cdots \ge \lambda^t_d.
\]
Then $M^{t+\frac{1}{2}}$ must take the form
\begin{equation}
M^{t+\frac{1}{2}}
= U^t \Diag(\xi_1,\dots,\xi_d)(U^t)^\top,
\quad
\xi \in \mathcal{D}^k
:= \left\{\xi\in\mathbb{R}^d:\; 0\le \xi_j\le 1,\ \sum_{j=1}^d \xi_j = k \right\},
\label{eq:M_eigendecomp}
\end{equation}
where $\xi\in \Dc^k$ because $M^{t+\frac{1}{2}}\in \Fc^k$.

Substituting \eqref{eq:M_eigendecomp} into \eqref{eq:M_simplified}, the problem
reduces to
\begin{equation}
\min_{\xi\in\mathcal{D}^k}
\; -\sum_{j=1}^d \lambda_j^t \xi_j
+ \sum_{j=1}^d \xi_j \log \xi_j .
\label{eq:xi_problem}
\end{equation}
This is a strictly convex optimization problem and hence admits a unique
minimizer.

We introduce a Lagrange multiplier $\alpha\in\mathbb{R}$ for the constraint
$\sum_{j=1}^d \xi_j = k$, and multipliers $\tau_j\ge 0$ for the constraints
$\xi_j\le 1$. (The nonnegativity constraints $\xi_j\ge 0$ are inactive since the
entropy term enforces $\xi_j>0$ at the optimum.)
The Lagrangian is
\[
\mathcal{L}(\xi,\alpha,\tau)
= -\sum_{j=1}^d \lambda_j^t \xi_j
+ \sum_{j=1}^d \xi_j \log \xi_j
+ \alpha\!\left(k-\sum_{j=1}^d \xi_j\right)
+ \sum_{j=1}^d \tau_j(\xi_j-1).
\]

The KKT stationarity condition yields
\[
-\lambda_j^t + \log \xi_j + 1 - \alpha + \tau_j = 0,
\qquad j=1,\dots,d,
\]
which implies
\begin{equation}
\xi_j = \exp(\lambda_j^t + \alpha - 1 - \tau_j).
\label{eq:xi_exp}
\end{equation}
The complementary slackness condition $\tau_j(\xi_j-1)=0$ implies the following:
\begin{itemize}
\item if $\exp(\lambda_j^t + \alpha - 1)\le 1$, then $\tau_j=0$ and
      $\xi_j=\exp(\lambda_j^t + \alpha - 1)$;
\item if $\exp(\lambda_j^t + \alpha - 1)>1$, then $\tau_j>0$ and $\xi_j=1$.
\end{itemize}
Combining both cases, we obtain
\begin{equation}
\xi_j
= \min\!\left\{ \exp(\lambda_j^t + \alpha - 1),\; 1 \right\},
\qquad j=1,\dots,d.
\label{eq:xi_solution}
\end{equation}

Finally, the scalar $\alpha$ is uniquely determined by the constraint in \eqref{eq:M_eigendecomp}
\[
\sum_{j=1}^d\xi_j = \sum_{j=1}^d \min\!\left\{ \exp(\lambda_j^t + \alpha - 1),\; 1 \right\} = k.
\]
Substituting \eqref{eq:xi_solution} into \eqref{eq:M_eigendecomp} yields the
closed-form expression of $M^{t+\frac{1}{2}}$ in \eqref{eq: omega update 1}.
The formula for $M^{t+1}$ in \eqref{eq: omega update 2} follows analogously.

\subsection{Proof of Proposition \ref{cor: tightness sufficient}}
\label{proof of prop: instances}
We use two facts. First, $\phi(\omega)\geq\langle \Sigma(\omega), P\rangle$ for every $\omega\in\Delta^L$ and $P\in\Pc^k$, because $\Pc^k\subseteq\Fc^k$. Second, a linear function of $\omega$ is minimized over the simplex at a vertex, so that $\min_\omega\langle \Sigma(\omega), P\rangle = \min_l\langle \Sigma^{(l)}, P\rangle$. In each case we exhibit a vertex $e_{l}$ of the simplex that is a least favorable mixture and whose matrix $\Sigma(e_l) = \Sigma^{(l)}$ has the stated eigengap.

\medskip
\noindent\textbf{Case (a).} Choose a source $l$ with $\Delta_l>0$. Under \eqref{eq: shared model revision}, $\Sigma^{(l)}$ has top-$k$ eigenspace ${\rm col}(V)$ and eigengap $\Delta_l$. By Ky Fan's principle, $\phi(e_l)=s_l$. Since $s_1=\cdots=s_L$, for every $\omega\in\Delta^L$,
\[
\phi(\omega)\geq\langle\Sigma(\omega),VV^\intercal\rangle
=\sum_{m=1}^L\omega_m s_m=s_l=\phi(e_l).
\]
Thus, $\omega^*=e_l$ is a least favorable mixture with eigengap $\delta^*=\Delta_l>0$.

\medskip
\noindent\textbf{Case (b).} For every $\omega\in\Delta^L$, $\phi(\omega)\geq\langle\Sigma(\omega), P_{l_0}\rangle = \sum_l\omega_l\langle\Sigma^{(l)}, P_{l_0}\rangle\geq\langle\Sigma^{(l_0)}, P_{l_0}\rangle = \phi(e_{l_0})$, where the first inequality is the first fact, the second uses the assumption on $P_{l_0}$, and the last equality is Ky Fan's principle for $\Sigma^{(l_0)}$. As a result, $e_{l_0}$ is a least favorable mixture, and $\Sigma(e_{l_0}) = \Sigma^{(l_0)}$ has an eigengap by assumption. If $\Sigma^{(l)}\succeq\Sigma^{(l_0)}$ for all $l$, then $\langle\Sigma^{(l)} - \Sigma^{(l_0)}, P_{l_0}\rangle\geq 0$ because $P_{l_0}\succeq 0$, so the assumption on $P_{l_0}$ holds.

\medskip
\noindent\textbf{Case (c).} Let $g_l := \lambda_k(\Sigma^{(l)}) - \lambda_{k+1}(\Sigma^{(l)}) > 0$. Since $P$ is the projection onto the top-$k$ eigenspace of $\Sigma^{(l)}$ for every batch $l$, it holds that $P$ is also the projection onto the top-$k$
 eigenspace of $\Sigma(\omega)$ for every $\omega\in \Delta^L$, with eigengap at least $\sum_l\omega_lg_l\geq \min_l g_l$. By Ky Fan's principle
$\phi(\omega) = \langle \Sigma(\omega), P\rangle = \sum_l \omega_l c = c$ for every $\omega\in \Delta^L$, $\phi$ is constant. It implies that every $\omega$ minimizes $\phi(\cdot)$, with eigengap at least $\min_l g_l>0$.

\subsection{Proof of Theorem \ref{thm: fantope tight}}
\label{proof of thm: fantope tight}
We use two facts. First, by Sion's theorem and \eqref{eq: omega_star}, the optimal value of the relaxed problem \eqref{eq: fantope MPCA} equals $\min_{\omega\in\Delta^L}\phi(\omega) = \phi(\omega^*)$ for every least favorable mixture $\omega^*$. Second, for any $M\in\Fc^k$, the function $\omega\mapsto\langle\Sigma(\omega), M\rangle$ is linear on the simplex, so its minimum is attained at a vertex: $\min_{\omega\in\Delta^L}\langle \Sigma(\omega), M\rangle = \min_{1\leq l\leq L}\langle \Sigma^{(l)}, M\rangle$.

\medskip
\noindent\textbf{Step 1: every relaxed solution maximizes $\langle \Sigma(\omega^*), \cdot\rangle$ over $\Fc^k$.} 

Let $M^*$ solve \eqref{eq: fantope MPCA} and let $\omega^*$ be any least favorable mixture. By the first fact, $\min_\omega\langle \Sigma(\omega), M^*\rangle = \phi(\omega^*)$, and since $\omega^*$ is one feasible mixture,
\[
\langle \Sigma(\omega^*), M^*\rangle\ \geq\ \min_{\omega\in\Delta^L}\langle \Sigma(\omega), M^*\rangle\ =\ \phi(\omega^*)\ =\ \max_{M\in\Fc^k}\langle \Sigma(\omega^*), M\rangle .
\]
Hence $M^*$ maximizes $\langle \Sigma(\omega^*), M\rangle$ over $\Fc^k$, for \emph{\textbf{every}} $\omega^*$. 

\medskip
\noindent\textbf{Step 2: under an eigengap, the maximizer is unique.} 

Now let $\omega^*$ satisfy Condition~\ref{cond: eigengap}, write $\Sigma(\omega^*) = \sum_{i=1}^d \lambda_i v_i v_i^\intercal$ with $\lambda_1\geq\dots\geq\lambda_d$ and orthonormal $v_1,\dots,v_d$, so that $\lambda_k > \lambda_{k+1}$, and let $P^* := \sum_{i\leq k}v_iv_i^\intercal$ be the projection onto the top-$k$ eigenspace of $\Sigma(\omega^*)$, which is well defined because of the eigengap. 

For $M\in\Fc^k$, set $m_i := v_i^\intercal M v_i$. Then $m_i\in[0,1]$ because $0\preceq M\preceq {\bf I}_d$, and $\sum_i m_i = {\rm Tr}(M) = k$. Hence
\begin{equation}
    \langle\Sigma(\omega^*), M\rangle = \sum_{i=1}^d \lambda_i m_i \leq \sum_{i\leq k}\lambda_i = \phi(\omega^*),
    \label{eq:temp-thm-fantope-tight}
\end{equation}
which is Ky Fan's maximum principle, and since $\lambda_k > \lambda_{k+1}$, equality holds if and only if $m_i = 1$ for $i\leq k$ and $m_i = 0$ for $i > k$. 

Note that $m_i=1$, together with ${\bf I}_d - M\succeq 0$, forces $({\bf I}_d - M)v_i = 0$, i.e., $Mv_i = v_i$ for $i\leq k$, because $0 = 1-m_i = v_i^\intercal ({\bf I}_d - M) v_i\geq 0$. Similarly, $m_i=0$, together with $M\succeq 0$, forces $M v_i = 0$ for $i>k$. 
With orthonormal $v_1,...,v_d$, it holds
$$
M = M \mathbf{I}_d = M \sum_{i=1}^d v_i v_i^\intercal = \sum_{i\leq k} (Mv_i) v_i^\intercal + \sum_{i>k} (Mv_i) v_i^\intercal = \sum_{i\leq k} v_i v_i^\intercal = P^*.
$$
Therefore, the equality in \eqref{eq:temp-thm-fantope-tight} holds if and only if $M=P^*$. That is, the projection $P^*$ is the unique maximizer of $\langle \Sigma(\omega^*), M\rangle$ over $\Fc^k$.

\medskip
\noindent\textbf{Step 3: uniqueness and tightness.} 

By Steps 1 and 2, every solution of the relaxed StablePCA \eqref{eq: fantope MPCA} equals $P^*$. A solution exists, because \eqref{eq: fantope MPCA} maximizes a continuous function over a compact set, so $P^*$ is the unique solution of the relaxed StablePCA \eqref{eq: fantope MPCA}, with optimal value $\phi(\omega^*) = \sum_{i\leq k}\lambda_i(\Sigma(\omega^*))$ by the first fact. 

Since $P^*\in\Pc^k\subseteq\Fc^k$, the optimal value of the original StablePCA \eqref{eq: def diStablePCA} must lie between $\min_\omega\langle\Sigma(\omega), P^*\rangle$ and the relaxed optimal value,
\[
\min_{\omega\in \Delta^L}\langle\Sigma(\omega), P^*\rangle \leq \max_{P\in \Pc^k}\min_{\omega\in \Delta^L}\langle\Sigma(\omega), P\rangle \leq \max_{M\in \Fc^k}\min_{\omega\in \Delta^L}\langle\Sigma(\omega), M\rangle=\min_{\omega\in \Delta^L}\langle \Sigma(\omega), P^*\rangle.
\]
Therefore, all quantities in \eqref{eq: tight values} coincide
\[
\phi(\omega^*) =\max_{P\in \Pc^k}\min_{\omega\in \Delta^L}\langle\Sigma(\omega), P\rangle =\max_{M\in \Fc^k}\min_{\omega\in \Delta^L}\langle\Sigma(\omega), M\rangle=\min_{\omega\in \Delta^L}\langle \Sigma(\omega), P^*\rangle = \min_{l}\langle\Sigma^{(l)}, P^*\rangle,
\]
where the last equality is due to the second fact.

Note that every solution of the original StablePCA \eqref{eq: def diStablePCA} attains the common value, hence solves the relaxed StablePCA \eqref{eq: fantope MPCA}. And we have established that the solution of the relaxed StablePCA is $P^*$ uniquely, therefore the solution of the original StabelPCA is unique and equals $P^*$.

\medskip
\noindent\textbf{Step 4: independence of the choice of $\omega^*$.} 

In Step 3, we have shown that every solution of the relaxed StablePCA equals $P^*$, the projection onto the top-$k$ eigenspace of $\Sigma(\omega^*)$. 

Suppose that $\omega^{**}$ is another least favorable mixture with an eigengap, Steps 2 and 3 applied to $\omega^{**}$ show that the projection onto the top-$k$ eigenspace of $\Sigma(\omega^{**})$ is the unique solution of the relaxed StablePCA. This implies that all least favorable mixtures with an eigengap share the same top-$k$ eigenspace.

\subsection{Proof of Theorem \ref{thm: conv rates alg}}
\label{proof of thm: conv rates alg}

We prove the optimization-error bound of Theorem~\ref{thm: conv rates alg} in Section~\ref{sec: theory}. To this end, we introduce the following Theorem, which establishes the primal-dual gap of the output $(\widehat{M}_T, \widehat{\omega}_T)$ of Mirror-Prox algorithm \ref{algo: mp}. The proof is provided in Appendix \ref{proof of thm: conv rates alg saddle}.

\begin{Theorem}[Primal-dual gap of Mirror-Prox]
\label{thm: conv rates alg saddle}
With the step sizes $\eta_M,\eta_\omega$ specified in Theorem \ref{thm: conv rates alg},
the primal-dual gap of the output {$(\widehat{M}_T, \widehat{\omega}_T)$} of  Algorithm \ref{algo: mp} satisfies
\begin{equation}
\label{eq: saddle output}
    \max_{M\in \Fc^k} \left\langle\widehat{\Sigma}({\widehat{\omega}_T}), M\right\rangle - \min_{\omega\in \Delta^L}\left\langle\widehat{\Sigma}(\omega), {\widehat{M}_T}\right\rangle \leq  \max_{l\in [L]}\|\widehat{\Sigma}^{(l)}\|_{\rm op}\cdot \frac{8k\sqrt{k\log(d/k)\log L}}{T}.
\end{equation}
\end{Theorem}
This result implies that the output ${(\widehat{M}_T, \widehat{\omega}_T)}$ of Algorithm \ref{algo: mp} is an $\mathcal{O}(T^{-1})$-nearly saddle point of the empirical Fantope-relaxed StablePCA objective \eqref{eq: finite sample}; see Definition \ref{def: saddle point}. Moreover, since $\widehat{\omega}_T\in \Delta^L$, it holds
\[
\max_{M \in \Fc^k} \min_{\omega \in \Delta^L}\left\langle \widehat{\Sigma}(\omega), M \right\rangle \leq \max_{M\in \Fc^k}\left\langle\widehat{\Sigma}({\widehat{\omega}_T}), M\right\rangle.
\]
Combining the inequality above with~\eqref{eq: saddle output} yields the following inequality:
\[
\begin{aligned}
    \max_{M \in \Fc^k} \min_{\omega \in \Delta^L}\left\langle \widehat{\Sigma}(\omega), M \right\rangle - \min_{\omega\in \Delta^L}\left\langle\widehat{\Sigma}(\omega), {\widehat{M}_T}\right\rangle 
    &\leq \max_{M\in \Fc^k}\left\langle\widehat{\Sigma}({\widehat{\omega}_T}), M\right\rangle - \min_{\omega\in \Delta^L}\left\langle\widehat{\Sigma}(\omega), {\widehat{M}_T}\right\rangle \\
    &\leq \max_{l\in [L]}\|\widehat{\Sigma}^{(l)}\|_{\rm op}\cdot \frac{8k\sqrt{k\log(d/k)\log L}}{T},
\end{aligned}
\]
which completes the proof.

\subsection{Proof of Theorem \ref{thm: new conv}}
\label{proof of thm: new conv}
Throughout this proof and the next, $C$ denotes a constant depending only on the sub-Gaussian parameter $\sigma$ of Condition~\ref{cond: subgauss}, whose value may change from line to line, and we recall that $n = \min_l n_l$ and that $\mathbf{r}_{\max}$ is the effective rank in \eqref{eq: effective rank}.

For any $M\in \Fc^k,\omega\in \Delta^L$, it holds that
\[
\left|\langle \widehat{\Sigma}(\omega), M\rangle - \langle\Sigma(\omega), M\rangle\right| = \left|\sum_{l=1}^L \omega_l \cdot \langle \widehat{\Sigma}^{(l)} - \Sigma^{(l)}, M\rangle\right| \leq \sup_{l\in [L]}\left|\langle\widehat{\Sigma}^{(l)} - \Sigma^{(l)}, M\rangle\right|.
\]

To control the right-hand side of the above inequality uniformly over $M\in \Fc^k$, we invoke the following lemma, which establishes a uniform concentration inequality that bounds the discrepancy between the empirical and population covariance matrices over $\Fc^k$. The proof is provided in Appendix \ref{proof of lemma: obj val diff}. Its rate is governed by the effective rank rather than by the ambient dimension.
    \begin{Lemma}
    \label{lemma: obj val diff}
        Suppose Condition \ref{cond: subgauss} holds. Then for any $t\geq 0$, with probability at least $1-2L e^{-t}$:
        \[
        \max_{l\in [L]}\left\|\widehat{\Sigma}^{(l)} - \Sigma^{(l)}\right\|_{\rm op} \leq C_\sigma \max_{l\in [L]}\|\Sigma^{(l)}\|_{\rm op} \cdot \left(\sqrt{\frac{\mathbf{r}_{\max}+t}{n}} + \frac{\mathbf{r}_{\max}+t}{n}\right),
        \]
        and consequently
        \[
        \max_{l\in [L]}\left|\|\widehat{\Sigma}^{(l)}\|_{\rm op} - \|\Sigma^{(l)}\|_{\rm op}\right| \leq C_\sigma \max_{l\in [L]}\|\Sigma^{(l)}\|_{\rm op} \cdot \left(\sqrt{\frac{\mathbf{r}_{\max}+t}{n}} + \frac{\mathbf{r}_{\max}+t}{n}\right),
        \]
        \[
        \max_{l\in [L]}\sup_{M\in \Fc^k}\left|\langle\widehat{\Sigma}^{(l)} - \Sigma^{(l)}, M\rangle\right| \leq C_\sigma \max_{l\in [L]}\|\Sigma^{(l)}\|_{\rm op} \cdot k \left(\sqrt{\frac{\mathbf{r}_{\max}+t}{n}} + \frac{\mathbf{r}_{\max}+t}{n}\right),
        \]
        where $\mathbf{r}_{\max}$ is the effective rank in \eqref{eq: effective rank} and $C_\sigma = C\max\{\sigma^2,\sigma^4\}$ for a universal constant $C>0$, so that $C_\sigma$ depends only on $\sigma$.
    \end{Lemma}

Then we apply Lemma \ref{lemma: obj val diff} to conclude that with probability at least $1-2L e^{-t}$,
\begin{equation*}
    \forall M\in \Fc^k, \omega\in \Delta^L, \quad \left|\langle \widehat{\Sigma}(\omega), M\rangle - \langle\Sigma(\omega), M\rangle\right|\leq C \max_{l\in [L]}\|\Sigma^{(l)}\|_{\rm op} \cdot k \left(\sqrt{\frac{\mathbf{r}_{\max}+t}{n}} + \frac{\mathbf{r}_{\max}+t}{n}\right). 
    \label{eq: element wise diff}
\end{equation*}
Therefore, with high probability at least $1-2L e^{-t}$, it holds $\forall M\in \Fc^k$,
\begin{equation} 
    \begin{aligned}
        \left|\min_{\omega\in \Delta^L} \langle \widehat{\Sigma}(\omega), M\rangle - \min_{\omega\in \Delta^L}\langle\Sigma(\omega), M\rangle\right|
        &\leq \max_{\omega\in \Delta^L}\left|\langle \widehat{\Sigma}(\omega), M\rangle - \langle\Sigma(\omega), M\rangle\right|\\
        &\leq  C \max_{l\in [L]}\|\Sigma^{(l)}\|_{\rm op} \cdot k \left(\sqrt{\frac{\mathbf{r}_{\max}+t}{n}} + \frac{\mathbf{r}_{\max}+t}{n}\right).
    \end{aligned}
    \label{eq: max diff}
\end{equation}

We now turn to compare the output $\widehat{M}_T$ of Algorithm \ref{algo: mp} to the population optimum. Recall that $M^*$ denotes a population StablePCA solution in \eqref{eq: fantope MPCA}, i.e.,
\[
M^* \in \argmax_{M\in \Fc^k}\min_{\omega\in\Delta^L} \langle\Sigma(\omega), M\rangle.
\]
With probability at least $1-2Le^{-t}$, we proceed as follows:
\[
\begin{aligned}
    \min_{\omega\in\Delta^L}
\langle \Sigma(\omega), \widehat{M}_T\rangle
&\ge
\min_{\omega\in\Delta^L}
\langle \widehat{\Sigma}(\omega), \widehat{M}_T\rangle
-
C k\!\left(
\sqrt{\frac{\mathbf{r}_{\max}+t}{n}}
+
\frac{\mathbf{r}_{\max}+t}{n}
\right)
\max_{l}\|\Sigma^{(l)}\|_{\rm op}
&& \text{(by \eqref{eq: max diff})} \\[0.3em]
&\ge
\max_{M\in\Fc^k}
\min_{\omega\in\Delta^L}
\langle \widehat{\Sigma}(\omega), M\rangle
-
\frac{8k\sqrt{k\log(d/k)\log L}}{T}
\max_{l}\|\widehat{\Sigma}^{(l)}\|_{\rm op} \\
&\quad
-
C k\!\left(
\sqrt{\frac{\mathbf{r}_{\max}+t}{n}}
+
\frac{\mathbf{r}_{\max}+t}{n}
\right)
\max_{l}\|\Sigma^{(l)}\|_{\rm op}
&& \text{(Theorem~\ref{thm: conv rates alg})} \\[0.3em]
\end{aligned}
\]
Applying the second result in Lemma \ref{lemma: obj val diff} which connects $\|\widehat{\Sigma}^{(l)}\|_{\rm op}$ and $\|\Sigma^{(l)}\|_{\rm op}$, noting that $(\mathbf{r}_{\max}+t)/n\leq 1$ for $t\leq n-\mathbf{r}_{\max}$ so that $\max_l\|\widehat{\Sigma}^{(l)}\|_{\rm op}\leq(1+2C_\sigma)\max_l\|\Sigma^{(l)}\|_{\rm op}$, and removing the cross-term, we establish that
\[
\begin{aligned}
\min_{\omega\in\Delta^L}
\langle \Sigma(\omega), \widehat{M}_T\rangle &\geq \max_{M\in\Fc^k}
\min_{\omega\in\Delta^L}
\langle \widehat{\Sigma}(\omega), M\rangle\\
&\quad -
C k\left(\sqrt{\frac{\mathbf{r}_{\max}+t}{n}}+\frac{\mathbf{r}_{\max}+t}{n} + \frac{\sqrt{k\log(d/k)\log L}}{T}\right)
\max_{l}\|{\Sigma}^{(l)}\|_{\rm op}.
\end{aligned}
\]

Moreover, since $M^*\in \Fc^k$, we continue to establish that
\begin{align*}
\max_{M\in\Fc^k}
\min_{\omega\in\Delta^L}
\langle \widehat{\Sigma}(\omega), M\rangle
&\ge
\min_{\omega\in\Delta^L}
\langle \widehat{\Sigma}(\omega), M^*\rangle\\
&\ge
\min_{\omega\in\Delta^L}
\langle \Sigma(\omega), M^*\rangle\\
&\quad - C k\!\left(
\sqrt{\frac{\mathbf{r}_{\max}+t}{n}}
+ \frac{\mathbf{r}_{\max}+t}{n}
\right)
\max_{l}\|\Sigma^{(l)}\|_{\rm op}
\quad \text{(by \eqref{eq: max diff})} \\
&= \max_{M\in \Fc^k}\min_{\omega\in\Delta^L}
\langle \Sigma(\omega), M\rangle\\
&\quad - C k\!\left(
\sqrt{\frac{\mathbf{r}_{\max}+t}{n}}
+ \frac{\mathbf{r}_{\max}+t}{n}
\right)
\max_{l}\|\Sigma^{(l)}\|_{\rm op}
\quad \text{(def.\ of $M^*$)}
\end{align*}

Combining the above two results and rearranging terms yields the claimed convergence bound: for any value $t\in[0, n-\mathbf{r}_{\max}]$, with probability at least $1-2Le^{-t}$,
\[
\begin{aligned}
&\max_{M\in\Fc^k}
\min_{\omega\in\Delta^L}
\langle \Sigma(\omega), M\rangle
-
\min_{\omega\in\Delta^L}
\langle \Sigma(\omega), \widehat{M}_T\rangle\\
&\le
Ck\!\left(\sqrt{\frac{\mathbf{r}_{\max}+t}{n}}
+ \frac{\mathbf{r}_{\max}+t}{n}
+
\frac{\sqrt{k\log(d/k)\log L}}{T}
\right)
\max_{l\in[L]}\|\Sigma^{(l)}\|_{\rm op}.
\end{aligned}
\]
Since $t\leq n-\mathbf{r}_{\max}$, we have $(\mathbf{r}_{\max}+t)/n\leq\sqrt{(\mathbf{r}_{\max}+t)/n}$, and the quadratic term is absorbed into the square-root term, which gives the statement of Theorem~\ref{thm: new conv}.

\subsection{Proof of Theorem \ref{thm: global conv rates alg}}
\label{proof of thm: global conv rates alg}

By Theorem~\ref{thm: conv rates alg} and the definition of
    $\tau$ in~\eqref{eq: certificate},
    \begin{equation}
        \begin{aligned}
        &\max_{M\in \Fc^k}\min_{\omega\in \Delta^L}\left\langle \widehat{\Sigma}(\omega), M\right\rangle - \min_{\omega\in \Delta^L}\left\langle \widehat{\Sigma}(\omega), \widehat{P}_T\right\rangle \\
        &\leq \max_{M\in \Fc^k}\min_{\omega\in \Delta^L}\left\langle \widehat{\Sigma}(\omega), M\right\rangle - \min_{\omega\in \Delta^L}\left\langle \widehat{\Sigma}(\omega), \widehat{M}_T\right\rangle + \left\{\min_{\omega\in \Delta^L}\left\langle \widehat{\Sigma}(\omega), \widehat{M}_T\right\rangle - \min_{\omega\in \Delta^L}\left\langle \widehat{\Sigma}(\omega), \widehat{P}_T\right\rangle\right\} \\
        &\leq \frac{8k\sqrt{k\log(d/k)\log L}\cdot \max_{l\in [L]}\|\widehat{\Sigma}^{(l)}\|_{\rm op}}{T} +\tau.
    \end{aligned}
    \label{eq: empirical objective gap of P_T}
    \end{equation}
    Since $\Fc^k\supseteq \Pc^k$, we have
    \[
    \max_{M\in \Fc^k}\min_{\omega\in \Delta^L}\left\langle \widehat{\Sigma}(\omega), M\right\rangle \geq \max_{P\in \Pc^k}\min_{\omega\in \Delta^L}\left\langle \widehat{\Sigma}(\omega), P\right\rangle.
    \]
    Combining the above two results yields the stated objective gap bound as follows:
    \[
    \begin{aligned}
        \max_{P\in \Pc^k}\min_{\omega\in \Delta^L}\left\langle \widehat{\Sigma}(\omega), P\right\rangle - \min_{\omega\in \Delta^L}\left\langle \widehat{\Sigma}(\omega), \widehat{P}_T\right\rangle 
        &\leq \max_{M\in \Fc^k}\min_{\omega\in \Delta^L}\left\langle \widehat{\Sigma}(\omega), M\right\rangle -\min_{\omega\in \Delta^L}\left\langle \widehat{\Sigma}(\omega), \widehat{P}_T\right\rangle \\
        &\leq \frac{8k\sqrt{k\log(d/k)\log L}\cdot \max_{l\in [L]}\|\widehat{\Sigma}^{(l)}\|_{\rm op}}{T} +\tau.
    \end{aligned}
    \]
Then we apply Lemma \ref{lemma: obj val diff}, which controls the finite-sample errors, to the above inequality exactly as in the proof of Theorem~\ref{thm: new conv}, with the quadratic term $(\mathbf{r}_{\max}+t)/n$ absorbed into $\sqrt{(\mathbf{r}_{\max}+t)/n}$ because $t\leq n-\mathbf{r}_{\max}$, and complete the proof.

\subsection{Proofs of Theorems \ref{thm: conv relaxed eigengap}, \ref{thm: conv for tau}, and \ref{thm: conv original eigengap}}
\label{proof of thm: conv for tau}

Throughout the proof, write
\[
f(M) := \min_{\omega\in\Delta^L}\langle\Sigma(\omega), M\rangle = \min_{l\in[L]}\langle\Sigma^{(l)}, M\rangle
\qquad\text{and}\qquad
\widehat{f}(M) := \min_{\omega\in\Delta^L}\langle\widehat{\Sigma}(\omega), M\rangle = \min_{l\in[L]}\langle\widehat{\Sigma}^{(l)}, M\rangle
\]
for the population and empirical objectives, where the second equalities hold because a linear function of $\omega$ is minimized over the simplex at a vertex. Write 
$${\rm OPT} := \max_{M\in\Fc^k}f(M),$$ for the optimal value of the relaxed StablePCA \eqref{eq: fantope MPCA}, and, for the sampling errors,
\[
e_{\rm op} := \max_{l\in[L]}\big\|\widehat{\Sigma}^{(l)} - \Sigma^{(l)}\big\|_{\rm op}
\qquad\text{and}\qquad
e(M) := \max_{l\in[L]}\big|\langle\widehat{\Sigma}^{(l)} - \Sigma^{(l)}, M\rangle\big|\quad\text{for } M\in\Fc^k .
\]
Since $|\min_l a_l - \min_l b_l|\leq\max_l|a_l - b_l|$ for any two families of reals $\{a_l\}$ and $\{b_l\}$, we have
\begin{equation}
    \big|f(M) - \widehat{f}(M)\big|\leq e(M)\qquad\text{for every } M\in\Fc^k .
    \label{eq: proof tau - f vs fhat}
\end{equation}
Finally, let $\epsilon_T := \max_{M\in\Fc^k}\widehat{f}(M) - \widehat{f}(\widehat{M}_T)\geq 0$ denote the optimization error of $\widehat{M}_T$ for the empirical problem \eqref{eq: empirical stablepca}, which by Theorem~\ref{thm: conv rates alg} satisfies
\begin{equation}
    \epsilon_T\leq\frac{8\hat{\rho}_{\max}\,k\sqrt{k\log(d/k)\log L}}{T} .
    \label{eq: proof tau - eps T}
\end{equation}
The proof proceeds in five steps. 
\begin{itemize}
    \item Step 1 bounds the objective gap $G := {\rm OPT} - f(\widehat{M}_T)$ deterministically in terms of $e(P^*)$, $e_{\rm op}$, and $\|\widehat{M}_T - P^*\|_F$.
    \item Step 2 uses the eigengap to convert $\|\widehat{M}_T - P^*\|_F$ back into $G$ and solves the resulting inequality. 
    \item Step 3 bounds the sampling errors with high probability, which proves Theorem~\ref{thm: conv relaxed eigengap}.
    \item Step 4 bounds the certificate, which proves Theorem~\ref{thm: conv for tau}.
    \item Step 5 treats the projected output, which proves Theorem~\ref{thm: conv original eigengap}.
\end{itemize}

\medskip
\noindent\textbf{Step 1: a deterministic bound on the objective gap.}
Under Condition~\ref{cond: eigengap}, Theorem~\ref{thm: fantope tight} shows that $P^*$ is the unique solution of \eqref{eq: fantope MPCA}, so that $f(P^*) = {\rm OPT}$. By \eqref{eq: proof tau - f vs fhat} applied at $P^*$ and at $\widehat{M}_T$, and since $\widehat{f}(P^*)\leq\max_{M\in\Fc^k}\widehat{f}(M) = \widehat{f}(\widehat{M}_T) + \epsilon_T$ by the definition of $\epsilon_T$,
\begin{equation}
    G = f(P^*) - f(\widehat{M}_T)\leq\big[\widehat{f}(P^*) + e(P^*)\big] - \big[\widehat{f}(\widehat{M}_T) - e(\widehat{M}_T)\big]\leq\epsilon_T + e(P^*) + e(\widehat{M}_T) .
    \label{eq: proof tau - step1a}
\end{equation}
To bound $e(\widehat{M}_T)$, write $\Delta_M := \widehat{M}_T - P^*$. By the triangle inequality and H\"older's inequality $|\langle A, B\rangle|\leq\|A\|_{\rm op}\|B\|_*$,
\begin{equation}
    e(\widehat{M}_T)\leq e(P^*) + \max_{l\in[L]}\big|\langle\widehat{\Sigma}^{(l)} - \Sigma^{(l)}, \Delta_M\rangle\big|\leq e(P^*) + e_{\rm op}\,\|\Delta_M\|_* .
    \label{eq: proof tau - step1b}
\end{equation}
We now show that
\begin{equation}
    \|\Delta_M\|_*\leq 2\sqrt{k}\,\|\Delta_M\|_F .
    \label{eq: proof tau - nuclear frobenius}
\end{equation}
Since $\widehat{M}_T\succeq 0$, we have $\Delta_M\succeq -P^*$, and by Weyl's monotonicity theorem $\lambda_i(\Delta_M)\geq\lambda_i(-P^*)$ for every $i\in[d]$, with eigenvalues in nonincreasing order. The matrix $-P^*$ has the eigenvalue $0$ with multiplicity $d-k$ and the eigenvalue $-1$ with multiplicity $k$, so $\lambda_i(-P^*) = 0$ for $i\leq d-k$, and hence $\Delta_M$ has at most $k$ negative eigenvalues. Moreover, ${\rm Tr}(\Delta_M) = {\rm Tr}(\widehat{M}_T) - {\rm Tr}(P^*) = k - k = 0$, so the positive eigenvalues of $\Delta_M$ sum to the same value as the absolute values of its negative eigenvalues, and therefore
\[
\|\Delta_M\|_* = \sum_{i=1}^d|\lambda_i(\Delta_M)| = 2\sum_{i:\,\lambda_i(\Delta_M) < 0}|\lambda_i(\Delta_M)|\leq 2\sqrt{k}\Big(\sum_{i:\,\lambda_i(\Delta_M)<0}\lambda_i(\Delta_M)^2\Big)^{1/2}\leq 2\sqrt{k}\,\|\Delta_M\|_F ,
\]
by the Cauchy--Schwarz inequality over the at most $k$ negative eigenvalues. Combining \eqref{eq: proof tau - step1a}, \eqref{eq: proof tau - step1b}, and \eqref{eq: proof tau - nuclear frobenius} gives
\begin{equation}
    G\leq\epsilon_T + 2e(P^*) + 2\sqrt{k}\,e_{\rm op}\,\|\widehat{M}_T - P^*\|_F .
    \label{eq: proof tau - step1}
\end{equation}

\medskip
\noindent\textbf{Step 2: localization through the eigengap.}
We use the following curvature inequality, first introduced in \cite{vu2013fantope}, which converts the objective gap at $\Sigma(\omega^*)$ into a distance from $P^*$.
    \begin{Lemma}[Lemma 3.1 of \cite{vu2013fantope}]
        Let $A\in \RR^{d\times d}$ be a symmetric matrix and $E$ be the projection onto the subspace spanned by the eigenvectors of $A$ corresponding to its $k$ largest eigenvalues $\lambda_1\geq \lambda_2 \geq \cdots\geq \lambda_k$. If $\delta_A = \lambda_k - \lambda_{k+1}>0$, then
        \[
        \frac{\delta_A}{2}\|E-F\|_F^2 \leq \langle A, E-F\rangle,
        \]
        for all $F$ satisfying that $0\preceq F\preceq {\bf I}$ and ${\rm Tr}(F) = k$.
        \label{lemma: curvation}
    \end{Lemma}
Applying this lemma with $A = \Sigma(\omega^*)$, $E = P^*$, $F = \widehat{M}_T$, and $\delta_A = \delta^*$ yields
\[
\frac{\delta^*}{2}\|\widehat{M}_T - P^*\|_F^2\leq\langle\Sigma(\omega^*), P^* - \widehat{M}_T\rangle .
\]
By Ky Fan's maximum principle and Sion's minimax theorem, as in \eqref{eq: omega_star}, and because $\omega^*\in\Delta^L$,
\[
\langle\Sigma(\omega^*), P^*\rangle = \phi(\omega^*) = \min_{\omega\in\Delta^L}\phi(\omega) = {\rm OPT}
\qquad\text{and}\qquad
\langle\Sigma(\omega^*), \widehat{M}_T\rangle\geq\min_{\omega\in\Delta^L}\langle\Sigma(\omega), \widehat{M}_T\rangle = f(\widehat{M}_T) .
\]
Therefore $\langle\Sigma(\omega^*), P^* - \widehat{M}_T\rangle\leq{\rm OPT} - f(\widehat{M}_T) = G$, and
\begin{equation}
    \|\widehat{M}_T - P^*\|_F^2\leq\frac{2G}{\delta^*} .
    \label{eq: proof tau - curvature}
\end{equation}
Substituting \eqref{eq: proof tau - curvature} into \eqref{eq: proof tau - step1} and writing $a := \epsilon_T + 2e(P^*)$ and $b := 2\sqrt{2k/\delta^*}\,e_{\rm op}$, we obtain $G\leq a + b\sqrt{G}\leq a + b^2/2 + G/2$, where the last step is the elementary inequality $b\sqrt{G}\leq b^2/2 + G/2$. Rearranging, $G\leq 2a + b^2 = 2\epsilon_T + 4e(P^*) + 8k\,e_{\rm op}^2/\delta^*$. Alternatively, since $\|\Delta_M\|_*\leq\|\widehat{M}_T\|_* + \|P^*\|_* = 2k$, the inequalities \eqref{eq: proof tau - step1a} and \eqref{eq: proof tau - step1b} give directly $G\leq\epsilon_T + 2e(P^*) + 2k\,e_{\rm op}$, without using the eigengap. Together,
\begin{equation}
    G\leq 2\epsilon_T + 4e(P^*) + \min\left\{\frac{8k\,e_{\rm op}^2}{\delta^*},\ 2k\,e_{\rm op}\right\} .
    \label{eq: proof tau - step2}
\end{equation}
This is a deterministic inequality. The eigengap has localized the problem: the uniform control of the sampling errors over the Fantope enters only through $e_{\rm op}^2$, which is multiplied by the factor $1/\delta^*$, while the remaining sampling error $e(P^*)$ is evaluated at the fixed matrix $P^*$.

\vspace{0.5em}
\noindent\textbf{Step 3: the sampling errors.}
By Lemma~\ref{lemma: obj val diff}, with probability at least $1 - 2Le^{-t}$,
\begin{equation}
    e_{\rm op}\leq C_\sigma\rho_{\max}\left(\sqrt{\frac{\mathbf{r}_{\max}+t}{n}} + \frac{\mathbf{r}_{\max}+t}{n}\right)\leq 2C_\sigma\rho_{\max}\sqrt{\frac{\mathbf{r}_{\max}+t}{n}}
    \qquad\text{and}\qquad
    \hat{\rho}_{\max}\leq(1 + 2C_\sigma)\rho_{\max},
    \label{eq: proof tau - E max}
\end{equation}
where we used $t\leq n - \mathbf{r}_{\max}$, so that $(\mathbf{r}_{\max}+t)/n\leq 1$. We next bound $e(P^*)$. Fix $l\in[L]$, write $P^* = \sum_{j=1}^kv_jv_j^\intercal$ with orthonormal $v_1,\dots,v_k\in\RR^d$, and let $W_i := \langle X_i^{(l)}X_i^{(l)\intercal}, P^*\rangle = \sum_{j=1}^k\langle X_i^{(l)}, v_j\rangle^2$ for $i\in[n_l]$, so that
\[
\langle\widehat{\Sigma}^{(l)} - \Sigma^{(l)}, P^*\rangle = \frac{1}{n_l}\sum_{i=1}^{n_l}\big(W_i - \EE W_i\big)
\]
is an average of independent and identically distributed centered random variables. By the triangle inequality for the sub-exponential norm $\|\cdot\|_{\psi_1}$, the inequality $\|\xi^2\|_{\psi_1}\leq\|\xi\|_{\psi_2}^2$ \citep[Lemma 2.7.6]{vershynin2018high}, and Condition~\ref{cond: subgauss},
\[
\|W_i\|_{\psi_1}\leq\sum_{j=1}^k\big\|\langle X_i^{(l)}, v_j\rangle^2\big\|_{\psi_1}\leq\sum_{j=1}^k\big\|\langle X_i^{(l)}, v_j\rangle\big\|_{\psi_2}^2\leq\sigma^2\sum_{j=1}^kv_j^\intercal\Sigma^{(l)}v_j = \sigma^2\langle\Sigma^{(l)}, P^*\rangle\leq\sigma^2k\rho_{\max} ,
\]
and centering changes the $\psi_1$-norm by at most a universal factor \citep[Section 2.7]{vershynin2018high}. Bernstein's inequality for sums of independent sub-exponential random variables \citep[Theorem 2.8.1]{vershynin2018high} then gives, for any $t\geq 0$, with probability at least $1 - 2e^{-t}$,
\[
\big|\langle\widehat{\Sigma}^{(l)} - \Sigma^{(l)}, P^*\rangle\big|\leq C\sigma^2k\rho_{\max}\left(\sqrt{\frac{t}{n_l}} + \frac{t}{n_l}\right)\leq 2C\sigma^2k\rho_{\max}\sqrt{\frac{t}{n}} ,
\]
using $n_l\geq n\geq t$. A union bound over $l\in[L]$ shows that, with probability at least $1 - 2Le^{-t}$,
\begin{equation}
    e(P^*)\leq 2C\sigma^2k\rho_{\max}\sqrt{\frac{t}{n}} .
    \label{eq: proof tau - e Pstar}
\end{equation}
This bound involves neither $d$ nor $\mathbf{r}_{\max}$: it is the sampling error of the $L$ explained variances $\langle\widehat{\Sigma}^{(l)}, P^*\rangle$ of the fixed subspace $P^*$. On the intersection of the two events, which has probability at least $1 - 4Le^{-t}$, substituting \eqref{eq: proof tau - eps T}, \eqref{eq: proof tau - E max}, and \eqref{eq: proof tau - e Pstar} into \eqref{eq: proof tau - step2} yields
\[
G\leq C'\rho_{\max}k\left(\sqrt{\frac{t}{n}} + \min\left\{\frac{\rho_{\max}}{\delta^*}\cdot\frac{\mathbf{r}_{\max}+t}{n},\ \sqrt{\frac{\mathbf{r}_{\max}+t}{n}}\right\} + \frac{\sqrt{k\log(d/k)\log L}}{T}\right) = \mathcal{E}_{n,T}
\]
for a constant $C'$ depending only on $\sigma$, which is the first claim of Theorem~\ref{thm: conv relaxed eigengap}. The second claim follows from \eqref{eq: proof tau - curvature}:
\begin{equation}
    \|\widehat{M}_T - P^*\|_F^2\leq\frac{2G}{\delta^*}\leq\frac{2\,\mathcal{E}_{n,T}}{\delta^*} .
    \label{eq: projection error step 2}
\end{equation}
Finally, recalling that $\widehat{P}_T = \argmin_{P\in \Pc^k} \|P-\widehat{M}_T\|_F$ in \eqref{eq: proj} and that $P^*\in\Pc^k$, we have
\begin{equation}
    \|\widehat{P}_T-\widehat{M}_T \|_F^2 \leq \|P^* - \widehat{M}_T\|_F^2 \leq \frac{2\,\mathcal{E}_{n,T}}{\delta^*} .
    \label{eq: upper bound of projection error}
\end{equation}

\medskip
\noindent\textbf{Step 4: bounding the certificate $|\tau|$.}

    By definition,
\[
\left|\tau\right|
=
\left|\min_{1\le l\le L}
\langle \widehat{\Sigma}^{(l)}, \widehat{M}_T\rangle - \min_{1\le l\le L}
\langle \widehat{\Sigma}^{(l)}, \widehat{P}_T\rangle\right| \leq \max_{1\leq l\leq L}\left|\langle \widehat{\Sigma}^{(l)}, \widehat{M}_T-\widehat{P}_T\rangle\right| .
\]
By H\"older's inequality,
\begin{equation}
    \left|\tau\right| \leq \max_{l\in [L]}\|\widehat{\Sigma}^{(l)}\|_{\rm op}\cdot \|\widehat{M}_T - \widehat{P}_T\|_*.
    \label{eq: upper bound of tau}
\end{equation}

Next, we establish the upper bound of $\|\widehat{M}_T - \widehat{P}_T\|_*$. For notation convenience, we define
\[
\Delta_T := \widehat{M}_T - \widehat{P}_T.
\]
As $\widehat{M}_T \in \Fc^k$ and $\widehat{P}_T\in \Pc^k$, we have
\[
\Tr(\Delta_T) = \Tr(\widehat{M}_T) - \Tr(\widehat{P}_T) = k-k = 0.
\]
Therefore, the total positive eigenvalues equal the total negative eigenvalues:
\[
\sum_{i: \lambda_i(\Delta_T)>0} \lambda_i (\Delta_T) = -\sum_{i: \lambda_i(\Delta_T)<0} \lambda_i(\Delta_T).
\]
Hence,
\begin{equation}
    \|\Delta_T\|_* = \sum_{i=1}^d|\lambda_i(\Delta_T)| = 2\sum_{i: \lambda_i(\Delta_T)<0}(-\lambda_i(\Delta_T)).
    \label{eq: Delta_T nuclear}
\end{equation}

More importantly, note that $\widehat{P}_T$ is the Frobenius projection of $\widehat{M}_T$ onto $\Pc^k$. Let
\[
\widehat{M}_T = U \Diag(\lambda_1,\lambda_2,...,\lambda_d) U^\intercal, \quad \textrm{with}\quad \lambda_1\geq \dots\geq \lambda_d, \; 0\leq \lambda_i\leq 1, \; \sum_{i=1}^d \lambda_i = k.
\]
Then $\widehat{P}_T$ admits the form as:
\[
\widehat{P}_T = U\Diag(\underbrace{1,\dots, 1}_{k}, \underbrace{0,\dots, 0}_{d-k}) U^\intercal,
\]
which implies that
\[
\Delta_T = \widehat{M}_T - \widehat{P}_T = U \Diag(\lambda_1 - 1,...\lambda_k-1, \lambda_{k+1},\dots, \lambda_{d}) U^\intercal.
\]
Consequently, $\Delta_T$ has at most $k$ negative eigenvalues.

We then continue \eqref{eq: Delta_T nuclear} to establish the upper bound of $\|\Delta_T\|_*$. By Cauchy-Schwarz,
\begin{equation}
\frac{1}{2}\|\Delta_T\|_* = \sum_{i:\lambda_i(\Delta_T)<0} (-\lambda_i) \leq \sqrt{k}\left(\sum_{i:\lambda_i(\Delta_T)<0} \lambda_i^2(\Delta_T)\right)^{1/2} \leq \sqrt{k}\|\Delta_T\|_F.
\label{eq: proof tau - Delta T nuclear frobenius}
\end{equation}
Therefore, it follows from \eqref{eq: upper bound of tau} that
\[
\left|\tau\right|^2 \leq 4k\max_{l\in [L]}\|\widehat{\Sigma}^{(l)}\|_{\rm op}^2\cdot \|\widehat{M}_T - \widehat{P}_T\|_{F}^2.
\]
Together with \eqref{eq: upper bound of projection error} and the bound $\hat{\rho}_{\max}\leq(1+2C_\sigma)\rho_{\max}$ in \eqref{eq: proof tau - E max}, we obtain $|\tau|^2\leq 8k(1+2C_\sigma)^2\rho_{\max}^2\,\mathcal{E}_{n,T}/\delta^*$, which is Theorem~\ref{thm: conv for tau}.

\medskip
\noindent\textbf{Step 5: the projected output (Theorem~\ref{thm: conv original eigengap}).}

On the event of the previous steps, the inequalities \eqref{eq: projection error step 2} and \eqref{eq: upper bound of projection error} give $\|\widehat{M}_T - P^*\|_F^2\leq 2\mathcal{E}_{n,T}/\delta^*$ and $\|\widehat{P}_T - \widehat{M}_T\|_F^2\leq 2\mathcal{E}_{n,T}/\delta^*$. Therefore
\[
\|\widehat{P}_T - P^*\|_F^2\leq 2\|\widehat{P}_T - \widehat{M}_T\|_F^2 + 2\|\widehat{M}_T - P^*\|_F^2\leq\frac{8\,\mathcal{E}_{n,T}}{\delta^*} ,
\]
which is the second bound of Theorem~\ref{thm: conv original eigengap}. For the first bound, by Theorem~\ref{thm: fantope tight} the optimal values of \eqref{eq: fantope MPCA} and \eqref{eq: def diStablePCA} coincide, so the objective gap of $\widehat{P}_T$ for the original problem decomposes as $\{{\rm OPT} - f(\widehat{M}_T)\} + \{f(\widehat{M}_T) - f(\widehat{P}_T)\}$. The first term is at most $\mathcal{E}_{n,T}$ by Theorem~\ref{thm: conv relaxed eigengap}. For the second, since a minimum over $l$ is bounded by the same index, $f(\widehat{M}_T) - f(\widehat{P}_T)\leq\max_l\langle\Sigma^{(l)}, \widehat{M}_T - \widehat{P}_T\rangle\leq\rho_{\max}\|\widehat{M}_T - \widehat{P}_T\|_*\leq 2\sqrt{k}\,\rho_{\max}\|\widehat{M}_T - \widehat{P}_T\|_F$, by H\"older's inequality and \eqref{eq: proof tau - Delta T nuclear frobenius}, and $\|\widehat{M}_T - \widehat{P}_T\|_F\leq\sqrt{2\mathcal{E}_{n,T}/\delta^*}$. Hence $f(\widehat{M}_T) - f(\widehat{P}_T)\leq 2\sqrt{2}\,\rho_{\max}\sqrt{k\mathcal{E}_{n,T}/\delta^*}\leq 3\rho_{\max}\sqrt{k\mathcal{E}_{n,T}/\delta^*}$, which gives the first bound.

\subsection{Proof of Theorem \ref{thm: conv rates alg saddle}}
\label{proof of thm: conv rates alg saddle}

Recall that Lemma~\ref{thm: MP conv general} provides an $\mathcal{O}(T^{-1})$ convergence rate
for general convex--concave minimax problems under Condition~\ref{cond: mirror and smooth}.
Thus, to prove Theorem~\ref{thm: conv rates alg saddle}, it suffices to verify that
Condition~\ref{cond: mirror and smooth} holds for the empirical StablePCA objective
\[
\min_{M\in\mathcal{F}^k}\max_{\omega\in\Delta^L}\;f(M,\omega),
\qquad
f(M,\omega):= -\sum_{l=1}^L \omega_l \,\langle \widehat\Sigma^{(l)}, M\rangle.
\]

We identify the primal and dual domains in Lemma~\ref{thm: MP conv general} as
$\mathcal{X}=\mathcal{F}^k,\mathcal{Y}=\Delta^L,$
and choose the mirror maps
\[
\psi_{\Fc^k}(M)=\psi_1(M):=\Tr(M\log M),
\qquad
\psi_{\Delta^L}(\omega)=\psi_2(\omega):=\sum_{l=1}^L \omega_l\log\omega_l.
\]
Let $D_{\psi_1}$ and $D_{\psi_2}$ be the induced Bregman divergences, specified in  \eqref{eq: bregman}.

The following lemma verifies Condition~\ref{cond: mirror and smooth} for the empirical Fantope-relaxed StablePCA. The proof is provided in Appendix \ref{proof of lem: stablepca_condition_verify}.

\begin{Lemma}
\label{lem: stablepca_condition_verify}
The mirror maps $\psi_1(M)$ and $\psi_2(\omega)$ are strongly convex such that:
\begin{itemize}
    \item $\psi_1$ is $\frac{1}{2k}$-strongly convex on $\mathcal{F}^k$
    with respect to the Schatten-$1$ (nuclear) norm $\|\cdot\|_*$.
    Moreover,
    \[
    R_{\Fc^k}^2
    :=\sup_{M\in\mathcal{F}^k}\psi_1(M)-\inf_{M\in\mathcal{F}^k}\psi_1(M)
    = k\log \frac{d}{k}.
    \]
    \item $\psi_2$ is $1$-strongly convex on $\Delta^L$
    with respect to the $\ell_1$ norm $\|\cdot\|_1$.
    Moreover,
    \[
    R_{\Delta^L}^2
    :=\sup_{\omega\in\Delta^L}\psi_2(\omega)-\inf_{\omega\in\Delta^L}\psi_2(\omega)
    = \log L.
    \]
\end{itemize}
    Furthermore, $f(M,\omega)$ is $(\beta_{11},\beta_{12},\beta_{21},\beta_{22})$-smooth
    in the sense of Condition~\ref{cond: mirror and smooth} under the norm pair
    $\|M\|_*$ and $\|\omega\|_1$, with
    \[
    (\beta_{11},\beta_{12},\beta_{21},\beta_{22})
    =
    \Big(0,\;\beta,\;\beta,\;0\Big),
    \qquad
    \beta := \max_{1\le l\le L}\|\widehat\Sigma^{(l)}\|_{\rm op}.
    \]
\end{Lemma}

By Lemma~\ref{lem: stablepca_condition_verify}, Condition~\ref{cond: mirror and smooth}
holds with
\[
\mu_{\mathcal{X}}=\frac{1}{2k},\quad \mu_{\mathcal{Y}}=1,
\qquad
(\beta_{11},\beta_{12},\beta_{21},\beta_{22})=(0,\beta,\beta,0),
\qquad
R_{\mathcal{X}}^2= k\log \frac{d}{k},\quad R_{\mathcal{Y}}^2=\log L.
\]
Substituting these quantities into Lemma~\ref{thm: MP conv general} gives the desired
$\mathcal{O}(T^{-1})$ duality-gap bound for the averaged midpoint iterates
\[
\widehat M_T := \frac{1}{T}\sum_{t=0}^{T-1} M_{t+\frac12},
\qquad
\widehat\omega_T := \frac{1}{T}\sum_{t=0}^{T-1} \omega_{t+\frac12}.
\]

Specifically, by setting
\[
\eta
= \frac{\mu_{\mathcal{X}}\wedge\mu_{\mathcal{Y}}}
{2\max\left\{
\frac{\beta_{11}}{a},
\frac{\beta_{22}}{b},
\frac{\beta_{12}}{\sqrt{ab}},
\frac{\beta_{21}}{\sqrt{ab}}
\right\}} = \frac{\frac{1}{2k}}{2\cdot \frac{\beta}{\sqrt{ab}}} = \frac{\sqrt{ab}}{4k\beta},
\]
Lemma~\ref{thm: MP conv general} gives
\[
\begin{aligned}
    \max_{\omega\in\Delta^L}
f\!\left(\widehat{M}_T,\, \omega\right)
-
\min_{M\in\mathcal{F}^k}
f\!\left(M,\, \widehat{\omega}_T\right)
&\le
2\max\left\{
\frac{\beta_{11}}{a},
\frac{\beta_{22}}{b},
\frac{\beta_{12}}{\sqrt{ab}},
\frac{\beta_{21}}{\sqrt{ab}}
\right\}
\frac{aR_{\mathcal{X}}^{2}+bR_{\mathcal{Y}}^{2}}
{\mu_{\mathcal{X}}\wedge\mu_{\mathcal{Y}}}
\cdot \frac{1}{T} \\
&\leq \frac{4k\beta}{\sqrt{ab}}\left(ak\log \frac{d}{k} + b\log L\right) \frac{1}{T}.
\end{aligned}
\]

We take $a = \log L$ and $b = k\log (d/k)$, then the step size becomes
\[
\eta = \frac{1}{4\max_{l\in [L]}\|\widehat{\Sigma}^{(l)}\|_{\rm op}}\sqrt{\frac{\log L\,\log(d/k)}{k}}.
\]
By \eqref{eq: MP general first} and \eqref{eq: MP general second}, it holds
\[
\eta_M = \frac{\eta}{a} = \frac{1}{4\max_{l\in [L]}\|\widehat{\Sigma}^{(l)}\|_{\rm op}}\sqrt{\frac{\log(d/k)}{k\log L}}, \qquad \eta_\omega = \frac{\eta}{b} = \frac{1}{4\max_{l\in [L]}\|\widehat{\Sigma}^{(l)}\|_{\rm op}}\sqrt{\frac{\log L}{k^3\log(d/k)}}.
\]
And the final primal-dual gap simplifies as
\[
\max_{\omega\in\Delta^L}
f\!\left(\widehat{M}_T,\, \omega\right)
-
\min_{M\in\mathcal{F}^k}
f\!\left(M,\, \widehat{\omega}_T\right)
\le \frac{8k\sqrt{k\log(d/k)\log L}}{T} \max_{l\in [L]}\|\widehat{\Sigma}^{(l)}\|_{\rm op}.
\]
which completes the proof.

\subsection{Proof of Proposition \ref{prop: alternative reformulation}}
\label{proof of prop: alternative reformulation}
Fix a source $l$ and a projection $P\in\Pc^k$. Since $X - PX = ({\bf I}_d - P)X$ and ${\bf I}_d - P$ is a projection,
\[
\EE\|X^{(l)} - PX^{(l)}\|_2^2 = \langle \Sigma^{(l)}, {\bf I}_d - P\rangle = {\rm Tr}(\Sigma^{(l)}) - \langle\Sigma^{(l)}, P\rangle ,
\]
and by Ky Fan's maximum principle, $\min_{P'\in\Pc^k}\EE\|X^{(l)} - P'X^{(l)}\|_2^2 = {\rm Tr}(\Sigma^{(l)}) - \sum_{i=1}^k\lambda_i^{(l)} = \sum_{i=k+1}^d\lambda_i^{(l)}$. Moreover, $\langle c\,{\bf I}_d, P\rangle = ck$ for every scalar $c$, so that ${\rm Tr}(\Sigma^{(l)}) = \langle\frac1k\sum_{i=1}^d\lambda_i^{(l)}{\bf I}_d, P\rangle$ and $\sum_{i=1}^k\lambda_i^{(l)} = \langle\frac1k\sum_{i=1}^k\lambda_i^{(l)}{\bf I}_d, P\rangle$. Hence the unexplained variance and the regret of source $l$ are the linear functions
\[
\EE\|X^{(l)} - PX^{(l)}\|_2^2 = -\Big\langle \Sigma^{(l)} - \frac{1}{k}\sum_{i=1}^{d}\lambda_i^{(l)}{\bf I}_d,\ P\Big\rangle,
\qquad
{\rm Regret}_{\TT^{(l)}}(P) = -\Big\langle \Sigma^{(l)} - \frac{1}{k}\sum_{i=1}^{k}\lambda_i^{(l)}{\bf I}_d,\ P\Big\rangle .
\]
SquaredPCA \eqref{eq: squaredPCA} minimizes the maximum over $l$ of the first expression, i.e., maximizes $\min_l\langle\Sigma^{(l)} - \frac1k\sum_{i\leq d}\lambda_i^{(l)}{\bf I}_d, P\rangle$; FairPCA \eqref{eq: fairPCA} minimizes $\max_{\QQ\in\Cc}{\rm Regret}_\QQ(P) = \max_l{\rm Regret}_{\TT^{(l)}}(P)$ by Lemma~\ref{lemma: fairpca equiv}, i.e., maximizes $\min_l\langle\Sigma^{(l)} - \frac1k\sum_{i\leq k}\lambda_i^{(l)}{\bf I}_d, P\rangle$. Finally, for any matrices $A^{(1)},\dots,A^{(L)}$, the function $\omega\mapsto\sum_l\omega_l\langle A^{(l)}, P\rangle$ is linear on the simplex and attains its minimum at a vertex, so $\min_l\langle A^{(l)}, P\rangle = \min_{\omega\in\Delta^L}\sum_l\omega_l\langle A^{(l)}, P\rangle$, which gives the stated forms.

\section{Proofs of Technical Lemmas}

\subsection{Proof of Lemma \ref{lemma: eigenvalue function}}
\label{proof of lemma: eigenvalue function}
Recall that $\phi(\omega)$ admits the form in \eqref{eq: omega_star}:
\begin{equation}
\phi(\omega)
=
\max_{M\in\Fc^k}
\sum_{l=1}^L \omega_l \langle \Sigma^{(l)}, M\rangle .
\label{eq:phi-variational}
\end{equation}

\vspace{0.5em}
\noindent\textbf{Continuity and Convexity.}

For each fixed $M\in\Fc^k$, the map
\(
\omega \mapsto \sum_{l=1}^L \omega_l \langle \Sigma^{(l)}, M\rangle
\)
is affine in $\omega$. Since $\phi(\omega)$ is the pointwise maximum over $M\in\Fc^k$
of affine functions, it follows that $\phi$ is convex on $\Delta^L$.
Moreover, because $\Fc^k$ is compact and the function
$(\omega,M)\mapsto \sum_{l=1}^L \omega_l \langle \Sigma^{(l)}, M\rangle$
is continuous, the maximum in \eqref{eq:phi-variational} depends continuously on $\omega$.
Hence, $\phi$ is continuous on $\Delta^L$.

\vspace{0.5em}
\noindent\textbf{Differentiability under an eigengap.}

Fix $\omega\in\Delta^L$ such that
\(
\lambda_k(\Sigma(\omega)) > \lambda_{k+1}(\Sigma(\omega)).
\)
Under this eigengap condition, the top-$k$ eigenspace of $\Sigma(\omega)$ is unique,
and therefore the maximizer of
\(
\max_{M\in\Fc^k} \langle \Sigma(\omega), M\rangle
\)
is unique. Denote this maximizer by $P_k(\omega)$, the orthogonal projection matrix
onto the top-$k$ eigenspace of $\Sigma(\omega)$.

By Danskin's theorem, since the maximizer in \eqref{eq:phi-variational} is unique at $\omega$,
the function $\phi$ is differentiable at $\omega$, and its gradient is given by
\[
\nabla \phi(\omega)
=
\big(
\langle \Sigma^{(1)}, P_k(\omega)\rangle,\,
\ldots,\,
\langle \Sigma^{(L)}, P_k(\omega)\rangle
\big)^\intercal.
\]

Combining the above arguments completes the proof.

\subsection{Proof of Lemma \ref{lem: subgrad_phi}}
\label{proof of lem: subgrad_phi}
    For any symmetric matrix $A\in\mathbb{R}^{d\times d}$,
Ky Fan's maximum principle states that
\begin{equation}
\sum_{i=1}^k \lambda_i(A)
=
\max_{P\in\Pc^k} \langle A, P\rangle,
\label{eq: kyfan}
\end{equation}
Applying \eqref{eq: kyfan} to $\Sigma(\omega)$ gives
\begin{equation}
\phi(\omega)
=
\max_{P\in\Pc^k} \langle \Sigma(\omega), P\rangle.
\label{eq: phi_as_max}
\end{equation}

Let $P(\omega)\in\argmax_{P\in\Pc^k} \langle \Sigma(\omega), P\rangle$ be any maximizer.
Equivalently, $P(\omega)=U(\omega)(U(\omega))^\intercal$ where the columns of $U(\omega)$ span a top-$k$ eigenspace of $\Sigma(\omega)$.
Then
\begin{equation}
\phi(\omega)=\langle \Sigma(\omega), P(\omega)\rangle.
\label{eq: opt_at_omega}
\end{equation}

For any $\omega'\in\Delta^L$, since $\phi(\omega')$ is the maximum over $P$,
we can evaluate it at $P(\omega)$ to obtain
\begin{equation}
\phi(\omega')
=
\max_{P\in\Pc^k}\langle \Sigma(\omega'),P\rangle
\ge
\langle \Sigma(\omega'), P(\omega)\rangle.
\label{eq: lower_by_same_P}
\end{equation}
Subtracting \eqref{eq: opt_at_omega} from \eqref{eq: lower_by_same_P} yields
\[
\phi(\omega')-\phi(\omega)
\ge
\langle \Sigma(\omega')-\Sigma(\omega),\, P(\omega)\rangle.
\]
Using the linearity $\Sigma(\omega')-\Sigma(\omega)=\sum_{l=1}^L (\omega'_l-\omega_l)\Sigma^{(l)}$,
\[
\langle \Sigma(\omega')-\Sigma(\omega),\, P(\omega)\rangle
=
\sum_{l=1}^L (\omega'_l-\omega_l)\langle \Sigma^{(l)}, P(\omega)\rangle
=
\langle g(\omega),\, \omega'-\omega\rangle,
\]
where $g_l(\omega)=\langle\Sigma^{(l)},P(\omega)\rangle$.
Therefore,
\[
\phi(\omega') \ge \phi(\omega) + \langle g(\omega),\, \omega'-\omega\rangle,
\qquad \forall\,\omega'\in\Delta^L,
\]
which proves that $g(\omega)\in\partial \phi(\omega)$.

\subsection{Proof of Lemma \ref{lemma: fairpca equiv}}
\label{proof of lemma: fairpca equiv}
Given a distribution $\QQ$, the regret over a projection matrix $P\in \Pc^k$ is defined as:
\[
{\rm Regret}_\QQ(P) = \EE_{\QQ}\|X-PX\|_2^2 - \min_{P'\in \Pc^k}\EE_{\QQ}\|X-P'X\|_2^2.
\]
Recall that the uncertainty set $\Cc$ consists of all mixtures of the source distributions. We take the worst-case regret over $\Cc$:
\[
\max_{\QQ\in \mathcal{C}}{\rm Regret}_\QQ(P) = \max_{\omega\in \Delta^L}\left\{\EE_{X\sim \sum_{l=1}^L\omega_l\TT^{(l)}}\|X-PX\|_2^2 - \min_{P'\in \Pc^k}\EE_{X\sim \sum_{l=1}^L\omega_l\TT^{(l)}}\|X-P'X\|_2^2\right\}.
\]

We now simplify the right-hand side. First, by linearity of expectation,
\[
\EE_{X\sim \sum_{l=1}^L\omega_l\TT^{(l)}}\|X-PX\|_2^2 = \sum_{l=1}^L \omega_l\cdot \EE_{\TT^{(l)}}\|X-PX\|_2^2.
\]
Second, the optimal rank-$k$ reconstruction error under $\omega$-weighted mixture is
\[
\min_{P'\in \Pc^k}\EE_{X\sim \sum_{l=1}^L\omega_l\TT^{(l)}}\|X-P'X\|_2^2
=\sum_{i=k+1}^d\lambda_i\left(\Sigma(\omega)\right),
\]
which is a concave function of $\omega$, since the sum of the bottom eigenvalues are concave for $\Sigma(\omega)$ and $\Sigma(\omega)$ itself is linear in $\omega$. As a consequence, the function 
\[
\omega\longmapsto \EE_{X\sim \sum_{l=1}^L\omega_l\TT^{(l)}}\|X-PX\|_2^2 - \min_{P'\in \Pc^k}\EE_{X\sim \sum_{l=1}^L\omega_l\TT^{(l)}}\|X-P'X\|_2^2,
\]
is convex in $\omega$, since the first term is linear in $\omega$, while the second term is concave and appears with a minus sign.

Therefore, after taking maximization over $\omega\in \Delta^L$, the above convex function attains its maximum at an extreme point of $\Delta^L$, i.e., at one of the vertices $\{e_1,\dots, e_L\}$. It follows that
\[
\max_{\QQ\in \mathcal{C}}{\rm Regret}_\QQ(P) = \max_{l\in [L]} \left\{\EE_{X\sim \TT^{(l)}}\|X-PX\|_2^2 - \min_{P'\in \Pc^k}\EE_{X\sim \TT^{(l)}}\|X-P'X\|_2^2\right\} = \max_{l\in [L]}{\rm Regret}_{\TT^{(l)}}(P).
\]
Taking the minimization over $P\in \Pc^k$ on both sides completes the proof.

\subsection{Proof of Lemma \ref{thm: MP conv general}}
\label{proof: theorem MP general}

The proof follows the standard extragradient analysis for smooth
convex-concave minimax problems, which is outlined as follows:
\begin{enumerate}
    \item We first collect several technical lemmas concerning Bregman geometry and
the Mirror-Prox updates.
    \item We then show that each iteration satisfies a one-step inequality controlling
the primal--dual gap.
    \item Finally, telescoping this inequality over iterations yields the desired
$\mathcal{O}(1/T)$ convergence rate for the ergodic averages of the midpoint iterates.
\end{enumerate}

\medskip
Now, we present the preliminaries of the analysis.
\begin{Definition}[Bregman projection]
Let $\mathcal{Z}$ be a closed convex set and let $\psi$ be a mirror map defined
on a convex domain $\mathrm{dom}(\psi)\supseteq\mathcal{Z}$, with associated Bregman divergence $D_\psi.$
For any $\nu$ in the differentiability domain of $\psi$, its Bregman projection onto
$\mathcal{Z}$ is defined as
\[
\Pi_{\mathcal{Z}}^\psi(\nu)
:= \argmin_{z\in\mathcal{Z}} D_\psi(z,\nu).
\]
\end{Definition}

The following lemma collects two standard properties of Bregman divergences.
The first identity is often referred to as the \emph{three-point identity},
while the second expresses the optimality condition of Bregman projection. We omit the proof as it essentially follows the results in \citet{bubeck2015convex}.
\begin{Lemma}[Three-point identity and projection inequality]
\label{lemma: three point}
For $x,y$ in the differentiability domain of $\psi$ and any $z$ with finite $\psi(z)$,
\begin{equation}
\left\langle \nabla \psi(x)-\nabla \psi(y),\; x-z\right\rangle 
= D_\psi(x,y)+D_\psi(z,x)-D_\psi(z,y).
\label{eq: bregman identity}
\end{equation}
Let $\Zc$ be a closed convex set on which $\psi$ is finite. If $x$ and $\Pi_{\mathcal Z}^\psi(x)$ lie in the differentiability domain of $\psi$, the following inequality holds for every $z\in\Zc$:
\[
\big\langle
\nabla\psi(\Pi_{\mathcal{Z}}^\psi(x))-\nabla\psi(x),\;
z-\Pi_{\mathcal{Z}}^\psi(x)
\big\rangle \ge 0.
\]
\end{Lemma}

The next lemma rewrites the Mirror-Prox steps as Bregman projections in the
dual space.
This gives a dual-space interpretation when the mirror gradient can be inverted. The proof is provided in Appendix \ref{proof of lemma: MP equiv}.

\begin{Lemma}[Equivalent Mirror-Prox formulation]
\label{lemma: MP equiv}
Suppose the dual points $\nu_{t+\frac12}$ and $\nu_{t+1}$ satisfying the gradient equations below exist in the differentiability domain of $\psi$. Then the first step update \eqref{eq: MP general first} can be written as
\[
z_{t+\frac12}
= \Pi_{\mathcal{Z}}^\psi(\nu_{t+\frac12}),
\qquad \textrm{with $\nu_{t+\frac{1}{2}}$ satisfies}\;
\nabla\psi(\nu_{t+\frac12})
= \nabla\psi(z_t) - \eta\, g(z_t).
\]
Similarly, the second step update \eqref{eq: MP general second} admits the form
\[
z_{t+1}
= \Pi_{\mathcal{Z}}^\psi(\nu_{t+1}),
\qquad \textrm{with $\nu_{t+1}$ satisfies}\;
\nabla\psi(\nu_{t+1})
= \nabla\psi(z_t) - \eta\, g(z_{t+\frac12}).
\]
\end{Lemma}

The following lemma summarizes the geometric properties of the mirror map
and the smoothness of the saddle-point operator on $\mathcal{Z}$. The proof is provided in Appendix \ref{proof of lemma: properties on Z}.

\begin{Lemma}
\label{lemma: properties on Z}
Under Condition~\ref{cond: mirror and smooth}, the mirror map $\psi$ is
$(\mu_{\mathcal{X}}\wedge\mu_{\mathcal{Y}})$-strongly convex with respect to
$\|\cdot\|_{\mathcal{Z}}$, i.e.,
\[
D_\psi(z,z') \ge
\frac{\mu_{\mathcal{X}}\wedge\mu_{\mathcal{Y}}}{2}
\|z-z'\|_{\mathcal{Z}}^2.
\]
Moreover, the operator $g(z)$ is $\beta$-Lipschitz with respect to
$\|\cdot\|_{\mathcal{Z}}$, where
\[
\beta
= 2\max\left\{
\frac{\beta_{11}}{a},
\frac{\beta_{22}}{b},
\frac{\beta_{12}}{\sqrt{ab}},
\frac{\beta_{21}}{\sqrt{ab}}
\right\}.
\]
\end{Lemma}

Now we are prepared to present the proof of Lemma \ref{thm: MP conv general}.
We study the midpoint iterate
$z_{t+\frac12}=(x_{t+\frac12},y_{t+\frac12})$.

\vspace{0.5em}
\noindent\textbf{Step-1: Upper bound the primal-dual gap.}

Fix $y_{t+\frac12}$.
Since $x\mapsto f(x,y_{t+\frac12})$ is convex on $\mathcal{X}$, by the
first-order condition of convexity, for any $x\in\mathcal{X}$,
\[
f(x_{t+\frac12},y_{t+\frac12})-f(x,y_{t+\frac12})
\le
\left\langle \nabla_x f(x_{t+\frac12},y_{t+\frac12}),\; x_{t+\frac12}-x\right\rangle.
\]
Similarly, fix $x_{t+\frac12}$.
Since $y\mapsto f(x_{t+\frac12},y)$ is concave on $\mathcal{Y}$, the
first-order condition of concavity implies that for any $y\in\mathcal{Y}$,
\[
f(x_{t+\frac12},y)-f(x_{t+\frac12},y_{t+\frac12})
\le
-\left\langle \nabla_y f(x_{t+\frac12},y_{t+\frac12}),\; y_{t+\frac12}-y\right\rangle.
\]
Adding the above two results yields, for any $(x,y)\in\mathcal{X}\times\mathcal{Y}$,
\[
f(x_{t+\frac12},y) - f(x,y_{t+\frac12})
\le
\left\langle \nabla_x f(x_{t+\frac12},y_{t+\frac12}),\; x_{t+\frac12}-x\right\rangle
-\left\langle \nabla_y f(x_{t+\frac12},y_{t+\frac12}),\; y_{t+\frac12}-y\right\rangle.
\]
Using the definition $g(z)=(\nabla_x f(x,y),-\nabla_y f(x,y))$ in
\eqref{eq: g(z) def}, the right-hand side equals
$\langle g(z_{t+\frac12}), z_{t+\frac12}-z\rangle$ and thus
\begin{equation}
f(x_{t+\frac12},y) - f(x,y_{t+\frac12})
\le
\left\langle g(z_{t+\frac12}),\; z_{t+\frac12}-z\right\rangle.
\label{eq: dual gap fixed point}
\end{equation}
We next decompose the RHS of \eqref{eq: dual gap fixed point}. Note that
\[
\left\langle g(z_{t+\frac12}), z_{t+\frac12}-z\right\rangle
=
\left\langle g(z_{t+\frac12}), z_{t+1}-z\right\rangle
+
\left\langle g(z_{t+\frac12}), z_{t+\frac12}-z_{t+1}\right\rangle.
\]
Next, add and subtract $g(z_t)$ inside the second inner product:
\[
\begin{aligned}
\left\langle g(z_{t+\frac12}), z_{t+\frac12}-z_{t+1}\right\rangle
&=
\left\langle g(z_t), z_{t+\frac12}-z_{t+1}\right\rangle
+
\left\langle g(z_{t+\frac12})-g(z_t), z_{t+\frac12}-z_{t+1}\right\rangle.
\end{aligned}
\]
Therefore,
\begin{equation}
\begin{aligned}
\left\langle g(z_{t+\frac12}), z_{t+\frac12}-z\right\rangle
&=
\left\langle g(z_{t+\frac12}), z_{t+1}-z\right\rangle
+
\left\langle g(z_t), z_{t+\frac12}-z_{t+1}\right\rangle
+\left\langle g(z_{t+\frac12})-g(z_t), z_{t+\frac12}-z_{t+1}\right\rangle\\
&=: \textrm{Term-1} + \textrm{Term-2} + \textrm{Term-3}.
\end{aligned}
\label{eq: key equality}
\end{equation}

\vspace{0.5em}
\noindent\textbf{Step-2: Upper bound Term-1.}

By the first-order optimality condition for the second prox update \eqref{eq: MP general second}, for every $z\in\mathcal Z$,
\[
\left\langle\eta g(z_{t+\frac12})+\nabla\psi(z_{t+1})-\nabla\psi(z_t),\;z-z_{t+1}\right\rangle\geq0.
\]
Rearranging yields
\[
\left\langle g(z_{t+\frac12}),z_{t+1}-z\right\rangle
\leq\frac1\eta\left\langle\nabla\psi(z_t)-\nabla\psi(z_{t+1}),z_{t+1}-z\right\rangle.
\]
This argument does not require an inverse of the mirror gradient.
Now apply the three-point identity \eqref{eq: bregman identity} with
$(x,y,z)=(z_{t+1},z_t,z)$:
\[
\left\langle \nabla\psi(z_{t+1})-\nabla\psi(z_t),\; z_{t+1}-z\right\rangle 
=
D_\psi(z_{t+1},z_t)+D_\psi(z,z_{t+1})-D_\psi(z,z_t).
\]
Substituting this identity into the previous bound gives
\begin{equation}
\left\langle g(z_{t+\frac12}), z_{t+1}-z\right\rangle
\le
\frac{1}{\eta}
\big[
D_\psi(z,z_t)-D_\psi(z,z_{t+1})-D_\psi(z_{t+1},z_t)
\big].
\label{eq: term 1 key equality}
\end{equation}

\vspace{0.5em}
\noindent\textbf{Step-3: Upper bound Term-2.}

The first-order optimality condition for \eqref{eq: MP general first}, evaluated at $z_{t+1}$, gives
\[
\left\langle\eta g(z_t)+\nabla\psi(z_{t+\frac12})-\nabla\psi(z_t),\;z_{t+1}-z_{t+\frac12}\right\rangle\geq0.
\]
Hence
\[
\left\langle g(z_t),z_{t+\frac12}-z_{t+1}\right\rangle
\leq\frac1\eta\left\langle\nabla\psi(z_t)-\nabla\psi(z_{t+\frac12}),z_{t+\frac12}-z_{t+1}\right\rangle.
\]
Apply \eqref{eq: bregman identity} with $(x,y,z)=(z_{t+\frac12},z_t,z_{t+1})$:
\[
\left\langle \nabla\psi(z_t)-\nabla\psi(z_{t+\frac12}),\; z_{t+\frac12}-z_{t+1}\right\rangle
=
D_\psi(z_{t+1},z_t)-D_\psi(z_{t+1},z_{t+\frac12})-D_\psi(z_{t+\frac12},z_t).
\]
Therefore,
\[
\left\langle g(z_t), z_{t+\frac12}-z_{t+1}\right\rangle
\le
\frac{1}{\eta}
\Big[
D_\psi(z_{t+1},z_t)-D_\psi(z_{t+1},z_{t+\frac12})-D_\psi(z_{t+\frac12},z_t)
\Big].
\]
Finally, by strong convexity of $\psi$ on $\mathcal{Z}$ (Lemma~\ref{lemma: properties on Z}),
\[
D_\psi(z_{t+1},z_{t+\frac12})
\ge \frac{\mu}{2}\|z_{t+1}-z_{t+\frac12}\|_{\mathcal{Z}}^2,
\qquad
D_\psi(z_{t+\frac12},z_t)
\ge \frac{\mu}{2}\|z_{t+\frac12}-z_t\|_{\mathcal{Z}}^2,
\]
where $\mu=\mu_{\mathcal{X}}\wedge\mu_{\mathcal{Y}}$.
Substituting these lower bounds yields
\begin{equation}
\left\langle g(z_t), z_{t+\frac12}-z_{t+1}\right\rangle
\le
\frac{1}{\eta}
\left[
D_\psi(z_{t+1},z_t)
-\frac{\mu}{2}\|z_{t+1}-z_{t+\frac12}\|_{\mathcal{Z}}^2
-\frac{\mu}{2}\|z_{t+\frac12}-z_t\|_{\mathcal{Z}}^2
\right].
\label{eq: term 2 key equality}
\end{equation}

\vspace{0.5em}
\noindent\textbf{Step-4: Upper bound Term-3.}

By Cauchy--Schwarz,
\[
\left\langle g(z_{t+\frac12})-g(z_t),\; z_{t+\frac12}-z_{t+1}\right\rangle
\le
\|g(z_{t+\frac12})-g(z_t)\|_{\mathcal{Z},*}\;
\|z_{t+\frac12}-z_{t+1}\|_{\mathcal{Z}}.
\]
By the $\beta$-Lipschitz property of $g$ (Lemma~\ref{lemma: properties on Z}),
\[
\|g(z_{t+\frac12})-g(z_t)\|_{\mathcal{Z},*}
\le
\beta\|z_{t+\frac12}-z_t\|_{\mathcal{Z}}.
\]
Combining the previous two inequalities gives
\[
\left\langle g(z_{t+\frac12})-g(z_t),\; z_{t+\frac12}-z_{t+1}\right\rangle
\le
\beta\|z_{t+\frac12}-z_t\|_{\mathcal{Z}}\;
\|z_{t+\frac12}-z_{t+1}\|_{\mathcal{Z}}.
\]
Applying Young’s inequality $ab\le \frac12 a^2+\frac12 b^2$ yields
\begin{equation}
\left\langle g(z_{t+\frac12})-g(z_t),\; z_{t+\frac12}-z_{t+1}\right\rangle
\le
\frac{\beta}{2}\|z_{t+\frac12}-z_{t+1}\|_{\mathcal{Z}}^2
+
\frac{\beta}{2}\|z_{t+\frac12}-z_t\|_{\mathcal{Z}}^2.
\label{eq: term 3 key equality}
\end{equation}

\vspace{0.5em}
\noindent\textbf{Step-5: Combine three terms and complete the proof.}

Plugging \eqref{eq: term 1 key equality}, \eqref{eq: term 2 key equality},
and \eqref{eq: term 3 key equality} into \eqref{eq: key equality} yields
\[
\begin{aligned}
\left\langle g(z_{t+\frac12}), z_{t+\frac12}-z\right\rangle
&\le
\frac{1}{\eta}\big[D_\psi(z,z_t)-D_\psi(z,z_{t+1})\big]\\
&\quad+
\left(\frac{\beta}{2}-\frac{\mu}{2\eta}\right)\|z_{t+\frac12}-z_{t+1}\|_{\mathcal{Z}}^2
+
\left(\frac{\beta}{2}-\frac{\mu}{2\eta}\right)\|z_{t+\frac12}-z_t\|_{\mathcal{Z}}^2.
\end{aligned}
\]
Choosing $\eta=\mu/\beta$ yields
\[
\left\langle g(z_{t+\frac12}), z_{t+\frac12}-z\right\rangle
\le
\frac{D_\psi(z,z_t)-D_\psi(z,z_{t+1})}{\eta}.
\]
Combining with \eqref{eq: dual gap fixed point}, we obtain
\[
f(x_{t+\frac12},y)-f(x,y_{t+\frac12})
\le
\frac{D_\psi(z,z_t)-D_\psi(z,z_{t+1})}{\eta}.
\]

Summing over $t=0,\dots,T-1$ gives
\begin{equation}
    \sum_{t=0}^{T-1}\big[f(x_{t+\frac12},y)-f(x,y_{t+\frac12})\big]
\le
\frac{D_\psi(z,z_0)-D_\psi(z,z_T)}{\eta}
\le
\frac{D_\psi(z,z_0)}{\eta},
\label{eq: summation over T}
\end{equation}
where we used the fact that $D_\psi(z,z_T)\ge 0$.

Now we bound $D_\psi(z,z_0)$.
Since $z_0\in\argmin_{z\in\mathcal{Z}\cap \Dc_\Zc}\psi(z)$ and $\psi$ is convex,
the first-order optimality condition implies that for all $z\in\mathcal{Z}\cap \Dc_\Zc$,
\[
\langle \nabla\psi(z_0), z-z_0\rangle \ge 0.
\]
Thus,
\[
\begin{aligned}
D_\psi(z,z_0)
&=
\psi(z)-\psi(z_0)-\langle \nabla\psi(z_0), z-z_0\rangle\\
&\le \psi(z)-\psi(z_0)
\le \sup_{u\in\mathcal{Z}}\psi(u)-\inf_{u\in\mathcal{Z}}\psi(u).
\end{aligned}
\]
Recalling $\psi(z)=a\psi_{\mathcal{X}}(x)+b\psi_{\mathcal{Y}}(y)$, we obtain
\[
\sup_{u\in\mathcal{Z}}\psi(u)-\inf_{u\in\mathcal{Z}}\psi(u)
\le aR_{\mathcal{X}}^2 + bR_{\mathcal{Y}}^2,
\]
where $R_{\mathcal{X}}^2$ and $R_{\mathcal{Y}}^2$ are defined in
Condition~\ref{cond: mirror and smooth}.
Hence,
\[
D_\psi(z,z_0)\le aR_{\mathcal{X}}^2 + bR_{\mathcal{Y}}^2.
\]

Together with \eqref{eq: summation over T}, dividing the telescoping inequality by $T$ yields
\[
\frac{1}{T}\sum_{t=0}^{T-1}\big[f(x_{t+\frac12},y)-f(x,y_{t+\frac12})\big]
\le
\frac{aR_{\mathcal{X}}^2+bR_{\mathcal{Y}}^2}{\eta T}.
\]
With $\eta=\mu/\beta = \frac{\mu_\Xc\wedge \mu_\Yc}{\beta}$, this becomes
\[
\frac{1}{T}\sum_{t=0}^{T-1}\big[f(x_{t+\frac12},y)-f(x,y_{t+\frac12})\big]
\le
\frac{\beta(aR_{\mathcal{X}}^2+bR_{\mathcal{Y}}^2)}{(\mu_\Xc\wedge\mu_\Yc)\cdot  T}.
\]

By convexity of $f(\cdot,y)$ and concavity of $f(x,\cdot)$, Jensen's inequality
implies
\[
f\!\left(\frac{1}{T}\sum_{t=0}^{T-1}x_{t+\frac12},\, y\right)
-
f\!\left(x,\, \frac{1}{T}\sum_{t=0}^{T-1}y_{t+\frac12}\right)
\le
\frac{\beta(aR_{\mathcal{X}}^2+bR_{\mathcal{Y}}^2)}{(\mu_\Xc\wedge\mu_\Yc)\cdot  T}.
\]
Since the inequality holds for all $(x,y)\in\mathcal{X}\times\mathcal{Y}$, we obtain
\[
\max_{y\in\mathcal{Y}}
f\!\left(\frac{1}{T}\sum_{t=0}^{T-1}x_{t+\frac12},\, y\right)
-
\min_{x\in\mathcal{X}}
f\!\left(x,\, \frac{1}{T}\sum_{t=0}^{T-1}y_{t+\frac12}\right)
\le
\frac{\beta(aR_{\mathcal{X}}^2+bR_{\mathcal{Y}}^2)}{(\mu_\Xc\wedge\mu_\Yc)\cdot  T}.
\]
Substituting the expression of $\beta$ in Lemma~\ref{lemma: properties on Z}
completes the proof.

\subsection{Proof of Lemma \ref{lemma: obj val diff}}
\label{proof of lemma: obj val diff}
Fix a source $l$ and drop the superscript, writing $X_i$, $\Sigma$, $\widehat{\Sigma}$, and $n$ for $X_i^{(l)}$, $\Sigma^{(l)}$, $\widehat{\Sigma}^{(l)}$, and $n_l$. Write also $\mathbf{r}:=\mathbf{r}(\Sigma)$. If $\Sigma=0$, then $\EE\|X_i\|_2^2={\rm Tr}(\Sigma)=0$, so $X_i=0$ almost surely and $\widehat\Sigma=\Sigma=0$; the desired bound is trivial. We may therefore assume $\Sigma\neq0$, in which case $\mathbf r={\rm Tr}(\Sigma)/\|\Sigma\|_{\rm op}$. We show that for any $t\geq 0$, with probability at least $1 - 2e^{-t}$,
\begin{equation}
    \|\widehat{\Sigma} - \Sigma\|_{\rm op}\leq C_\sigma\|\Sigma\|_{\rm op}\left(\sqrt{\frac{\mathbf{r}+t}{n}} + \frac{\mathbf{r}+t}{n}\right).
    \label{eq: proof lemma cov effrank}
\end{equation}
The three claims of the lemma then follow. The first is from $n_l\geq n$ and $\mathbf{r}(\Sigma^{(l)})\leq\mathbf{r}_{\max}$ together with a union bound over $l\in[L]$. The second is from the triangle inequality $|\|\widehat{\Sigma}\|_{\rm op} - \|\Sigma\|_{\rm op}|\leq\|\widehat{\Sigma} - \Sigma\|_{\rm op}$. And the third is from H\"older's inequality,
\[
\sup_{M\in \Fc^k}\left|\langle\widehat{\Sigma} - \Sigma, M\rangle\right| \leq
\sup_{M\in \Fc^k}\|M\|_{*} \|\widehat{\Sigma} - \Sigma\|_{\rm op}  = k\|\widehat{\Sigma} - \Sigma\|_{\rm op},
\]
where the last equality holds because $\|M\|_* = {\rm Tr}(M) = k$ for every $M\in \Fc^k$. For Gaussian data, \eqref{eq: proof lemma cov effrank} is the concentration bound of \citet{koltchinskii2017concentration}, who also show that its rate cannot be improved. We give a short proof under Condition~\ref{cond: subgauss} based on the matrix deviation inequality of \citet[Theorem 1.4]{liaw2017simple}.

\medskip
\noindent\textbf{Step 1: reduction to isotropic sub-Gaussian vectors.} 

Let $S := {\rm col}(\Sigma)\subseteq\RR^d$, let $p := \dim S = {\rm rank}(\Sigma)$, let $U\in\mathcal{O}^{d\times p}$ have columns forming an orthonormal basis of $S$, so that $UU^\intercal = \Pi_S$ is the orthogonal projection onto $S$, and let $\Sigma^{+}$ denote the Moore--Penrose pseudo-inverse of $\Sigma$. Since $\EE\|({\bf I}_d - \Pi_S)X_i\|_2^2 = {\rm Tr}\big(({\bf I}_d - \Pi_S)\Sigma\big) = 0$, we have $X_i\in S$ almost surely. Define $Z_i := U^\intercal(\Sigma^{+})^{1/2}X_i\in\RR^p$. Since $\Sigma^{1/2}UU^\intercal = \Sigma^{1/2}\Pi_S = \Sigma^{1/2}$ and $\Sigma^{1/2}(\Sigma^{+})^{1/2} = \Pi_S$,
\[
\Sigma^{1/2}UZ_i = \Sigma^{1/2}UU^\intercal(\Sigma^{+})^{1/2}X_i = \Sigma^{1/2}(\Sigma^{+})^{1/2}X_i = \Pi_SX_i = X_i\quad\text{almost surely},
\]
where only the last equality uses $X_i\in S$. Moreover, $\EE[Z_iZ_i^\intercal] = U^\intercal(\Sigma^{+})^{1/2}\Sigma(\Sigma^{+})^{1/2}U = U^\intercal\Pi_SU = {\bf I}_p$, so that $Z_i$ is isotropic, and for every $z\in\RR^p$, $\langle Z_i, z\rangle = \langle X_i, (\Sigma^{+})^{1/2}Uz\rangle$, whence by Condition~\ref{cond: subgauss}
\[
\|\langle Z_i, z\rangle\|_{\psi_2}\leq\sigma\Big(z^\intercal U^\intercal(\Sigma^{+})^{1/2}\Sigma(\Sigma^{+})^{1/2}Uz\Big)^{1/2} = \sigma\big(z^\intercal U^\intercal\Pi_SUz\big)^{1/2} = \sigma\|z\|_2 .
\]
Hence the $n\times p$ matrix $B$ whose rows are $Z_1^\intercal,\dots,Z_n^\intercal$ has independent, isotropic, sub-Gaussian rows with $\max_i\|Z_i\|_{\psi_2}\leq\sigma$. For a unit vector $x\in\RR^d$, let $y := U^\intercal\Sigma^{1/2}x\in\RR^p$; then $\langle X_i, x\rangle = \langle Z_i, y\rangle$ and $\|y\|_2 = \|\Sigma^{1/2}x\|_2$ because $\Sigma^{1/2}x\in S$, and therefore
\begin{equation}
    x^\intercal(\widehat{\Sigma} - \Sigma)x = \frac1n\sum_{i=1}^n\langle X_i, x\rangle^2 - x^\intercal\Sigma x = \frac{1}{n}\|By\|_2^2 - \|y\|_2^2 .
    \label{eq: proof lemma quad form}
\end{equation}

\medskip
\noindent\textbf{Step 2: the matrix deviation inequality on an ellipsoid.} 

Let $\mathcal{T} := \{U^\intercal\Sigma^{1/2}x: \|x\|_2\leq 1\}\subseteq\RR^p$. Its radius is ${\rm rad}(\mathcal{T}) := \sup_{y\in\mathcal{T}}\|y\|_2 = \|\Sigma\|_{\rm op}^{1/2}$, and its Gaussian width, defined with $g\sim\mathcal{N}(0, {\bf I}_p)$, is
\[
w(\mathcal{T}) := \EE\sup_{y\in\mathcal{T}}\langle y, g\rangle = \EE\|\Sigma^{1/2}Ug\|_2\leq\big(\EE\|\Sigma^{1/2}Ug\|_2^2\big)^{1/2} = \sqrt{{\rm Tr}(U^\intercal\Sigma U)} = \sqrt{{\rm Tr}(\Sigma)} = \|\Sigma\|_{\rm op}^{1/2}\sqrt{\mathbf{r}},
\]
where ${\rm Tr}(U^\intercal\Sigma U) = {\rm Tr}(\Sigma\Pi_S) = {\rm Tr}(\Sigma)$. The matrix deviation inequality \citep[Theorem 1.4]{liaw2017simple} states that, for an $n\times p$ matrix $B$ with independent isotropic sub-Gaussian rows of sub-Gaussian norm at most $\sigma$, and for any bounded set $\mathcal{T}\subseteq\RR^p$ and any $u\geq 0$, with probability at least $1 - 2e^{-u^2}$,
\[
\sup_{y\in\mathcal{T}}\Big|\|By\|_2 - \sqrt{n}\,\|y\|_2\Big|\leq C\sigma^2\big[w(\mathcal{T}) + u\cdot{\rm rad}(\mathcal{T})\big],
\]
where $C>0$ is a universal constant. Taking $u = \sqrt{t}$ and dividing by $\sqrt n$, with probability at least $1 - 2e^{-t}$, for every $y\in\mathcal{T}$,
\[
\Big|\frac{\|By\|_2}{\sqrt n} - \|y\|_2\Big|\leq\epsilon := C\sigma^2\|\Sigma\|_{\rm op}^{1/2}\left(\sqrt{\frac{\mathbf{r}}{n}} + \sqrt{\frac{t}{n}}\right)\leq C\sigma^2\|\Sigma\|_{\rm op}^{1/2}\sqrt{\frac{2(\mathbf{r}+t)}{n}} .
\]

\medskip
\noindent\textbf{Step 3: from norms to quadratic forms.} 

On the same event, for every $y\in\mathcal{T}$, since $\|y\|_2\leq\|\Sigma\|_{\rm op}^{1/2}$,
\[
\Big|\frac{\|By\|_2^2}{n} - \|y\|_2^2\Big| = \Big|\frac{\|By\|_2}{\sqrt n} - \|y\|_2\Big|\cdot\Big(\frac{\|By\|_2}{\sqrt n} + \|y\|_2\Big)\leq\epsilon\,(2\|y\|_2 + \epsilon)\leq 2\epsilon\|\Sigma\|_{\rm op}^{1/2} + \epsilon^2 .
\]
By \eqref{eq: proof lemma quad form}, the left-hand side ranges over $|x^\intercal(\widehat{\Sigma} - \Sigma)x|$ for unit vectors $x\in\RR^d$, whose supremum is $\|\widehat{\Sigma} - \Sigma\|_{\rm op}$ because $\widehat{\Sigma} - \Sigma$ is symmetric. Substituting the bound on $\epsilon$,
\[
\|\widehat{\Sigma} - \Sigma\|_{\rm op}\leq 2\sqrt{2}\,C\sigma^2\|\Sigma\|_{\rm op}\sqrt{\frac{\mathbf{r}+t}{n}} + 2C^2\sigma^4\|\Sigma\|_{\rm op}\,\frac{\mathbf{r}+t}{n},
\]
which is \eqref{eq: proof lemma cov effrank} with $C_\sigma := 2\sqrt2\,C\max\{\sigma^2, C\sigma^4\}\leq 2\sqrt2\,C\max\{1, C\}\cdot\max\{\sigma^2,\sigma^4\}$, of the form stated in the lemma with the universal constant $2\sqrt2\,C\max\{1,C\}$ in place of $C$.

\subsection{Proof of Lemma \ref{lem: stablepca_condition_verify}}
\label{proof of lem: stablepca_condition_verify}

\noindent\textbf{Strong Convexity of $\psi_1(M)$.}

When $k=1$, Section 4.3 of \citet{bubeck2015convex} states that $\psi_1(X)$ is $\frac12$-strongly convex on $X\in \Fc^1$ with respect to the Schatten-1 (nuclear) norm $\|\cdot\|_*$. This means that
\begin{equation}
    \label{eq:bubeck_trace1_sc}
    D_{\psi_1}(X,Y) \geq \frac{1}{4}\|X-Y\|_*^2,\qquad \forall X,Y\in \Fc^1.
\end{equation}
For general $k\geq 1$, fix $M,M'\in \Fc^k$, we define
\[
X:= \frac{1}{k}M, \quad Y:= \frac{1}{k}M',
\]
then $X,Y\in \Fc^1$ satisfies that
\begin{equation}
    \label{eq:norm_scaling}
    \|X-Y\|_* = \frac{1}{k}\|M - M'\|_*.
\end{equation}

Moreover, recall the Bregman divergence associated with $\psi_1$ in \eqref{eq: Breg separate}:
\[
D_{\psi_1}(M, M') = {\rm Tr}\left(M(\log M - \log M')\right) = {\rm Tr}\left(kX\big(\log(kX)-\log(kY)\big)\right) = k D_{\psi_1}(X,Y).
\]
Together with \eqref{eq:bubeck_trace1_sc} and \eqref{eq:norm_scaling}, we obtain
\[
D_{\psi_1}(M,M') = kD_{\psi_1}(X,Y)\geq \frac{k}{4}\|X-Y\|_*^2 = \frac{1}{4k}\|M-M'\|_*^2.
\]
This establishes that $\psi_1$ is $\frac{1}{2k}$-strongly convex over $M\in \mathcal{F}^k$,
with respect to the nuclear norm $\|M\|_*$.

Note that $\psi_1(M) = \sum_i \lambda_i \log \lambda_i \leq 0$ is always true since $\lambda_i\in [0,1]$, where $\lambda_i$ denotes the $i$-th eigenvalue of $M$. Therefore, the maximum is attained at any vertex when $M$ is a rank-$k$ projector, with $\sup_{M\in \Fc^k}\psi_1(M) = 0$. 

Due to the convexity of $\psi_1(M)$, it is straightforward to see that the minimum of $\psi_1(M)=\sum_i \lambda_i \log \lambda_i$ is taken at when all $\lambda_i=\frac{k}{d}$, with $\inf_{M\in \Fc^k}\psi_1(M) = k\log (k/d)$. Therefore,
\[
R^2_{\Fc^k} = \sup_{M\in \Fc^k}\psi_1(M) - \inf_{M\in \Fc^k}\psi_1(M) = k\log \frac{d}{k}.
\]

\vspace{0.5em}
\noindent\textbf{Strong Convexity of $\psi_2(\omega)$.}

Recall in \eqref{eq: Breg separate}, we have
\[
\begin{aligned}
    D_\psi(\omega,\omega') =\sum_{l=1}^L \omega_l \log \frac{\omega_l}{\omega_l'} =: {\rm KL}(\omega\|\omega').
\end{aligned}
\]
Pinsker's inequality states that for any probability distributions $p,q$ on the simplex,
\[
{\rm KL}(p\|q) \geq \frac{1}{2}\|p-q\|_1^2.
\]
Applying it to $\omega,\omega'$ gives:
\[
D_{\psi_2}(\omega,\omega') = {\rm KL}(\omega\|\omega') \geq \frac{1}{2}\|\omega-\omega'\|_1^2,
\]
which indicates that $\psi_2$ is $1$-strongly convex over $\omega\in \Delta^L$.

Note that $\psi_2(\omega)\leq 0$ is always true, since $x\log x\leq 0$ for $x\in [0,1]$, and the maximum is attained at any vertex. The minimum is attained at the uniform distribution $\frac{1}{L}{\bf 1}_L$. Therefore,
\[
R^2_{\Delta^L}:= \sup_{\omega\in \Delta^L}\psi_2(\omega) - \inf_{\omega\in \Delta^L}\psi_2(\omega) = \log L.
\]

\vspace{0.5em}
\noindent\textbf{Smoothness of $f(M,\omega)$}

We first compute the partial gradients:
\[
\nabla_M f(M,\omega) = -\sum_{l=1}^L \omega_l\,\widehat\Sigma^{(l)},
\qquad
\nabla_\omega f(M,\omega)
=
\Big(
-\langle \widehat\Sigma^{(1)},M\rangle,\ldots,
-\langle \widehat\Sigma^{(L)},M\rangle
\Big)^\intercal .
\]
The dual norms of nuclear norm $\|M\|_*$ and $\ell_1$ norm $\|\omega\|_1$ are operator norm $\|M\|_{\rm op}$ and $\ell_\infty$-norm $\|\omega\|_\infty$,
respectively. Since $\nabla_M f(M,\omega)$ does not depend on $M$ and
$\nabla_\omega f(M,\omega)$ does not depend on $\omega$, we immediately have
\[
\beta_{11}=0,
\qquad
\beta_{22}=0.
\]

Next, for $\beta_{21}$, for any $M_1,M_2\in\mathcal{F}^k$ and any $l\in[L]$,
H\"older's inequality for nuclear norms gives
\[
\big|\langle \widehat\Sigma^{(l)}, M_1-M_2\rangle\big|
\le \|\widehat\Sigma^{(l)}\|_{\rm op}\,\|M_1-M_2\|_*.
\]
Taking the maximum over $l$ yields
\[
\|\nabla_\omega f(M_1,\omega)-\nabla_\omega f(M_2,\omega)\|_\infty
\le
\beta\,\|M_1-M_2\|_*, \quad \textrm{with}\;\; \beta=\widehat{\rho}_{\max}=\max_{l\in[L]}\|\widehat{\Sigma}^{(l)}\|_{\rm op}.
\]
so we may take $\beta_{21}=\beta$.

Finally, for $\beta_{12}$, for any $\omega_1,\omega_2\in\Delta^L$,
\[
\begin{aligned}
\|\nabla_M f(M,\omega_1)-\nabla_M f(M,\omega_2)\|_{\rm op}
&=
\left\|\sum_{l=1}^L (\omega_{1l}-\omega_{2l})\,\widehat\Sigma^{(l)}\right\|_{\rm op}\\
&\le
\sum_{l=1}^L |\omega_{1l}-\omega_{2l}|\,\|\widehat\Sigma^{(l)}\|_{\rm op}\\
&\le
\beta\,\|\omega_1-\omega_2\|_1,
\end{aligned}
\]
hence $\beta_{12}=\beta$.

\subsection{Proof of Lemma \ref{lemma: MP equiv}}
\label{proof of lemma: MP equiv}
By the definition of the Bregman divergence, we have
    \[
    \begin{aligned}
         z_{t+\frac12} &= \argmin_{z\in \Zc\cap \Dc_{\Zc}}D_\psi(z, \nu_{t+\frac12}) \\
         &= \argmin_{z\in \Zc\cap \Dc_{\Zc}}\psi(z) - \psi(\nu_{t+\frac12}) - \langle\nabla\psi(\nu_{t+\frac12}), z\rangle \\
         &= \argmin_{z\in \Zc\cap \Dc_{\Zc}}\psi(z)- \langle\nabla\psi(\nu_{t+\frac12}), z\rangle &&\textrm{as $\psi(\nu_{t+\frac12})$ is independent of $z$} \\
         &= \argmin_{z\in \Zc\cap \Dc_{\Zc}}\psi(z) - \langle\nabla\psi(z_t)-\eta g(z_t),z\rangle && \textrm{the expression of $\nabla\psi(\nu_{t+\frac12})$}\\
         &= \argmin_{z\in \Zc\cap \Dc_{\Zc}} D_\psi(z, z_t) + \eta\cdot \langle g(z_t), z\rangle.
    \end{aligned}
    \]
    Similarly, we shall show that
    \[
    \begin{aligned}
        z_{t+1} &= \argmin_{z\in \Zc\cap \Dc_{\Zc}}D_\psi(z,\nu_{t+1}) \\
        &= \argmin_{z\in \Zc \cap \Dc_{\Zc}}\psi(z) - \langle\nabla \psi(\nu_{t+1}), z \rangle\\
        &= \argmin_{z\in \Zc\cap \Dc_{\Zc}}\psi(z) - \langle \nabla \psi(z_t) - \eta \cdot g(z_{t+\frac12}), z\rangle \\
        &= \argmin_{z\in \Zc\cap \Dc_{\Zc}} D_\psi(z, z_t) + \eta\cdot \langle g(z_{t+\frac{1}{2}}), z\rangle.
    \end{aligned}
    \]

\subsection{Proof of Lemma \ref{lemma: properties on Z}}
\label{proof of lemma: properties on Z}

\vspace{0.5em}
\noindent\textbf{Strong convexity of $\psi$.}

We first show the strong convexity of $\psi(z)$.
By the $\mu_{\mathcal{X}}$-strong convexity of $\psi_{\mathcal{X}}$ on $\mathcal{X}$
with respect to $\|\cdot\|_{\mathcal{X}}$, for any $x,x'\in\mathcal{X}$,
\[
\psi_{\mathcal{X}}(x')
\ge
\psi_{\mathcal{X}}(x)
+
\langle \nabla \psi_{\mathcal{X}}(x),\, x'-x\rangle
+
\frac{\mu_{\mathcal{X}}}{2}\|x'-x\|_{\mathcal{X}}^{2}.
\]
Similarly, by the $\mu_{\mathcal{Y}}$-strong convexity of $\psi_{\mathcal{Y}}$ on $\mathcal{Y}$
with respect to $\|\cdot\|_{\mathcal{Y}}$, for any $y,y'\in\mathcal{Y}$,
\[
\psi_{\mathcal{Y}}(y')
\ge
\psi_{\mathcal{Y}}(y)
+
\langle \nabla \psi_{\mathcal{Y}}(y),\, y'-y\rangle
+
\frac{\mu_{\mathcal{Y}}}{2}\|y'-y\|_{\mathcal{Y}}^{2}.
\]

Let $z=(x,y)$ and $z'=(x',y')$, and define
\[
\psi(z) := a\,\psi_{\mathcal{X}}(x)+ b\,\psi_{\mathcal{Y}}(y).
\]
Multiplying the two inequalities above by $a$ and $b$ respectively and summing,
we obtain
\[
\begin{aligned}
\psi(z')
&=
a\psi_{\mathcal{X}}(x') + b\psi_{\mathcal{Y}}(y')\\
&\ge
a\psi_{\mathcal{X}}(x) + a\langle \nabla\psi_{\mathcal{X}}(x),\,x'-x\rangle
+ \frac{a\mu_{\mathcal{X}}}{2}\|x'-x\|_{\mathcal{X}}^{2}\\
&\quad+
b\psi_{\mathcal{Y}}(y) + b\langle \nabla\psi_{\mathcal{Y}}(y),\,y'-y\rangle
+ \frac{b\mu_{\mathcal{Y}}}{2}\|y'-y\|_{\mathcal{Y}}^{2}\\
&=
\psi(z)
+
\langle \nabla\psi(z),\, z'-z\rangle
+
\frac{a\mu_{\mathcal{X}}}{2}\|x'-x\|_{\mathcal{X}}^{2}
+
\frac{b\mu_{\mathcal{Y}}}{2}\|y'-y\|_{\mathcal{Y}}^{2}.
\end{aligned}
\]
Using the definition of $\|\cdot\|_{\mathcal{Z}}$,
\[
\|z'-z\|_{\mathcal{Z}}^{2} = a\|x'-x\|_{\mathcal{X}}^{2} + b\|y'-y\|_{\mathcal{Y}}^{2},
\]
we further have
\[
\frac{a\mu_{\mathcal{X}}}{2}\|x'-x\|_{\mathcal{X}}^{2}
+
\frac{b\mu_{\mathcal{Y}}}{2}\|y'-y\|_{\mathcal{Y}}^{2}
\ge
\frac{\mu_{\mathcal{X}}\wedge\mu_{\mathcal{Y}}}{2}
\left(a\|x'-x\|_{\mathcal{X}}^{2} + b\|y'-y\|_{\mathcal{Y}}^{2}\right)
=
\frac{\mu_{\mathcal{X}}\wedge\mu_{\mathcal{Y}}}{2}\|z'-z\|_{\mathcal{Z}}^{2}.
\]
Therefore,
\[
\psi(z')
\ge
\psi(z)
+
\langle \nabla\psi(z),\, z'-z\rangle
+
\frac{\mu_{\mathcal{X}}\wedge\mu_{\mathcal{Y}}}{2}\|z'-z\|_{\mathcal{Z}}^{2},
\]
which shows that $\psi$ is $(\mu_{\mathcal{X}}\wedge\mu_{\mathcal{Y}})$-strongly convex
with respect to $\|\cdot\|_{\mathcal{Z}}$.
As a consequence, the associated Bregman divergence satisfies
\[
D_{\psi}(z,z')
\ge
\frac{\mu_{\mathcal{X}}\wedge\mu_{\mathcal{Y}}}{2}\|z-z'\|_{\mathcal{Z}}^{2}.
\]

\vspace{0.5em}
\noindent\textbf{Lipschitz continuity of $g(z)$.}

Next, we prove that $g$ is $\beta$-Lipschitz.
The dual norm associated with $\|\cdot\|_{\mathcal{Z}}$ is
\[
\|u\|_{\mathcal{Z},*}
=
\sqrt{\frac{1}{a}\|u_x\|_{\mathcal{X},*}^{2} + \frac{1}{b}\|u_y\|_{\mathcal{Y},*}^{2}},
\qquad\text{for }u=(u_x,u_y)\in \mathcal{X}^*\times \mathcal{Y}^*.
\]
Let $z=(x,y)$ and $z'=(x',y')$.
Recall $g(z)=(\nabla_x f(x,y),-\nabla_y f(x,y))$, hence
\[
g(z)-g(z')
=
\big(\nabla_x f(x,y)-\nabla_x f(x',y'),\;
-(\nabla_y f(x,y)-\nabla_y f(x',y'))\big).
\]
By definition of $\|\cdot\|_{\mathcal{Z},*}$,
\begin{equation}
\|g(z)-g(z')\|_{\mathcal{Z},*}^{2}
=
\frac{1}{a}\|\nabla_x f(x,y)-\nabla_x f(x',y')\|_{\mathcal{X},*}^{2}
+
\frac{1}{b}\|\nabla_y f(x,y)-\nabla_y f(x',y')\|_{\mathcal{Y},*}^{2}.
\label{eq: lip g(z) main ineq}
\end{equation}

Using the triangle inequality,
\[
\|\nabla_x f(x,y)-\nabla_x f(x',y')\|_{\mathcal{X},*}
\le
\|\nabla_x f(x,y)-\nabla_x f(x',y)\|_{\mathcal{X},*}
+
\|\nabla_x f(x',y)-\nabla_x f(x',y')\|_{\mathcal{X},*}.
\]
By Condition~\ref{cond: mirror and smooth},
\[
\|\nabla_x f(x,y)-\nabla_x f(x',y)\|_{\mathcal{X},*}\le \beta_{11}\|x-x'\|_{\mathcal{X}},
\qquad
\|\nabla_x f(x',y)-\nabla_x f(x',y')\|_{\mathcal{X},*}\le \beta_{12}\|y-y'\|_{\mathcal{Y}},
\]
and therefore
\[
\|\nabla_x f(x,y)-\nabla_x f(x',y')\|_{\mathcal{X},*}
\le
\beta_{11}\|x-x'\|_{\mathcal{X}}+\beta_{12}\|y-y'\|_{\mathcal{Y}}.
\]
Similarly,
\[
\|\nabla_y f(x,y)-\nabla_y f(x',y')\|_{\mathcal{Y},*}
\le
\beta_{21}\|x-x'\|_{\mathcal{X}}+\beta_{22}\|y-y'\|_{\mathcal{Y}}.
\]

Substituting these bounds into \eqref{eq: lip g(z) main ineq} yields
\[
\begin{aligned}
\|g(z)-g(z')\|_{\mathcal{Z},*}^{2}
&\le
\frac{(\beta_{11}\|x-x'\|_{\mathcal{X}}+\beta_{12}\|y-y'\|_{\mathcal{Y}})^{2}}{a}
+
\frac{(\beta_{21}\|x-x'\|_{\mathcal{X}}+\beta_{22}\|y-y'\|_{\mathcal{Y}})^{2}}{b}.
\end{aligned}
\]
Using $(u+v)^2\le 2u^2+2v^2$, we obtain
\[
\begin{aligned}
\|g(z)-g(z')\|_{\mathcal{Z},*}^{2}
&\le
\frac{2\beta_{11}^{2}}{a}\|x-x'\|_{\mathcal{X}}^{2}
+
\frac{2\beta_{12}^{2}}{a}\|y-y'\|_{\mathcal{Y}}^{2}
+
\frac{2\beta_{21}^{2}}{b}\|x-x'\|_{\mathcal{X}}^{2}
+
\frac{2\beta_{22}^{2}}{b}\|y-y'\|_{\mathcal{Y}}^{2}\\
&=
2\left(\frac{\beta_{11}^{2}}{a}+\frac{\beta_{21}^{2}}{b}\right)\|x-x'\|_{\mathcal{X}}^{2}
+
2\left(\frac{\beta_{12}^{2}}{a}+\frac{\beta_{22}^{2}}{b}\right)\|y-y'\|_{\mathcal{Y}}^{2}.
\end{aligned}
\]
Now note that
\[
\frac{\beta_{11}^{2}}{a}+\frac{\beta_{21}^{2}}{b}
\le
2\max\left\{\frac{\beta_{11}^{2}}{a},\frac{\beta_{21}^{2}}{b}\right\}
=
2\max\left\{\left(\frac{\beta_{11}}{a}\right)^{2}a,\left(\frac{\beta_{21}}{\sqrt{ab}}\right)^{2}a\right\},
\]
and similarly
\[
\frac{\beta_{12}^{2}}{a}+\frac{\beta_{22}^{2}}{b}
\le
2\max\left\{\left(\frac{\beta_{12}}{\sqrt{ab}}\right)^{2}b,\left(\frac{\beta_{22}}{b}\right)^{2}b\right\}.
\]
Consequently,
\[
\|g(z)-g(z')\|_{\mathcal{Z},*}^{2}
\le
4\max\left\{
\frac{\beta_{11}^{2}}{a^{2}},
\frac{\beta_{22}^{2}}{b^{2}},
\frac{\beta_{12}^{2}}{ab},
\frac{\beta_{21}^{2}}{ab}
\right\}
\big(
a\|x-x'\|_{\mathcal{X}}^{2}+b\|y-y'\|_{\mathcal{Y}}^{2}
\big).
\]
Recalling $\|z-z'\|_{\mathcal{Z}}^{2}=a\|x-x'\|_{\mathcal{X}}^{2}+b\|y-y'\|_{\mathcal{Y}}^{2}$ and taking square roots, we obtain
\[
\|g(z)-g(z')\|_{\mathcal{Z},*}
\le
2\max\left\{
\frac{\beta_{11}}{a},
\frac{\beta_{22}}{b},
\frac{\beta_{12}}{\sqrt{ab}},
\frac{\beta_{21}}{\sqrt{ab}}
\right\}
\|z-z'\|_{\mathcal{Z}}.
\]
This completes the proof.

\end{document}